\documentclass{article}

\PassOptionsToPackage{numbers, compress}{natbib}
\usepackage[final]{neurips_2020}

\usepackage[utf8]{inputenc} 
\usepackage[T1]{fontenc}    
\usepackage{hyperref}       
\usepackage{url}            
\usepackage{booktabs}       
\usepackage{amsfonts}       
\usepackage{nicefrac}       
\usepackage{microtype}      

\usepackage[pdftex]{graphicx} 
\usepackage{subcaption}       
\usepackage{xcolor}           
\usepackage{amsmath}          
\usepackage{wrapfig}          
\usepackage{lipsum}           
\usepackage{booktabs}         
\usepackage{multirow}         
\usepackage{wrapfig}          
\usepackage{subcaption}       
\usepackage{amsmath}          
\usepackage{caption}          
\usepackage{cases}            
\usepackage{rotating}         
\newcommand{\comment}[1]{}

\newcommand*\samethanks[1][\value{footnote}]{\footnotemark[#1]}

\title{Meta-trained agents implement Bayes-optimal agents}

\author{%
  Vladimir Mikulik\thanks{Equal contribution}, Gr\'egoire Del\'etang\samethanks, Tom McGrath\samethanks, Tim Genewein\samethanks,\\
  \textbf{Miljan Martic, Shane Legg, Pedro A. Ortega}\thanks{Correspondence to \texttt{\{vmikulik|gdelt|mcgrathtom|timgen|pedroortega\}@google.com}}\\
  DeepMind\\London, UK
}

\begin{document}

\maketitle

\begin{abstract}
Memory-based meta-learning is a powerful technique to build agents that adapt fast to any task within a target distribution. A previous theoretical study has argued that this remarkable performance is because the meta-training protocol incentivises agents to behave Bayes-optimally. We empirically investigate this claim on a number of prediction and bandit tasks. Inspired by ideas from theoretical computer science, we show that meta-learned and Bayes-optimal agents not only behave alike, but they even share a similar computational structure, in the sense that one agent system can approximately simulate the other. Furthermore, we show that Bayes-optimal agents are fixed points of the meta-learning dynamics. Our results suggest that memory-based meta-learning might serve as a general technique for numerically approximating Bayes-optimal agents---that is, even for task distributions for which we currently don't possess tractable models.
\end{abstract}

\section{Introduction}

Within the paradigm of learning-to-learn, memory-based 
meta-learning is a powerful technique to create agents that 
adapt fast to any task drawn from a target 
distribution~\citep{bengio1991learning,
schmidhuber1996simple, thrun1998learning, 
hochreiter2001learning, santoro2016meta, wang2016learning}.
In addition, it has been claimed 
that meta-learning might be a key tool for creating 
systems that generalize to unseen 
environments~\citep{bengio2019ideas}. 
This claim is also partly supported by studies
in computational neuroscience, 
where experimental studies with 
human subjects have shown that fast skill adaptation
relies on task variation \citep{braun2009motor,
braun2010structure}. Due to this, understanding how 
meta-learned agents acquire their representational
structure and perform their computations is of paramount 
importance, as it can inform architectural 
choices, design of training tasks, and address questions about 
generalisation and safety in artificial intelligence.

Previous theoretical work has argued that agents 
that fully optimise a meta-learning objective are 
Bayes-optimal by construction,
because meta-learning objectives are Monte-Carlo approximations of Bayes-optimality objectives~\citep{ortega2019meta}. 
This is striking, as Bayes-optimal 
agents maximise returns (or minimise loss) by optimally
trading off exploration versus exploitation~\citep{duff2002optimal}. 
The theory also makes a stronger, structural claim: namely, 
that meta-trained agents 
perform Bayesian updates ``under the hood'', where the computations are implemented
via a state machine embedded in the memory dynamics that 
tracks the sufficient statistics of the uncertainties
necessary for solving the task class.

Here we set out to empirically investigate
the computational structure of meta-learned agents.
However, this comes with non-trivial challenges. Artificial neural networks are infamous for their hard-to-interpret computational structure: they achieve remarkable performance on challenging tasks, but the computations underlying that performance remain elusive. Thus, while much work in explainable machine learning focuses on the I/O behaviour or memory content, only few investigate the internal dynamics that give rise to them through careful bespoke analysis---see e.g.\ \citep{maheswaranathan2019reverse, maheswaranathan2020recurrent, tanaka2019deep, bau2018identifying, olah2018the, olah2020an, jaderberg2019human}. 

To tackle these challenges, we adapt a relation from theoretical computer science to machine learning systems. Specifically, to compare agents at their computational level \cite{marr2010vision}, we verify whether they can \emph{approximately simulate} each other. The \emph{quality} of the simulation can then be assessed in terms of both state and output similarity between the original and the simulation.

Thus, our main contribution is the investigation of the 
computational structure of RNN-based meta-learned solutions. 
Specifically, we compare the computations of 
meta-learned agents against the computations of 
Bayes-optimal agents in terms of their
behaviour and internal representations on a set 
of prediction and reinforcement learning tasks 
with known optimal solutions. 
We show that on these tasks:
\begin{itemize}
    \item Meta-learned agents \emph{behave} like 
    Bayes-optimal agents (Section~\ref{sec:behavioral_comparison}). That is, the predictions and actions 
    made by meta-learned agents are virtually indistinguishable 
    from those of Bayes-optimal agents.
   
    \item During the course of meta-training, meta-learners \emph{converge} to the Bayes-optimal solution (Section~\ref{sec:convergence-analysis}). We empirically show that Bayes-optimal policies are the fixed points of the learning dynamics.
    
    \item Meta-learned agents \emph{represent} tasks
    like Bayes-optimal agents (Section~\ref{sec:structural-comparison}). 
    Specifically, the computational
    structures correspond to state machines embedded in 
    (Euclidean) memory space, where the states encode 
    the sufficient statistics of the task and produce optimal actions. We can approximately simulate computations performed by meta-learned agents with computations performed by Bayes-optimal agents.
\end{itemize}

\section{Preliminaries}\label{sec:preliminaries}
\paragraph{Memory-based meta-learning}
Memory-based meta-learners are agents with memory that are trained on batches of finite-length
roll-outs, where each roll-out is performed on a task 
drawn from a distribution.
The emphasis on memory is crucial, as training then
performs a search in algorithm space to find a suitable
adaptive policy \citep{kolen2001field}.
The agent is often implemented as a neural network with recurrent
connections, like an RNN, most often using LSTMs
\citep{hochreiter2001learning, gers1999learning}, or GRUs 
\citep{cho2014learning}. Such a network
computes two functions~$f_w$ and $g_w$ using weights~$w \in \mathcal{W}$,
\begin{equation}\label{eq:agent-function}
  \begin{aligned}
    y_t &= f_w(x_t, s_{t-1}) &&\text{(output function)}\\
    s_t &= g_w(x_t, s_{t-1}), &&\text{(state-transition function)}
  \end{aligned}
\end{equation}
that map the current input and previous state pair~$(x_t, s_{t-1}) 
\in \mathcal{X} \times \mathcal{S}$ into the output~$y_t \in \mathcal{Y}$ and 
the next state~$s_t \in \mathcal{S}$ respectively.
Here, $\mathcal{X}$, $\mathcal{Y}$, $\mathcal{S}$, and~$\mathcal{W}$ are
all vector spaces over the reals~$\mathbb{R}$.
An input~$x_t \in \mathcal{X}$ encodes the instantaneous 
experience at time~$t$, such as e.g.\ the 
last observation, action, and feedback signal;
and an output~$y_t \in \mathcal{Y}$ contains e.g.\ the logits
for the current prediction or action probabilities.
RNN meta-learners are typically trained using backpropagation through time (BPTT)~\citep{robinson1987utility, werbos1988generalization}. For fixed weights~$w$, and combined with a fixed initial state~$s_0$, equations \eqref{eq:agent-function} define a state machine\footnote{More precisely, a Mealy machine \citep{mealy1955method, sipser1996introduction,  savage1998models}.}. This state machine can be seen as an adaptive policy or an online learning algorithm.

\paragraph{Bayes-optimal policies as state machines}
Bayes-optimal policies have a natural interpretation as state machines following~\eqref{eq:agent-function}. Every such policy can be seen as a state-transition function~$g$, which maintains sufficient statistics (i.e.,\ a summary of the past experience that is statistically sufficient to implement the prediction/action strategy) and an output function~$f$, which uses this information to produce optimal outputs (i.e.,\ the best action or prediction given the observed trajectory) \citep{duff2002optimal,
ortega2019meta}. For instance, to implement an optimal policy for a multi-armed bandit with (independent) Bernoulli rewards, it is sufficient to remember the number of successes and failures for each arm.

\paragraph{Comparisons of state machines via simulation}
To compare the policies of a meta-trained and a Bayes-optimal agent in terms of their \emph{computational structure}, we adapt a well-established methodology from the \emph{state-transition systems} literature \citep{clarke2018model, Baier2008PrinciplesOM, BAETEN2014399}. Specifically, we use the concept of \emph{simulation} to compare state machines. 

\comment{Intuitively, similarity between two state machines means that they realise the same underlying algorithm.\footnote{There exists a finer equivalence, \emph{bisimilarity}. We conjecture that our meta-learners are indeed approximately bisimilar to Bayes-optimal agents, but do not show so explicitly in this work.}}

Formally, we have the following. A \emph{trace} in a state machine is a sequence~$s_0 x_1 s_1 \cdots x_T s_T$ of transitions. Since the state machines we consider are deterministic, a given sequence of inputs $x_1, \ldots, x_T$ induces a unique trace in the state machine. A deterministic state machine~$M$ \emph{simulates} another machine~$N$, written $N \preceq M$, if every trace in~$N$ has a corresponding trace in~$M$ on which their output functions agree. More precisely, $N \preceq M$ if there exists a function~$\phi$ mapping the states of~$N$ into the states of~$M$ such that the following two conditions hold:
\begin{itemize}
    \item (\emph{transitions}) for any trace $s_0 x_1 s_1 \cdots s_T$ in $N$, the transformed trace $\phi(s_0) x_1 \phi(s_1) \cdots \phi(s_T)$ is also a trace in $M$; 
    \item (\emph{outputs}) for any state $s$ of $N$ and any input $x$, the output of machine $N$ at $(x,s)$ coincides with the output of machine $M$ at $(x,\phi(s))$.
\end{itemize}
Intuitively, this means there is a consistent way of interpreting every state in $N$ as a state in $M$, such that every computation in $N$ can be seen as a computation in $M$. When both $M \preceq N$ and $N \preceq M$ hold, then we consider both machines to be computationally equivalent.

\section{Methods}
\subsection{Tasks and agents}
\paragraph{Tasks}
Since our aim is to compare against Bayes-optimal policies, we consider~10 prediction and 4~reinforcement learning tasks for which the Bayes-optimal solution is analytically tractable. All tasks are episodic ($T=20$ time steps), and the task parameters~$\theta$ are drawn from a prior distribution~$p(\theta)$ at the beginning of each episode. A full list of tasks is shown in Figure~\ref{fig:comparison_all_tasks} and details are discussed in Appendix~\ref{sec:task-details}.

In prediction tasks the goal is to make probabilistic predictions of the next observation given past observations. All observations are drawn i.i.d.\ from an observational distribution. To simplify the computation of the optimal predictors, we chose observational distributions within the exponential family that have simple conjugate priors and posterior predictive distributions, namely: Bernoulli, categorical, exponential, and Gaussian. In particular, their Bayesian predictors have finite-dimensional sufficient statistics with simple update rules \citep{raiffa1961applied, bishop2006pattern, gelman2013bayesian}.

In reinforcement learning tasks the goal is to maximise the discounted cumulative sum of rewards in two-armed bandit problems~\citep{lattimore2018bandit}. We chose bandits with rewards that are Bernoulli- or Gaussian-distributed. The Bayes-optimal policies for these bandit tasks can be computed in polynomial time by pre-computing \emph{Gittins indices} \citep{gittins1979bandit, lattimore2018bandit, edwards2016towards}. Note that the bandit tasks, while conceptually simple, already require solving the exploration versus exploitation problem \citep{sutton2018reinforcement}.

\paragraph{RNN meta-learners}
Our RNN meta-learners consist of a three-layer network architecture: one fully connected layer (the encoder), followed by one LSTM layer (the memory), and one fully connected layer (the decoder) with a linear readout producing the final output, namely the parameters of the predictive distribution for the prediction tasks, and the logits of the softmax action-probabilities for the bandit tasks respectively. The width of each layer is the same and denoted by~$N$. We selected\footnote{Note that the (effective) network capacity needs to be large enough to at least represent the different states required by the Bayes-optimal solution. However, it is currently unknown how to precisely measure effective network capacity. We thus selected our architectures based on preliminary ablations that investigate convergence speed of training. See Appendix~\ref{sec:architecture-ablation} for details.} $N=32$ for prediction tasks and $N=256$ for bandit tasks. Networks were trained with BPTT~\citep{robinson1987utility, werbos1988generalization} and Adam~\citep{kingma2014adam}. In prediction tasks the loss function is the log-loss of the prediction. In bandit tasks the agents were trained to maximise the return (i.e.,\ the discounted cumulative reward) using the Impala \citep{espeholt2018impala} policy gradient algorithm. See Appendix \ref{sec:rnn-details} for details on network architectures and training.

\subsection{Behavioral analysis}\label{sec:behavioural-analysis-methods}

The aim of our behavioural analysis is to compare the input-output behaviour of a meta-learned~(${\rm RNN}$) and a Bayes-optimal agent~(${\rm Opt}$). For prediction tasks, we feed the same observations to both agent types and then compute their dissimilarity as the sum of the KL-divergences of the instantaneous predictions averaged over~$K$ trajectories, that is,
\begin{equation}
    d({\rm Opt}, {\rm RNN})
    = \frac{1}{K}\sum_{k=1}^K \sum_{t=1}^T D_{\rm KL}\bigl(
        \pi^{\rm Opt}_t \bigl\| \pi^{\rm RNN}_t \bigr) \label{eq:supervised-similarity}.
\end{equation}

Bandit tasks require a different dissimilarity measure: since there are multiple optimal policies, we cannot compare action probabilities directly. A dissimilarity measure that is invariant under optimal policies is the empirical reward difference:
\begin{equation}
    d({\rm Opt}, {\rm RNN})
    = \Bigl| \frac{1}{K}\sum_{k=1}^K \sum_{t=1}^T (r^{\rm Opt}_t - r^{\rm RNN}_t) \Bigr|
    \label{eq:bandit-similarity}
\end{equation}
where $r^{\rm Opt}$ and $r^{\rm RNN}$ are the empirical rewards collected during one episode. This dissimilarity measure only penalises policy deviations that entail reward differences.

\subsection{Convergence analysis}\label{sec:convergence-analysis-methods}

In our convergence analysis we investigate how the behaviour of meta-learners changes over the course of training. To characterise how a single RNN training run evolves, we evaluate the behavioural dissimilarity measures (Section~\ref{sec:behavioural-analysis-methods}), which compare RNN behaviour against Bayes-optimal behaviour, across many checkpoints of a training run. 
Additionally we study the RNN behaviour across \emph{multiple} training runs, which allows us to characterise convergence towards the Bayes-optimal solution. For this we use several RNN training runs (same architecture, different random initialisation), and at fixed intervals during training we compute pairwise behavioural distances between all meta-learners and the Bayes-optimal agent. The behavioural distance is computed using the Jensen-Shannon divergence\footnote{The Jensen-Shannon divergence is defined as $D_\text{JS}(X||Y) = \frac{1}{2}(D_\text{KL}(X||M) + D_\text{KL}(Y||M))$, where $M$ is the mixture distribution $(X + Y)/2$.} for prediction tasks and the absolute value of the cumulative regret for bandits. We visualise the resulting distance matrix in a 2D plot using multidimensional scaling (MDS)~\cite{borg2005modern}. 

\subsection{Structural analysis}\label{sec:structural-analysis-methods}

We base our structural analysis on the idea of simulation introduced in Section~\ref{sec:preliminaries}. Since here we also deal with continuous state, input, and output spaces, we relax the notion of simulation to \emph{approximate simulation}:
\begin{itemize}
    \item (\emph{Reference inputs}) As we cannot enumerate all the traces, we first sample a collection of input sequences from a reference distribution and then use the induced traces to compare state machines.
    \item (\emph{State and output comparison}) To assess the quality of a simulation, we first learn a map $\phi$ that embeds the states of one state machine into another, and then measure the dissimilarity. To do so, we introduce two measures of dissimilarity~$D_s$ and~$D_o$ to evaluate the state and output dissimilarity respectively. More precisely, consider assessing the quality of a state machine~$M$ simulating a machine~$N$ along a trace induced by the input sequence $x_1 \cdots x_T$. Then, the quality of the state embedding $D_s$ is measured as the mean-squared-error (MSE) between the embedded states $\phi(\mathcal{S}_N) \subset \mathcal{S}_M$ and the states $\mathcal{S}_M$ of~$M$ along the trace. Similarly, the quality of the output simulation $D_o$ is measured as the dissimilarity between the outputs generated from the states $\mathcal{S}_N$ and $\phi(\mathcal{S}_N)$ along the trace, that is, before and after the embedding respectively.
\end{itemize} 

In practice, we evaluate how well e.g.\ a meta-learned agent simulates a Bayes-optimal one by first finding an embedding~$\phi$ mapping Bayes-optimal states into meta-learned states  that minimises the state dissimilarity $D_s$, and then using said embedding to compute the output dissimilarity $D_o$. The mapping~$\phi$ is implemented as an MLP---see details in Appendix~\ref{sec:structural-comparison-details}.
We use \eqref{eq:supervised-similarity} and \eqref{eq:bandit-similarity} as output dissimilarity measures~$D_o$ in prediction and bandit tasks respectively.

Our approach is similar in spirit to~\citep{girard2005approximate}, but adapted to work in continuous observation spaces. 

\newpage
\section{Results}
\subsection{Behavioral Comparison}
\label{sec:behavioral_comparison}
\begin{figure}[ht]
    \centering
    \begin{subfigure}{\textwidth}
        \centering
        \parbox[b]{.58\textwidth}{\includegraphics[width=1\linewidth]{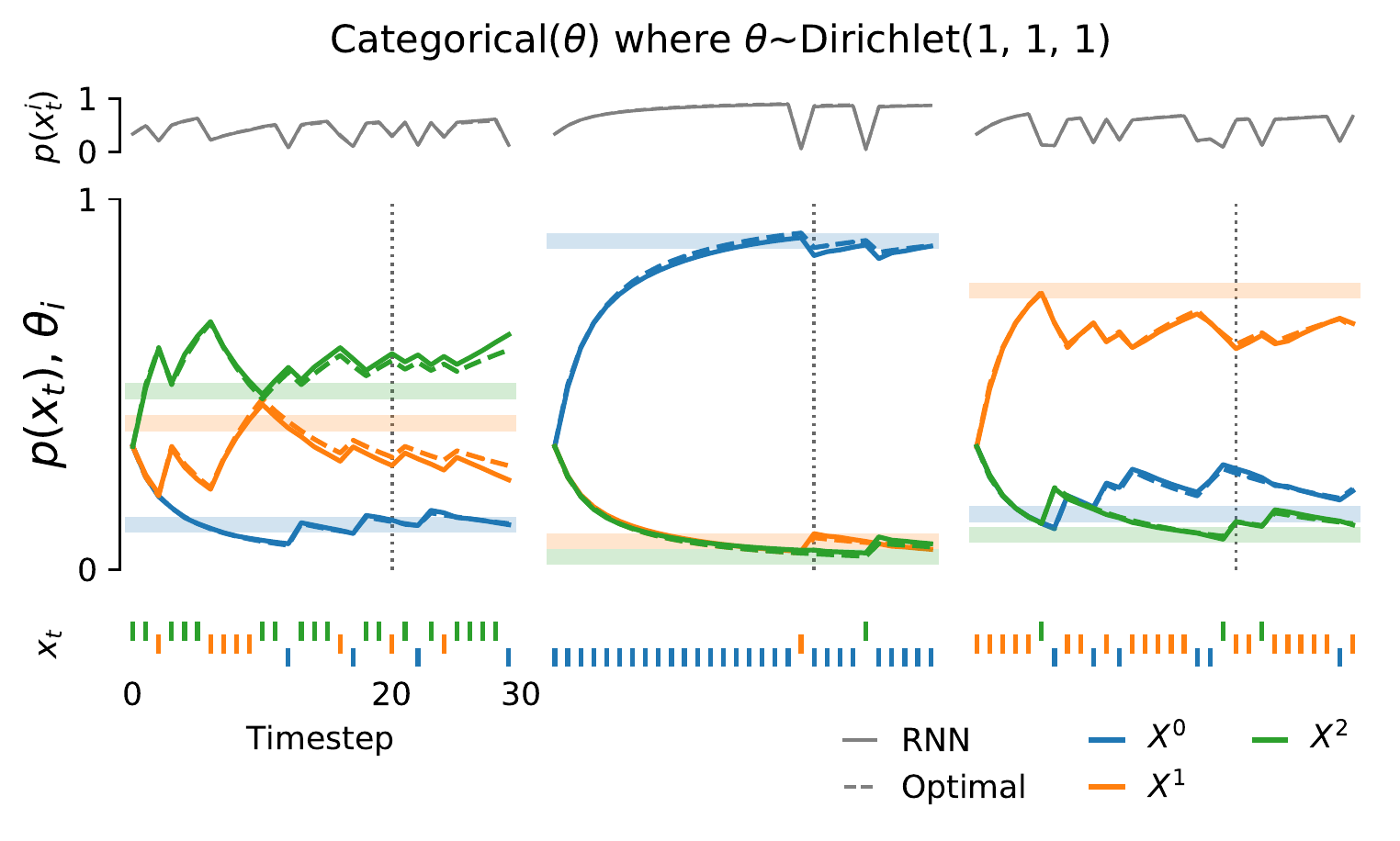}}
        \hfill%
        \parbox[b]{.4\textwidth}{%
        \subcaption{Categorical-variable prediction task. Bottom row shows observations, middle row shows predictions of the meta-learner (solid) and the Bayes-optimal predictor (dashed), that converge towards the environment's true values (shaded bars). Top-row shows the likelihood of the observation under the predictive distribution.
        }
        \label{fig:behavior_categorical}}
    \end{subfigure}
    \hfill
    \begin{subfigure}{\textwidth}
        \centering
        \parbox[b]{.58\textwidth}{\includegraphics[width=1\linewidth]{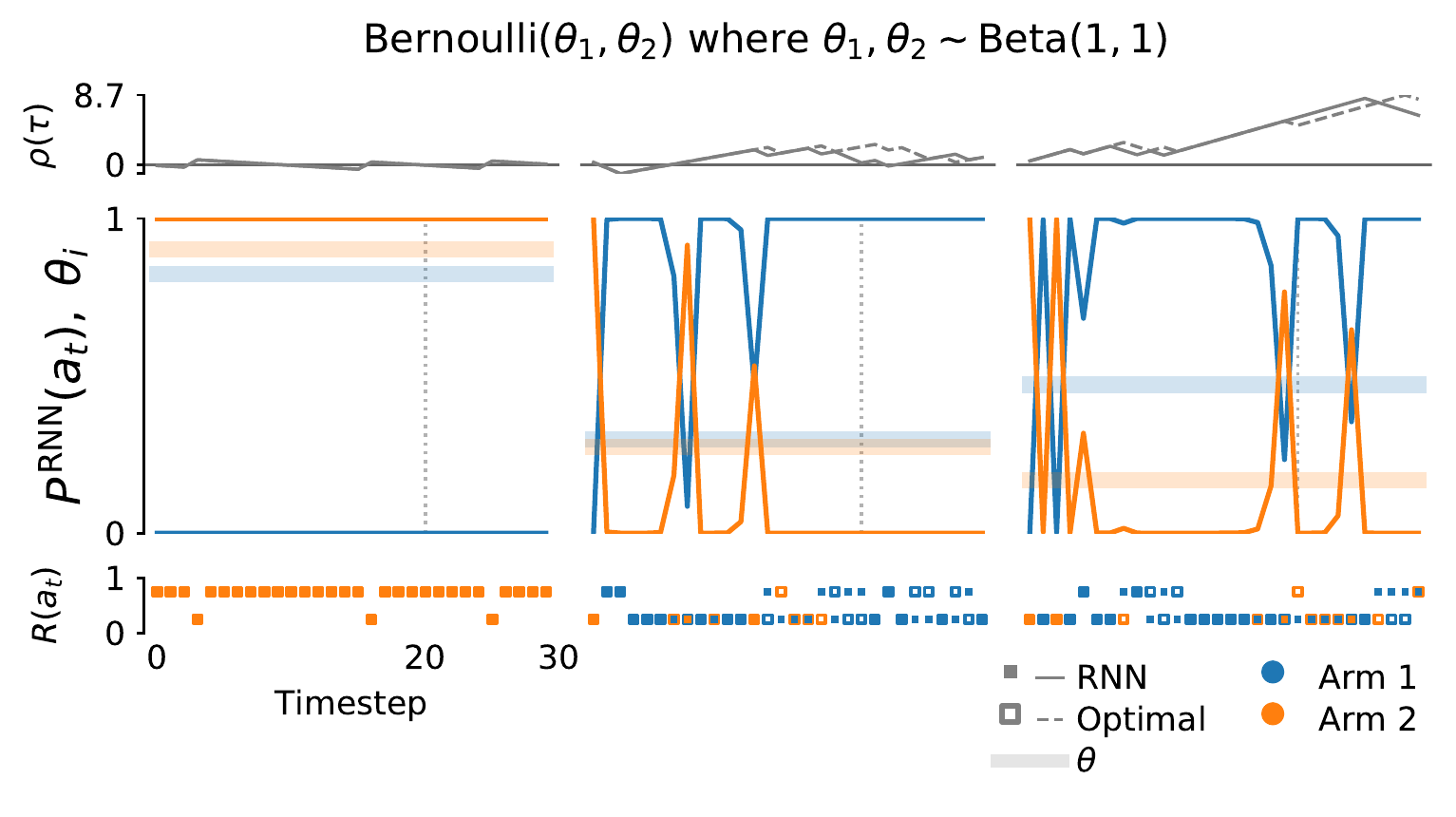}}
        \hfill%
        \parbox[b]{.4\textwidth}{%
        \subcaption{$2$-armed Bernoulli-bandit task. Bottom row shows inputs (action-reward pairs). Middle row shows action probabilities for meta-learner (solid) and expected arm payoffs as shaded bars. Action probabilities of the Bayes-optimal agent are not shown as it acts deterministically, selecting actions according to the highest Gittins index per time step. Top row shows cumulative regret for both agent-types.}
        \label{fig:behavior_bernoulli_bandit}}
    \end{subfigure}
    \caption{\footnotesize Illustrative behavioral comparison of a meta-learned agent and the Bayes-optimal agent on $3$ episodes (same environment random seed for both agents).  Meta-learned agents were trained with only $20$ time-steps; thus these results illustrate that the RNN generalizes to $30$ time-steps.
    }
    \label{fig:behavior}
\end{figure}

To compare the behavior between meta-learned and Bayes-optimal agents, we contrast their outputs for the same inputs. Consider for instance the two agents shown in Figure~\ref{fig:behavior}. Here we observe that the meta-learned and the Bayes-optimal agents behave in an almost identical manner: in the prediction case (Figure~\ref{fig:behavior_categorical}), the predictions are virtually indistinguishable and approach the true probabilities; and in the bandit case (Figure~\ref{fig:behavior_bernoulli_bandit}) the cumulative regrets are essentially the same\footnote{Recall that the regret is invariant under optimal policies.} whilst the policy converges toward pulling the best arm.

To quantitatively assess behavioral similarity between the meta-learners and the Bayes-optimal agents, we use the measures introduced in Section~\ref{sec:behavioural-analysis-methods}, namely \eqref{eq:supervised-similarity} for prediction tasks and \eqref{eq:bandit-similarity} for bandit tasks. For each task distribution, we averaged the performance of~$10$ meta-learned agents. 
The corresponding results in Figure~\ref{fig:comparison_all_tasks}a show that the trained meta-learners behave virtually indistinguishably from the Bayes-optimal agent. Results for reduced-memory agents, which cannot retain enough information to perform  optimally, are shown in Appendix~\ref{ss:reduced-mem}.

\subsection{Convergence}

\label{sec:convergence-analysis}
\begin{figure}[ht]
    \vspace{-10pt}
    \centering
    \begin{subfigure}{0.48\textwidth}
        \centering
        \includegraphics[width=\textwidth]{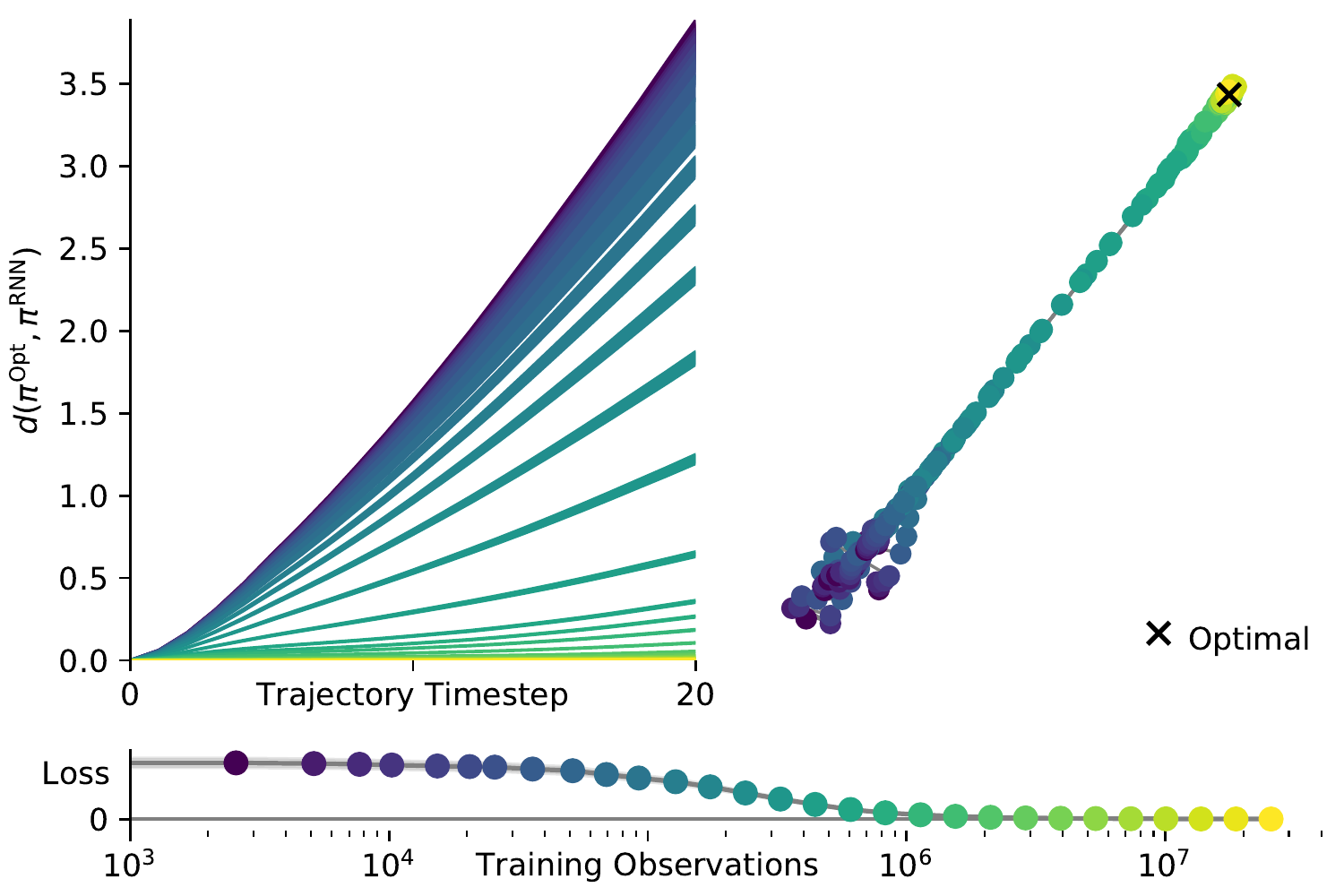}
        \vspace{-15pt}
        \caption{Categorical prediction task with parameters drawn from $\mathrm{Dirichlet}(1,1,1)$.}
        \label{fig:convergence_supervised}
    \end{subfigure}
    \hfill
    \begin{subfigure}{0.48\textwidth}
        \centering
        \includegraphics[width=\textwidth]{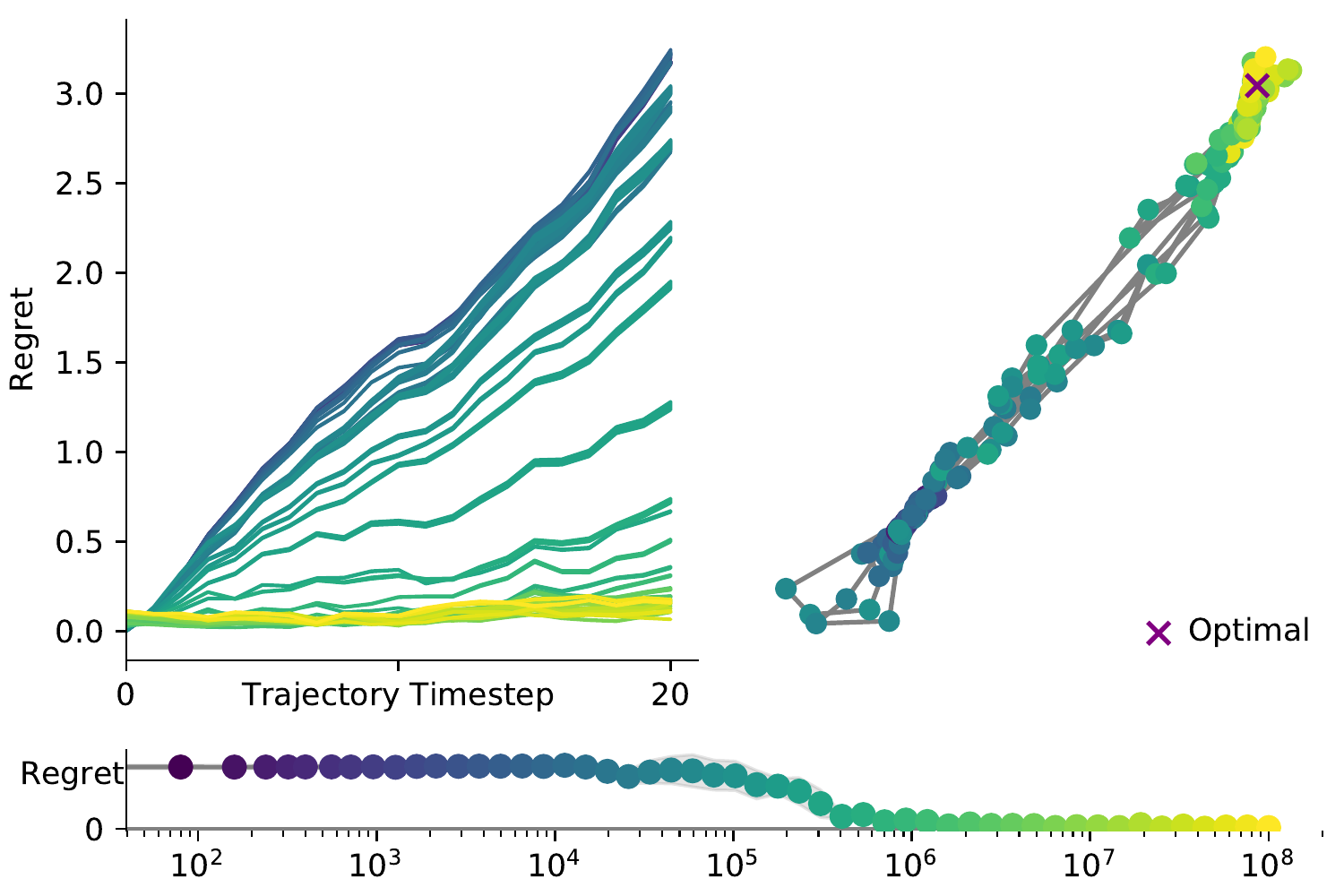}
        \vspace{-15pt}
        \caption{Two-armed Bernoulli-bandit task with biases drawn from $\mathrm{Beta}(1,1)$.}
        \label{fig:convergence_bandit}
    \end{subfigure}
    \caption{Policies evolve similarly towards the Bayes-optimal policy over the course of training for both the prediction~(a) and the bandit task~(b). Panels in each subfigure show, (clockwise from top left): evolution of the within-episode dissimilarity from the Bayes-optimal policy, averaged over $500$ trajectories; the evolution of~10 policies for different training runs (multidimensional scaling visualisation of pairwise behavioural distances; each curve is a separate run); and the training curves for the log-loss and regret respectively.}
    \label{fig:convergence}
\end{figure}

We investigate how the behavior of meta-learners changes over the course of training. Following the methodology introduced in Section~\ref{sec:convergence-analysis-methods} we show the evolution of behavior of a single training run (top-left panels in Figure~\ref{fig:convergence_supervised},~\ref{fig:convergence_bandit}). These results allow us to evaluate the meta-learners' ability to pick up on the environment's prior statistics and perform Bayesian evidence integration accordingly. As training progresses, agents learn to integrate the evidence in a near-optimal manner over the entire course of the episode. However, during training the improvements are not uniform throughout the episode. This `staggered' meta-learning, where different parts of the task are learned progressively, resembles results reported for meta-learners on nonlinear regression tasks in~\cite{rabinowitz2019meta}.

We also compared behavior across multiple training runs (top-right panels in Figure~\ref{fig:convergence_supervised},~\ref{fig:convergence_bandit}). Overall the results indicate that after some degree of heterogeneity early in training, all meta-learners converge in a very similar fashion to the Bayes-optimal behavior. This is an empirical confirmation of the theoretical prediction in~\cite{ortega2019meta} that the Bayes-optimal solution is the fixed-point of meta-learner training. Appendix~\ref{sec:structural-comparison-more} shows the convergence results for all tasks.

\subsection{Structural Comparison}
\label{sec:structural-comparison}
In this section we analyze the computational structure of the meta-learner who uses its internal state to store information extracted from observations required to act.

Following the discussion in Section~\ref{sec:structural-analysis-methods}, we determine the computational similarity of the meta-learning and Bayes-optimal agents via simulation. Our analysis is performed by projecting and then whitening both the RNN state (formed by concatenating both the cell- and hidden-states of the LSTM) and the Bayes-optimal state onto the first~$n$ principal components, where~$n$ is the dimensionality of the Bayes-optimal state/sufficient statistics. We find that these few components suffice to explain a large fraction of the variance of the RNN agent's state---see Appendix~\ref{sec:variance_explained}.
We then regress an MLP-mapping~$\phi$ from one (projected) agent state onto the other and compute $D_s$ and $D_o$. Importantly, this comparison is only meaningful if we ensure that both agents were exposed to precisely the same input history. This is easily achieved in prediction tasks by fixing the environment random seed. In bandit tasks we ensure that both agents experience the same action-reward pairs by using the trained meta-learner to generate input streams that are then also fed into the Bayes-optimal agent.

\begin{figure}[htb]
    \vspace{-10pt}
    \centering
    \begin{subfigure}{0.48\textwidth}
        \centering
        \includegraphics[width=\textwidth]{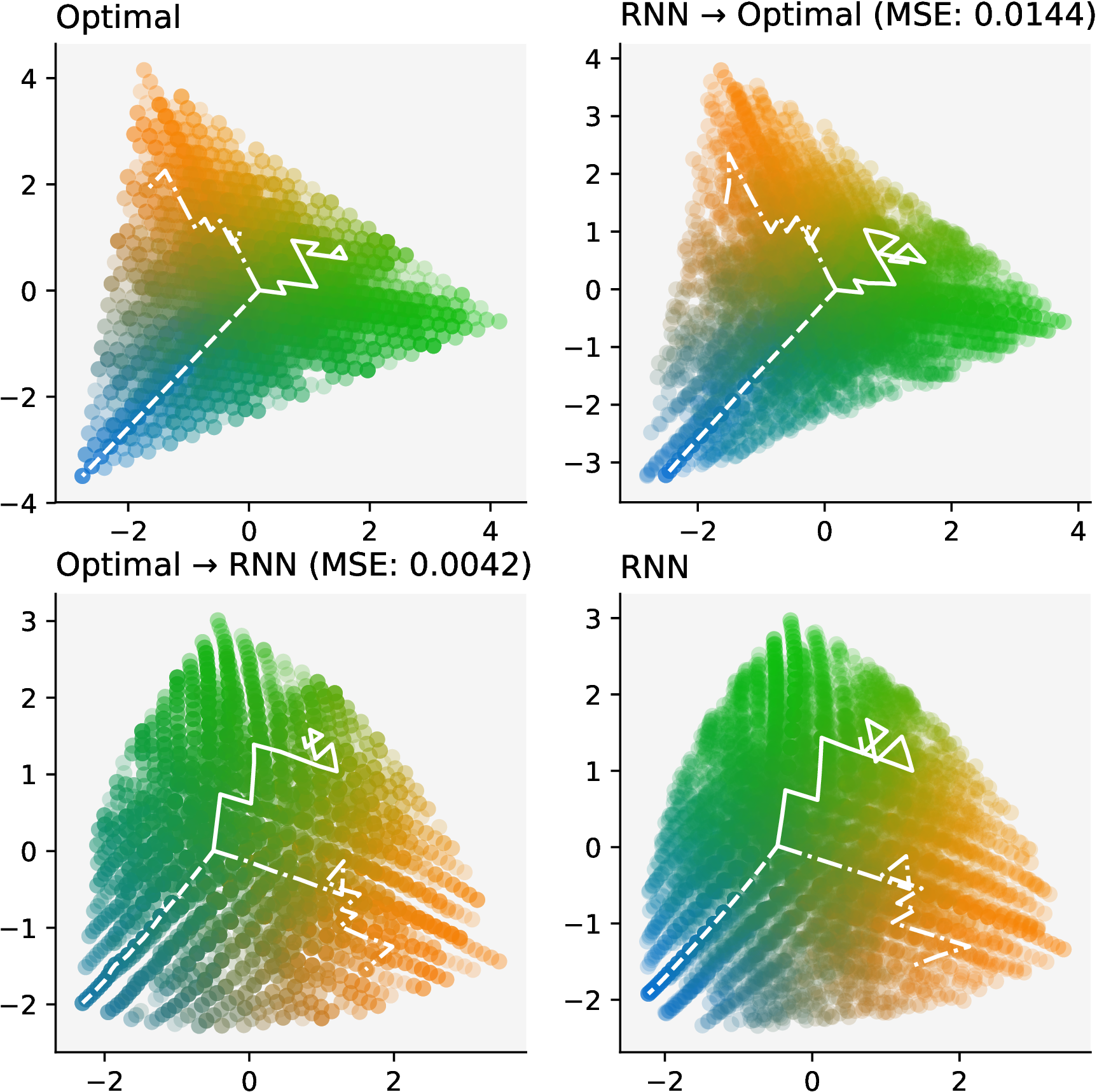}
        \caption{Categorical prediction task with parameters drawn from $\mathrm{Dirichlet}(1,1,1)$. The colors indicate the prediction probabilities emitted in each state. Three episode rollouts are shown.}
        \label{fig:structure_categorical}
    \end{subfigure}
    \hfill
    \begin{subfigure}{0.48\textwidth}
        \centering
        \includegraphics[width=\textwidth]{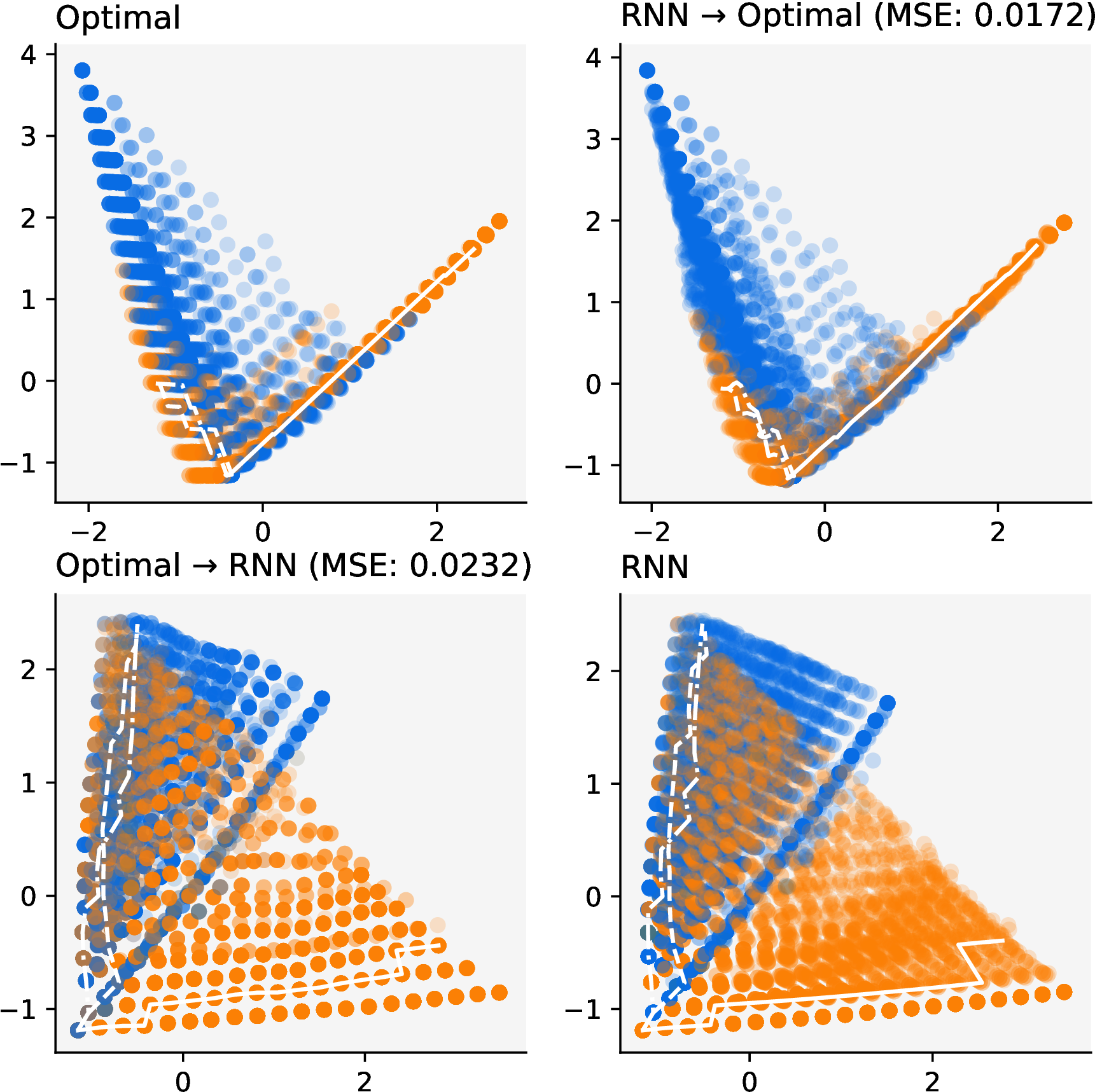}
        \caption{Two-armed Bernoulli-bandit task with biases drawn from $\mathrm{Beta}(1,1)$. The colors indicate the action-probabilities emitted in each state. Three episode rollouts are shown.}
        \label{fig:structure_gaussian_bandit}
    \end{subfigure}
    \caption{Structural comparison. Each sub-figure depicts: two agent state spaces, namely of the Bayes-optimal (top-left) and RNN states (bottom-right), projected onto the first two principal components; and two simulations, i.e.,\ the learned embeddings from the RNN into the Bayes-optimal states (top-right) and from the Bayes-optimal into the RNN states (bottom-left). The scores in the simulations indicate the MSE of the learned regression. The outputs emitted in each state are color-coded. Note that the color-codings in the simulations result from evaluating the output at the (potentially high-dimensional) embedded state (see Section~\ref{sec:structural-analysis-methods}). White lines indicate the same three episodes as shown in Figure~\ref{fig:behavior}.}
    \label{fig:structure}
\end{figure}

Figure~\ref{fig:structure} illustrates our method for assessing the computational similarity. We embedded\footnote{The embeddings were implemented as MLPs having three hidden layers with either 64 (prediction) or 256 (bandits) neurons each.} the state space of the Bayes-optimal agent into the state space of the meta-learned agent, and then we calculated the output from the embedded states. This embedding was also performed in the reverse direction. Visual inspection of this figure suggests that the meta-learned and the Bayes-optimal agents perform similar computations, as the panels resemble each other both in terms of states and outputs. We observed similar results for all other tasks (Appendix~\ref{sec:structural-comparison-more}). In contrast, we have observed that the computational structure of \emph{untrained} meta-learners does not resemble the one of Bayes-optimal agents (Appendix~\ref{sec:pca_random_init}). 

\begin{figure}[htb]
    \vspace{-10pt}  
    \centering
    \includegraphics[width=1\textwidth]{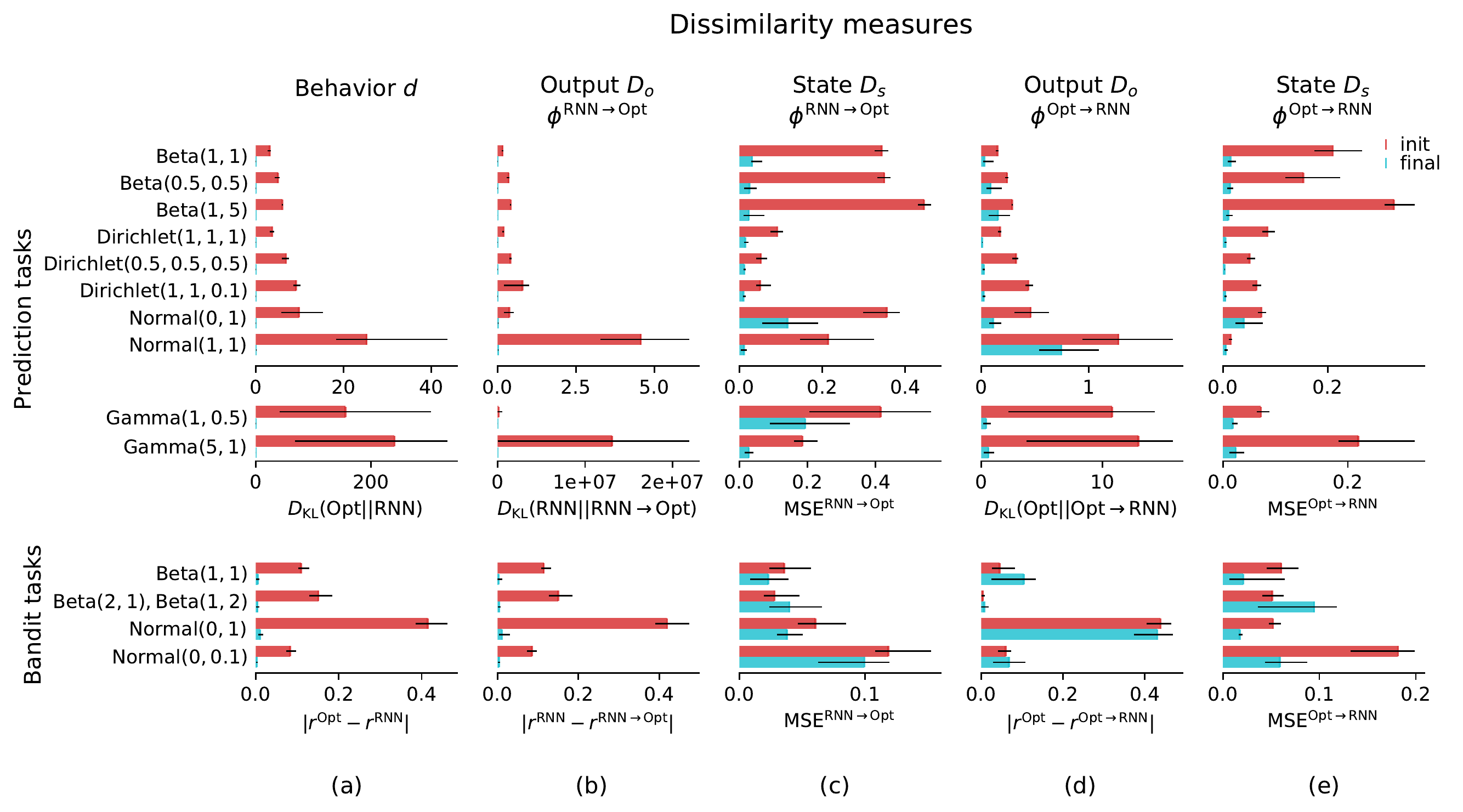}
    \vspace{-15pt}
    \caption{Behavioral and structural comparison for all tasks. Figure shows similarity measures across $K=500$ episodes of length $T=20$, and $10$ different training runs of the meta-learner (bars show median over training-runs, error bars denote $5$-$95$ quantiles). `init' denotes the untrained meta-learner, `final' denotes evaluation at the end of training. Numerical results are shown in Figure~\ref{fig:comparison_all_tasks_annotated} in the Appendix. Column~a: Behavioral dissimilarity between meta-learned agent and Bayes-optimal agent (see Section~\ref{sec:behavioral_comparison}). Columns~b \&~c: State- and Output-dissimilarity for $\mathrm{RNN}\rightarrow\mathrm{Opt}$. Columns~d \&~e: State- and Output-dissimilarity for $\mathrm{Opt}\rightarrow\mathrm{RNN}$. Low values of the state- and output-dissimilarity measures (simultaneously) indicate that the state machines implemented by $\mathrm{RNN}$ and $\mathrm{Opt}$ are structurally equivalent.}
    \label{fig:comparison_all_tasks}
    \vspace{-10pt}
\end{figure}

The quantitative results for the structural comparison for all tasks across~$10$ repetitions of meta-training are shown in Figure~\ref{fig:comparison_all_tasks}.
We find that for the trained meta-learner state-dissimilarity~$D_s$ is low in almost all cases. In bandit tasks, $D_s$ tends to be slightly larger in magnitude which is somewhat expected since the RNN-state dimensionality is much larger in bandit tasks. Additionally there is often no significant difference in $D_s$ between the untrained and the final agent---we suspect this to be an artefact of a reservoir effect~\citep{maass2004computational} (see Discussion). The output-dissimilarity~$D_o$ is low for both task types for $\mathrm{RNN}\rightarrow \mathrm{Opt}$, but not in the reverse direction. 
This indicates that the meta-learners are very well simulated by the Bayes-optimal agents, since both the state dissimilarity $D_s$ and the output dissimilarity~$D_o$ are almost negligible. In the reverse direction however, we observe that the meta-learned solutions do not always simulate the Bayes-optimal with high accuracy, as seen by the non-negigible output dissimilarity~$D_o$. We believe that this is because the sufficient statistics learned by the meta-learners are not minimal.

\section{Discussion and conclusions}
In this study we investigated whether memory-based meta-learning leads to solutions that are behaviourally and structurally equivalent to Bayes-optimal predictors. We found that behaviorally the Bayes-optimal solution constitutes a fixed-point of meta-learner training dynamics. Accordingly, trained meta-learners behave virtually indistinguishable from Bayes-optimal agents. We also found structural equivalence between the two agent types to hold to a large extent: meta-learners are well simulated by Bayes-optimal agents, but not necessarily vice versa. This failure of simulation is most likely a failure of injectivity: if a single state in one agent must be mapped to two distinct states in another then simulation is impossible. This occurs when two trajectories lead to the same state in one agent but not another (for instance if exchangeability has not been fully learned). We suspect that RNN meta-learners represent non-minimal sufficient statistics as a result of training. For instance, for Bernoulli prediction tasks the input sequences \emph{heads-tails-heads}, and \emph{tails-heads-heads} induce the same minimal sufficient statistics and thus lead to precisely the same internal state in the Bayes-optimal agent, but might lead to different states in the RNN agent. From a theoretical point of view this is not unexpected, since there is no explicit incentive during RNN training that would force representations to be minimal. Note that overly strong regularization can reduce the RNN's effective capacity to a point where it can no longer represent the number of states required by Bayes-optimal solution, which of course strictly rules out computational equivalence. 

A related issue can be observed in bandit tasks: even untrained meta-learners show low state-dissimilarity. We hypothesize that this is due to a ``reservoir effect''~\citep{maass2004computational}, that is the dynamics of high-dimensional untrained RNNs are highly likely to map each input history to a unique trajectory in memory space. Accordingly, the untrained RNN ``memorizes'' inputs perfectly---a verbose representation of the task's sufficient statistics.

Our work contributes to understanding the structure and computations implemented by recurrent neural networks. Focusing analysis on computational equivalence, as in our work, opens up the future possibility of separating different, heterogeneous agents into meaningful sets of equivalent classes, and study universal aspects of these agent-classes.

\subsection{Related work}
In this paper we study memory-based meta-learning through a Bayesian lens, showing that meta-learning objectives naturally induce Bayes-optimal behaviour at convergence. A number of previous works have attempted to devise new recurrent architectures to perform Bayes filtering in a number of settings, including time series prediction~\cite{lim2019recurrent}, state space modelling~\cite{krishnan2017structured}, and Kalman filtering~\cite{coskun2017long}. Other previous work has attempted to improve memory-based meta-learners' abilities by augmenting them with a memory, and using weights which adapt at different speeds~\cite{munkhdalai2017meta, munkhdalai2019metalearned}.

Another approach to meta-learning is optimiser-based meta-learning such as MAML~\citep{finn2017model}. In optimiser-based meta-learning models are trained to be able to adapt rapidly to new tasks via gradient descent. MAML has been studied from a Bayesian perspective, and shown to be a hierarchical Bayesian model~\cite{grant2018recasting}. Recent work suggests that solutions obtained by optimiser-based meta-learning might be more similar to those from memory-based meta-learning than previously thought~\citep{Raghu2020Rapid} .

In this paper we relate memory-based meta-learning to finite-state automata which track sufficient statistics of their inputs. The field of computational mechanics~\cite{shalizi2001computational} studies predictive automata (known as \emph{$\varepsilon$-machines)} which track the state of a stochastic process in order to predict its future states. The states of $\varepsilon$-machines are referred to as causal states, and have recently been used to augment recurrent agents in POMDPs~\cite{zhang2019learning}. Finite-state automata have also been considered as a model for decision-making agents in the \emph{situated automata} work of Rosenschein and Kaelbling~\cite{kaelbling1990action, rosenschein1995situated}. The states of situated automata track logical propositions about the state of the world instead of having a probabilistic interpretation, but are naturally suited to goal-directed agents.

There is considerable work on understanding recurrent neural networks on natural language tasks~\citep{belinkov2019analysis}, and in neuroscience~\citep{sohn2019bayesian, maheswaranathan2019universality, Dasgupta644534}, e.g. how relations between multiple trained models can illuminate computational mechanisms~\citep{bau2018identifying}, and the dynamics involved in contextual processing~\citep{maheswaranathan2020recurrent}. Computational analysis of internal dynamics of reinforcement learning agents has received less attention in the literature, though there are some notable examples: a multi-agent setting~\citep{jaderberg2019human} and Atari games~\citep{zahavy2016graying}. Using a related formalism to our approach, the authors of~\cite{koul2018learning} extract minimal finite-state machines (FSM) from the internal dynamics of Atari-agents. However their focus is on extracting small human-interpretable FSM, whereas we compare the computational structure of two agents in a fully automated, quantitative fashion.

In recent years a diverse range of tools to allow interpretability and explainability of deep networks have been developed, including saliency maps~\cite{erhan2009visualizing, simonyan2013deep, zeiler2014visualizing, shrikumar2016not, selvaraju2016grad, smilkov2017smoothgrad}, manual dissection of individual units~\cite{olah2020an, olah2018the, bau2020understanding} and training explainable surrogate models to mimic the output of deep networks~\cite{ribeiro2016should, koul2018learning}. Although our focus here is different - we seek to establish how a broad class of architectures behaves on a family of tasks, rather than explaining a specific network - the closest parallel is with the use of surrogate explainable models. In this case, the Bayes-optimal agent serves as an understood model, and we relate its (well-understood) behaviour to that of the meta-trained agent.

\vspace{-10pt}  
\paragraph{Scope and limitations}
We performed our empirical comparison on a range of tasks where optimal solutions are analytically and computationally tractable. The latter is typically no longer true in more complex tasks and domains. However, the simulation methodology used in this paper could be useful to compare agent-types against each other in more general settings, as it does not rely on either agent being Bayes-optimal. While most aspects of our methodology scale up well to more complex agents, the main difficulty is generating reference trajectories that cover a large (enough) fraction of possible experiences. Finally, our results show that when optimal policies are in the search space, and training converges to those policies, then the resulting policy will be Bayes-optimal. In more complex cases, one or both of these assumptions may no longer hold. Further study is needed to understand the kind of suboptimal solutions that are generated by meta-learning in this case.

\vspace{-10pt}  
\paragraph{Conclusions}
Our main contribution is to advance the understanding of RNN-based meta-learned solutions. We empirically confirm a recently published theoretical claim~\citep{ortega2019meta} that fully-converged meta-learners and Bayes-optimal agents are computationally equivalent. In particular, we showed that RNN meta-learners converge during training to the Bayes-optimal solution, such that trained meta-learners behave virtually indistinguishably from Bayes-optimal agents. Using a methodology related to the concept of \emph{simulation} in theoretical computer science, we additionally show (approximate) structural equivalence of the state-machines implemented by the RNN meta-learners and the Bayes-optimal agent. Our results suggest that memory-based meta-learning will drive learned policies towards Bayes-optimal behaviour, and will converge to this behaviour where possible.

\section{Broader Impact}
Our work helps advance and verify the current understanding of the nature of solutions that meta-learning brings about (our empirical work focused on modern recurrent neural network architectures and training algorithms, but we expect the findings to qualitatively hold for a large range of AI systems that are trained through meta-learning). Understanding how advanced AI and ML systems work is of paramount importance for safe deployment and reliable operation of such systems. This has also been recognized by the wider machine-learning community with a rapidly growing body of literature in this emerging field of ``Analysis and Understanding'' of deep learning. While increased understanding is likely to ultimately also contribute towards building more capable AI systems, thus potentially amplifying their negative aspects, we strongly believe that the merits of understanding how these systems work clearly outweigh the potential risks in this case.

We argue that understanding meta-learning on a fundamental level is important, since meta-learning subsumes many specific learning tasks and is thought to play an important role for AI systems that generalize well to novel situations. Accordingly we expect meta-learning to be highly relevant over the next decade(s) in AI research and in the development of powerful AI algorithms and applications. 
In this work we also show a proof-of-concept implementation for analysis methods that might potentially allow one to separate (heterogeneous) agents into certain equivalence classes, which would allow to safely generalize findings about an individual agent to the whole equivalence class. We believe that this might open up interesting future opportunities to boost the generality of analysis methods and automatic diagnostic tools for monitoring of AI systems.

\begin{ack}
We thank Jane Wang and Matt Botvinick for providing helpful comments on this work.
\end{ack}

\bibliographystyle{unsrt}
\bibliography{main}

\begin{thebibliography}{10}

\bibitem{bengio1991learning}
Y~Bengio, S~Bengio, and J~Cloutier.
\newblock Learning a synaptic learning rule.
\newblock In {\em IJCNN-91-Seattle International Joint Conference on Neural
  Networks}, volume~2, pages 969--vol. IEEE, 1991.

\bibitem{schmidhuber1996simple}
Juergen Schmidhuber, Jieyu Zhao, and MA~Wiering.
\newblock Simple principles of metalearning.
\newblock {\em Technical report IDSIA}, 69:1--23, 1996.

\bibitem{thrun1998learning}
Sebastian Thrun and Lorien Pratt.
\newblock Learning to learn: {I}ntroduction and overview.
\newblock In {\em Learning to learn}, pages 3--17. Springer, 1998.

\bibitem{hochreiter2001learning}
Sepp Hochreiter, A~Steven Younger, and Peter~R Conwell.
\newblock Learning to learn using gradient descent.
\newblock In {\em International Conference on Artificial Neural Networks},
  pages 87--94. Springer, 2001.

\bibitem{santoro2016meta}
Adam Santoro, Sergey Bartunov, Matthew Botvinick, Daan Wierstra, and Timothy
  Lillicrap.
\newblock Meta-learning with memory-augmented neural networks.
\newblock In {\em International Conference on Machine Learning}, pages
  1842--1850, 2016.

\bibitem{wang2016learning}
Jane~X Wang, Zeb Kurth-Nelson, Dhruva Tirumala, Hubert Soyer, Joel~Z Leibo,
  Remi Munos, Charles Blundell, Dharshan Kumaran, and Matt Botvinick.
\newblock Learning to reinforcement learn.
\newblock {\em arXiv preprint arXiv:1611.05763}, 2016.

\bibitem{bengio2019ideas}
Eliza Strickland.
\newblock Yoshua {B}engio, revered architect of {AI}, has some ideas about what
  to build next.
\newblock {\em IEEE Spectrum}, December 2019.

\bibitem{braun2009motor}
Daniel~A Braun, Ad~Aertsen, Daniel~M Wolpert, and Carsten Mehring.
\newblock Motor task variation induces structural learning.
\newblock {\em Current Biology}, 19(4):352--357, 2009.

\bibitem{braun2010structure}
Daniel~A Braun, Carsten Mehring, and Daniel~M Wolpert.
\newblock Structure learning in action.
\newblock {\em Behavioural Brain Research}, 206(2):157--165, 2010.

\bibitem{ortega2019meta}
Pedro~A Ortega, Jane~X Wang, Mark Rowland, Tim Genewein, Zeb Kurth-Nelson,
  Razvan Pascanu, Nicolas Heess, Joel Veness, Alex Pritzel, Pablo Sprechmann,
  et~al.
\newblock Meta-learning of sequential strategies.
\newblock {\em arXiv preprint arXiv:1905.03030}, 2019.

\bibitem{duff2002optimal}
Michael~O'Gordon Duff and Andrew Barto.
\newblock {\em Optimal {L}earning: {C}omputational procedures for
  {B}ayes-adaptive {M}arkov decision processes}.
\newblock PhD thesis, University of Massachusetts at Amherst, 2002.

\bibitem{maheswaranathan2019reverse}
Niru Maheswaranathan, Alex Williams, Matthew Golub, Surya Ganguli, and David
  Sussillo.
\newblock Reverse engineering recurrent networks for sentiment classification
  reveals line attractor dynamics.
\newblock In {\em Advances in Neural Information Processing Systems}, pages
  15670--15679, 2019.

\bibitem{maheswaranathan2020recurrent}
Niru Maheswaranathan and David Sussillo.
\newblock How recurrent networks implement contextual processing in sentiment
  analysis.
\newblock {\em arXiv preprint arXiv:2004.08013}, 2020.

\bibitem{tanaka2019deep}
Hidenori Tanaka, Aran Nayebi, Niru Maheswaranathan, Lane McIntosh, Stephen
  Baccus, and Surya Ganguli.
\newblock From deep learning to mechanistic understanding in neuroscience: the
  structure of retinal prediction.
\newblock In {\em Advances in Neural Information Processing Systems}, pages
  8535--8545, 2019.

\bibitem{bau2018identifying}
Anthony Bau, Yonatan Belinkov, Hassan Sajjad, Nadir Durrani, Fahim Dalvi, and
  James Glass.
\newblock Identifying and controlling important neurons in neural machine
  translation.
\newblock In {\em International Conference on Learning Representations}, 2019.

\bibitem{olah2018the}
Chris Olah, Arvind Satyanarayan, Ian Johnson, Shan Carter, Ludwig Schubert,
  Katherine Ye, and Alexander Mordvintsev.
\newblock The building blocks of interpretability.
\newblock {\em Distill}, 2018.

\bibitem{olah2020an}
Chris Olah, Nick Cammarata, Ludwig Schubert, Gabriel Goh, Michael Petrov, and
  Shan Carter.
\newblock An overview of early vision in {InceptionV1}.
\newblock {\em Distill}, 2020.

\bibitem{jaderberg2019human}
Max Jaderberg, Wojciech~M Czarnecki, Iain Dunning, Luke Marris, Guy Lever,
  Antonio~Garcia Castaneda, Charles Beattie, Neil~C Rabinowitz, Ari~S Morcos,
  Avraham Ruderman, et~al.
\newblock Human-level performance in {3D} multiplayer games with
  population-based reinforcement learning.
\newblock {\em Science}, 364(6443):859--865, 2019.

\bibitem{marr2010vision}
David Marr.
\newblock {\em Vision: A computational investigation into the human
  representation and processing of visual information}.
\newblock MIT Press, 2010.

\bibitem{kolen2001field}
John~F Kolen and Stefan~C Kremer.
\newblock {\em A field guide to dynamical recurrent networks}.
\newblock John Wiley \& Sons, 2001.

\bibitem{gers1999learning}
Felix~A Gers, J{\"u}rgen Schmidhuber, and Fred Cummins.
\newblock Learning to forget: {C}ontinual prediction with {LSTM}.
\newblock 1999.

\bibitem{cho2014learning}
Kyunghyun Cho, Bart Van~Merri{\"e}nboer, Caglar Gulcehre, Dzmitry Bahdanau,
  Fethi Bougares, Holger Schwenk, and Yoshua Bengio.
\newblock Learning phrase representations using {RNN} encoder-decoder for
  statistical machine translation.
\newblock {\em arXiv preprint arXiv:1406.1078}, 2014.

\bibitem{robinson1987utility}
AJ~Robinson and Frank Fallside.
\newblock {\em The utility driven dynamic error propagation network}.
\newblock University of Cambridge Department of Engineering Cambridge, 1987.

\bibitem{werbos1988generalization}
Paul~J Werbos.
\newblock Generalization of backpropagation with application to a recurrent gas
  market model.
\newblock {\em Neural networks}, 1(4):339--356, 1988.

\bibitem{mealy1955method}
George~H Mealy.
\newblock A method for synthesizing sequential circuits.
\newblock {\em The Bell System Technical Journal}, 34(5):1045--1079, 1955.

\bibitem{sipser1996introduction}
Michael Sipser.
\newblock Introduction to the theory of computation.
\newblock {\em ACM Sigact News}, 27(1):27--29, 1996.

\bibitem{savage1998models}
JE~Savage.
\newblock Models of computation. exploring the power of computing.
\newblock {\em Reading, MA}, 1998.

\bibitem{clarke2018model}
Edmund~M Clarke~Jr, Orna Grumberg, Daniel Kroening, Doron Peled, and Helmut
  Veith.
\newblock {\em Model checking}.
\newblock MIT press, 2018.

\bibitem{Baier2008PrinciplesOM}
Christel Baier and Joost-Pieter Katoen.
\newblock Principles of model checking.
\newblock 2008.

\bibitem{BAETEN2014399}
Jos~C.M. Baeten and Davide Sangiorgi.
\newblock Concurrency theory: {A} historical perspective on coinduction and
  process calculi.
\newblock In Jörg~H. Siekmann, editor, {\em Computational Logic}, volume~9 of
  {\em Handbook of the History of Logic}, pages 399 -- 442. North-Holland,
  2014.

\bibitem{raiffa1961applied}
Howard Raiffa and Robert Schlaifer.
\newblock Applied statistical decision theory.
\newblock 1961.

\bibitem{bishop2006pattern}
Christopher~M Bishop.
\newblock {\em Pattern recognition and machine learning}.
\newblock springer, 2006.

\bibitem{gelman2013bayesian}
Andrew Gelman, John~B Carlin, Hal~S Stern, David~B Dunson, Aki Vehtari, and
  Donald~B Rubin.
\newblock {\em Bayesian data analysis}.
\newblock CRC press, 2013.

\bibitem{lattimore2018bandit}
Tor Lattimore and Csaba Szepesv{\'a}ri.
\newblock Bandit algorithms.
\newblock {\em preprint}, page~28, 2018.

\bibitem{gittins1979bandit}
John~C Gittins.
\newblock Bandit processes and dynamic allocation indices.
\newblock {\em Journal of the Royal Statistical Society: Series B
  (Methodological)}, 41(2):148--164, 1979.

\bibitem{edwards2016towards}
Harrison Edwards and Amos Storkey.
\newblock Towards a neural statistician.
\newblock {\em arXiv preprint arXiv:1606.02185}, 2016.

\bibitem{sutton2018reinforcement}
Richard~S Sutton and Andrew~G Barto.
\newblock {\em Reinforcement learning: {A}n introduction}.
\newblock MIT Press, 2018.

\bibitem{kingma2014adam}
Diederik~P Kingma and Jimmy Ba.
\newblock Adam: A method for stochastic optimization.
\newblock {\em arXiv preprint arXiv:1412.6980}, 2014.

\bibitem{espeholt2018impala}
Lasse Espeholt, Hubert Soyer, Remi Munos, Karen Simonyan, Volodymir Mnih, Tom
  Ward, Yotam Doron, Vlad Firoiu, Tim Harley, Iain Dunning, et~al.
\newblock Impala: {S}calable distributed deep-{RL} with importance weighted
  actor-learner architectures.
\newblock {\em arXiv preprint arXiv:1802.01561}, 2018.

\bibitem{borg2005modern}
Ingwer Borg and Patrick~JF Groenen.
\newblock {\em Modern multidimensional scaling: {T}heory and applications}.
\newblock Springer Science \& Business Media, 2005.

\bibitem{girard2005approximate}
Antoine Girard and George~J Pappas.
\newblock Approximate bisimulations for nonlinear dynamical systems.
\newblock In {\em Proceedings of the 44th IEEE Conference on Decision and
  Control}, pages 684--689. IEEE, 2005.

\bibitem{rabinowitz2019meta}
Neil~C Rabinowitz.
\newblock Meta-learners' learning dynamics are unlike learners'.
\newblock {\em arXiv preprint arXiv:1905.01320}, 2019.

\bibitem{maass2004computational}
Wolfgang Maass and Henry Markram.
\newblock On the computational power of circuits of spiking neurons.
\newblock {\em Journal of Computer and System Sciences}, 69(4):593--616, 2004.

\bibitem{lim2019recurrent}
Bryan Lim, Stefan Zohren, and Stephen Roberts.
\newblock Recurrent neural filters: Learning independent bayesian filtering
  steps for time series prediction.
\newblock {\em arXiv preprint arXiv:1901.08096}, 2019.

\bibitem{krishnan2017structured}
Rahul~G Krishnan, Uri Shalit, and David Sontag.
\newblock Structured inference networks for nonlinear state space models.
\newblock In {\em Proceedings of the Thirty-First AAAI Conference on Artificial
  Intelligence}, pages 2101--2109, 2017.

\bibitem{coskun2017long}
Huseyin Coskun, Felix Achilles, Robert DiPietro, Nassir Navab, and Federico
  Tombari.
\newblock Long short-term memory kalman filters: Recurrent neural estimators
  for pose regularization.
\newblock In {\em Proceedings of the IEEE International Conference on Computer
  Vision}, pages 5524--5532, 2017.

\bibitem{munkhdalai2017meta}
Tsendsuren Munkhdalai and Hong Yu.
\newblock Meta networks.
\newblock {\em Proceedings of Machine Learning Research}, 70:2554, 2017.

\bibitem{munkhdalai2019metalearned}
Tsendsuren Munkhdalai, Alessandro Sordoni, Tong Wang, and Adam Trischler.
\newblock Metalearned neural memory.
\newblock In {\em Advances in Neural Information Processing Systems}, pages
  13331--13342, 2019.

\bibitem{finn2017model}
Chelsea Finn, Pieter Abbeel, and Sergey Levine.
\newblock Model-agnostic meta-learning for fast adaptation of deep networks.
\newblock In {\em Proceedings of the 34th International Conference on Machine
  Learning-Volume 70}, pages 1126--1135. JMLR. org, 2017.

\bibitem{grant2018recasting}
Erin Grant, Chelsea Finn, Sergey Levine, Trevor Darrell, and Thomas Griffiths.
\newblock Recasting gradient-based meta-learning as hierarchical bayes.
\newblock In {\em International Conference on Learning Representations}, 2018.

\bibitem{Raghu2020Rapid}
Aniruddh Raghu, Maithra Raghu, Samy Bengio, and Oriol Vinyals.
\newblock Rapid learning or feature reuse? {T}owards understanding the
  effectiveness of {MAML}.
\newblock In {\em International Conference on Learning Representations}, 2020.

\bibitem{shalizi2001computational}
Cosma~Rohilla Shalizi and James~P Crutchfield.
\newblock Computational mechanics: Pattern and prediction, structure and
  simplicity.
\newblock {\em Journal of Statistical Physics}, 104(3-4):817--879, 2001.

\bibitem{zhang2019learning}
Amy Zhang, Zachary~C Lipton, Luis Pineda, Kamyar Azizzadenesheli, Anima
  Anandkumar, Laurent Itti, Joelle Pineau, and Tommaso Furlanello.
\newblock Learning causal state representations of partially observable
  environments.
\newblock {\em arXiv preprint arXiv:1906.10437}, 2019.

\bibitem{kaelbling1990action}
Leslie~Pack Kaelbling and Stanley~J Rosenschein.
\newblock Action and planning in embedded agents.
\newblock {\em Robotics and Autonomous Systems}, 6(1-2):35--48, 1990.

\bibitem{rosenschein1995situated}
Stanley~J Rosenschein and Leslie~Pack Kaelbling.
\newblock A situated view of representation and control.
\newblock {\em Artificial Intelligence}, 73(1-2):149--173, 1995.

\bibitem{belinkov2019analysis}
Yonatan Belinkov and James Glass.
\newblock Analysis methods in neural language processing: A survey.
\newblock {\em Transactions of the Association for Computational Linguistics},
  7:49--72, 2019.

\bibitem{sohn2019bayesian}
Hansem Sohn, Devika Narain, Nicolas Meirhaeghe, and Mehrdad Jazayeri.
\newblock Bayesian computation through cortical latent dynamics.
\newblock {\em Neuron}, 103(5):934--947, 2019.

\bibitem{maheswaranathan2019universality}
Niru Maheswaranathan, Alex Williams, Matthew Golub, Surya Ganguli, and David
  Sussillo.
\newblock Universality and individuality in neural dynamics across large
  populations of recurrent networks.
\newblock In {\em Advances in Neural Information Processing Systems}, pages
  15603--15615, 2019.

\bibitem{Dasgupta644534}
Ishita Dasgupta, Eric Schulz, Joshua~B. Tenenbaum, and Samuel~J. Gershman.
\newblock A theory of learning to infer.
\newblock {\em bioRxiv}, 2019.

\bibitem{zahavy2016graying}
Tom Zahavy, Nir Ben-Zrihem, and Shie Mannor.
\newblock Graying the black box: Understanding {DQNs}.
\newblock In {\em International Conference on Machine Learning}, pages
  1899--1908, 2016.

\bibitem{koul2018learning}
Anurag Koul, Sam Greydanus, and Alan Fern.
\newblock Learning finite state representations of recurrent policy networks.
\newblock {\em arXiv preprint arXiv:1811.12530}, 2018.

\bibitem{erhan2009visualizing}
Dumitru Erhan, Yoshua Bengio, Aaron Courville, and Pascal Vincent.
\newblock Visualizing higher-layer features of a deep network.
\newblock 2009.

\bibitem{simonyan2013deep}
Karen Simonyan, Andrea Vedaldi, and Andrew Zisserman.
\newblock Deep inside convolutional networks: Visualising image classification
  models and saliency maps.
\newblock {\em arXiv preprint arXiv:1312.6034}, 2013.

\bibitem{zeiler2014visualizing}
Matthew~D Zeiler and Rob Fergus.
\newblock Visualizing and understanding convolutional networks.
\newblock In {\em European Conference on Computer Vision}, pages 818--833.
  Springer, 2014.

\bibitem{shrikumar2016not}
Avanti Shrikumar, Peyton Greenside, Anna Shcherbina, and Anshul Kundaje.
\newblock Not just a black box: Learning important features through propagating
  activation differences.
\newblock {\em arXiv preprint arXiv:1605.01713}, 2016.

\bibitem{selvaraju2016grad}
Ramprasaath~R Selvaraju, Abhishek Das, Ramakrishna Vedantam, Michael Cogswell,
  Devi Parikh, and Dhruv Batra.
\newblock Grad-cam: Why did you say that?
\newblock {\em arXiv preprint arXiv:1611.07450}, 2016.

\bibitem{smilkov2017smoothgrad}
Daniel Smilkov, Nikhil Thorat, Been Kim, Fernanda Vi{\'e}gas, and Martin
  Wattenberg.
\newblock Smoothgrad: removing noise by adding noise.
\newblock {\em arXiv preprint arXiv:1706.03825}, 2017.

\bibitem{bau2020understanding}
David Bau, Jun-Yan Zhu, Hendrik Strobelt, Agata Lapedriza, Bolei Zhou, and
  Antonio Torralba.
\newblock Understanding the role of individual units in a deep neural network.
\newblock {\em Proceedings of the National Academy of Sciences}, 2020.

\bibitem{ribeiro2016should}
Marco~Tulio Ribeiro, Sameer Singh, and Carlos Guestrin.
\newblock " why should i trust you?" explaining the predictions of any
  classifier.
\newblock In {\em Proceedings of the 22nd ACM SIGKDD International Conference
  on Knowledge Discovery and Data Mining}, pages 1135--1144, 2016.

\end{thebibliography}



\newpage
\appendix
\clearpage
\setcounter{page}{1} 

\section*{Supplementary Material}

\section{Task Details}\label{sec:task-details}

There is a total of $14$~tasks, out of which~$10$ are prediction and~$4$ are bandit tasks. 

\paragraph{Prediction:} The prediction tasks can be grouped according to their observational distributions:
\begin{itemize}
    \item \emph{Bernoulli:} The agent observes samples $x_t$ drawn from a Bernoulli distribution $\mathrm{Ber}(\theta)$. The prior distribution over the bias $\theta$ is given by a Beta distribution $\mathrm{Beta}(\alpha, \beta)$, where $\alpha>0$ and $\beta>0$ are the hyperparameters. We have three tasks with three respective prior distributions: $\mathrm{Beta}(1, 1)$, $\mathrm{Beta}(0.5, 0.5)$, and $\mathrm{Beta}(1, 5)$.
    \item \emph{Categorical:} The agent observes samples $x_t$ drawn from a categorical distribution $\mathrm{Cat}(\vec{\theta})$ where $\vec{\theta}=[\theta_1, \theta_2, \theta_3]^T$. The prior distribution over the bias parameters $\vec{\theta}$ is given by a Dirichlet distribution $\mathrm{Dirichlet}(\vec{\alpha})$, where $\vec{\alpha}=[\alpha_1, \alpha_2, \alpha_3]^T$ are the concentration parameters. We have three categorical tasks with three respective prior distributions: $\mathrm{Dirichlet}(1, 1, 1)$, $\mathrm{Dirichlet}(1, 1, 0.1)$, and $\mathrm{Dirichlet}(0.5, 0.5, 0.5)$.
    \item \emph{Exponential:} The agent observes samples $x_t$ drawn from an exponential distribution $\mathrm{Exp}(\lambda)$ where~$\lambda>0$ is the rate parameter. The prior distribution over the rate parameter $\lambda$ is given by a Gamma distribution $\mathrm{Gamma}(\alpha, \beta)$, where $\alpha>0$ is the shape and $\beta>0$ is the rate. We use two exponential prediction tasks: their priors are $\mathrm{Gamma}(1, 0.5)$ and $\mathrm{Gamma}(5, 1)$.
    \item \emph{Gaussian:} The agent observes samples $x_t$ drawn from a Gaussian distribution $\mathrm{Normal}(\mu, 1/\tau)$, where $\mu$ is an unknown mean and $\tau$ is a known precision. The prior distribution over $\mu$ is given by a Gaussian distribution $\mathrm{Normal}(m, 1/p)$, where~$m$ and~$p$ are the prior mean and precision parameters. We have two Gaussian prediction tasks: their priors are $\mathrm{Normal}(0, 1)$ and $\mathrm{Normal}(1, 1)$ and their precisions $\tau=1$ and $\tau=5$ respectively.
\end{itemize}

A prediction task proceeds as follows. As a concrete example, consider the Bernoulli prediction case---other distributions proceed analogously. In the very beginning of each episode, the bias parameter~$\theta$ is drawn from a fixed prior distribution $p(\theta) = \mathrm{Beta}(1, 1)$. This parameter is never shown to the agent. Then, in each turn~$t=1, 2, \ldots, T=20$, the agent makes a probabilistic prediction $\pi_t$ and then receives an observation~$x_t \sim p(x|\theta) = \mathrm{Ber}(\theta)$ drawn from the observational distribution. This leads to a prediction loss given by $-\log(\pi_t(x_t))$, where $\pi_t(x_t)$ is the predicted probability of the observation~$x_t$ at time~$t$. Then the next round starts.

\paragraph{Bandits:} As in the prediction case, the two-armed bandit tasks can also be grouped according to their reward distributions:
\begin{itemize}
    \item \emph{Bernoulli:} Upon pulling a lever $a \in \{1, 2\}$, the agent observes a reward sampled from a Bernoulli distribution $\mathrm{Ber}(\theta_a)$, where $\theta_a$ is the bias of arm~$a$. The prior distribution over each arm bias is given by a Beta distribution as in the prediction case. We have two Bernoulli bandit tasks: the first draws both biases from $\mathrm{Beta}(1, 1)$, and the second from $\mathrm{Beta}(2, 1)$ and $\mathrm{Beta}(1, 2)$ respectively.
    \item \emph{Gaussian:} Upon pulling a lever $a \in \{1, 2\}$, the agent observes a reward sampled from a Gaussian distribution $\mathrm{Normal}(\mu, \tau)$, where $\mu$ and $\tau$ are the unknown mean and the known precision of arm~$a$ respectively. As in the prediction case, the prior distribution over each arm mean is given by a Normal distribution. We have two Gaussian bandit tasks: the first with precision $\tau=1$ and prior $\mathrm{Normal}(0, 1)$ for both arms; and the second with precision $\tau=1$ and prior $\mathrm{Normal}(0, 0.1)$.
\end{itemize}

The interaction protocol for bandit tasks is as follows. For concreteness we pick the first Bernoulli bandit---but other bandits proceed analogously. In the very beginning of each episode, the arm biases~$\theta_1$ and~$\theta_2$ are drawn from a fixed prior distribution $p(\theta) = \mathrm{Beta}(1, 1)$. These parameters are never shown to the agent. Then, in each turn~$t=1, 2, \ldots, T$, the agent pulls a lever $a \sim \pi_t$ from its policy at time~$t$ and receives a reward~$r_t \sim p(r|\theta_a) = \mathrm{Ber}(\theta_a)$ drawn from the reward distribution. Then the next round starts. The agent's return is the discounted sum of rewards $\sum_t \gamma^t r_t$ with discount factor $\gamma=0.95$.

\section{Agent Details} \label{sec:agent-details}
\subsection{Bayes-optimal agents} \label{sec:opt-details}
Our Bayes-optimal agents act and predict according to the standard models in the literature. We briefly summarize this below.

\begin{table}[t]
    \centering
    \caption{Prediction rules for Bayes-optimal agents}\label{tab:task-updates}
    \begin{tabular}{cccc}
        \toprule
        Observation 
            & Prior
            & Update
            & Posterior Predictive \\
        \midrule
        Bernoulli$(\theta)$ 
            & Beta$(\alpha, \beta)$
            & $\alpha \leftarrow \alpha + x$; 
              $\beta \leftarrow \beta + (1-x)$
            & Bernoulli$(\frac{\alpha}{\alpha+\beta})$ \\
        Categorical$(\vec{\theta})$ 
            & Dirichlet$(\alpha_1, \alpha_2, \alpha_3)$ 
            & $\alpha_x \leftarrow \alpha_x + 1$
            & Categorical$(\frac{\alpha_i}{\sum_j \alpha_j})$\\
        Normal$(\mu, 1/\tau)$ 
            & Normal$(m, 1/p)$
            & $m \leftarrow \tfrac{pm + \tau x}{p + \tau}$;
              $p \leftarrow p + \tau$
            & Normal$(m, \frac{1}{p}+\frac{1}{\tau})$\\
        Exponential$(\lambda)$ 
            & Gamma$(\alpha, \beta)$
            & $\alpha \leftarrow \alpha + 1$, 
              $\beta \leftarrow \beta + x$
            & Lomax$(\alpha, \beta)$\\
        \bottomrule
    \end{tabular}
\end{table}

\paragraph{Prediction:} A Bayes-optimal agent makes predictions by combining a prior with observed data to form a posterior belief. Consider a Bernoulli environment that generates observations according to~$\mathrm{Bernoulli}(\theta)$, where in each episode~$\theta \sim \mathrm{Beta}(1, 1)$. In each turn~$t$, the agent makes a prediction according to the \emph{posterior predictive distribution}
\begin{equation}\label{eq:posterior-predictive}
    p(x_t|x_{<t}) = \int p(x_t|\theta) p(\theta|x_{<t}) d\theta,
\end{equation}
where the prior~$p(\theta|x_{<t})$ is the posterior of the previous turn (in the first step the agent uses its prior, which, for the optimal agent, coincides with the environment's prior).
Subsequently, the agent receives an observation~$x_t$, which and updates its posterior belief:
\begin{equation}\label{eq:bayesian-update}
    p(\theta|x_{\leq t}) \propto p(\theta|x_{<t}) p(x_t|\theta).
\end{equation}
Note that for the distributions used in our prediction tasks, the posterior can be parameterized by a small set of values: the minimal sufficient statistics (which compress the whole observation history~$x_{<t}$ into the minimal amount of information required to perform optimally).

For a Bernoulli predictor, the posterior predictive~\eqref{eq:posterior-predictive} is equal to
\[
    p(x_t|x_{<t}) 
    = p(x_t|\alpha,\beta) 
    = \mathrm{Ber}(\tfrac{\alpha}{\alpha + \beta}).
\]
where, $\alpha$ and $\beta$ are the sufficient statistics.
The posterior belief is given by
\[
    p(\theta|x_{\leq t}) 
    = p(\theta|\alpha',\beta') 
    = \mathrm{Beta}(\alpha', \beta'),
\]
where $\alpha' = \alpha + x_t$ and $\beta' = \beta + (1-x_t)$ are
the hyperparameters updated by the observation~$x_t$. For a full list of update and prediction rules, see Table~\ref{tab:task-updates}.

\paragraph{Bandits:}
A Bayes-optimal bandit player maintains beliefs for each arm's distribution over the rewards. For instance, if the rewards are distributed according to a Bernoulli law, then the agent keeps track of one~$(\alpha, \beta)$ sufficient-statistic pair per arm. The optimal arm to pull next is then given by
\begin{equation}\label{eq:optimal-action}
  a^\ast
  = \arg\max_a Q(a|\alpha_1, \beta_1, \alpha_1, \beta_1),
\end{equation}
where the Q-value is recursively defined as
\begin{equation}
\begin{aligned}\label{eq:q-value}
  Q(a|\alpha_1, \beta_1, \alpha_1, \beta_1) 
    &:= 0 
    && \text{if }t=T\\
  Q(a|\alpha_1, \beta_1, \alpha_1, \beta_1)
    &:= \sum_{r} p(r|\alpha_a, \beta_a) \Bigl\{ r 
    + \max_{a'} \gamma Q(a'|\alpha'_1, \beta'_1, \alpha'_2, \beta'_2)
    \Bigr\}
    && \text{if }t<T
\end{aligned}
\end{equation}
and where $\alpha'_1, \beta'_1, \alpha'_2, \beta'_2$ are the hyperparameters for the next step, updated in accordance to the action taken and the reward observed. Computing~\eqref{eq:optimal-action} naively is computationally intractable. Instead, one can pre-compute \emph{Gittins indices} in polynomial time, and use them as a replacement for the Q-values in~\eqref{eq:optimal-action} \citep{gittins1979bandit, lattimore2018bandit, edwards2016towards}. In particular, we have used the methods presented in \citep{edwards2016towards} to compute Gittins indices for the Bernoulli- and Gaussian-distributed rewards.

\subsection{RNN agents} \label{sec:rnn-details}

\paragraph{Prediction:}
We trained agents on the prediction tasks (episode length~$T=20$ steps) using supervised learning with a batch size of $128$ using BPTT unroll of $20$~timesteps, and a total training duration of $1e7$~steps. We used the Adam optimizer with learning rate $10^{-4}$, parameters $\beta_1=0.9$, $\beta_2=0.999$, and gradients clipped at magnitude 1. Networks were initialised with weights drawn from a truncated normal with standard deviation $1/\sqrt{N_{\rm in}}$, where $N_{\rm in}$ is the size of the input layer. We use the following output-parametrization: Bernoulli-predictions - single output corresponding to the log-probability (of observing ``heads''); Categorical predictions - $3$-D outputs corresponding to prediction logits; Normal predictions - $2$ linear outputs, one for mean and one for log-precision; Exponential predictions - $2$ linear outputs, one for $\log\alpha$ and one for $\log\beta$.

\paragraph{Bandit:}
We trained the reinforcement learners on bandit tasks (episode length~$T=20$ steps) with the Impala algorithm~\cite{espeholt2018impala} using a batch size of $16$ and discount factor $\gamma=0.95$ for a total number of $1e8$ training steps. The BPTT unroll length was 5 timesteps, and the learning rate was $2.5 \times 10^{-5}$. We used an entropy penalty of $0.003$ and value baseline loss scaling of $0.48$; i.e.,, the training objective was $\mathcal{L}_{\rm{VTrace}} + 0.003 \mathcal{L}_{\rm{Entropy}} + 0.48\mathcal{L}_{\rm{Value}}$. We used the same initialisation scheme as for the prediction tasks. RNN outputs in all bandit tasks were  $2$-dimensional action logits (one for each arm). Bandit agents are trained to minimize empirical (``sampled'') cumulative discounted rewards. For our behavioral and output dissimilarity measures we report expected reward instead of sampled reward (using the environment's ground-truth parameters to which the agent does not have access to)---this reduces the impact of sampling noise on our estimates. 

\section{Structural Comparison Details}\label{sec:structural-comparison-details}
We implement the map~$\phi$ from RNN agent states~$\mathcal{S}_N$ to optimal agent states~$\mathcal{S}_M$ using an MLP with three hidden layers, each of size~$64$ (prediction tasks where the RNN state is $64$-dimensional) or $256$ (bandit tasks where the RNN state is $512$-dimensional), with ReLu activations. We first project the high-dimensional RNN agent state space down to a lower-dimensional representation using PCA. The number of principal components is set to match the dimension of the minimal sufficient statistics required by the task. We trained the MLP using the Adam optimiser with learning rate~$0.001$, $\beta_1=0.9$, $\beta_2=0.999$ and batch size~$200$. The training set consisted of data from $500$ roll-outs---all results we report were evaluated on $500$ held out test-trajectories. 

State dissimilarity~$D_s$ is measured by providing the same inputs to both agents (same observations in prediction tasks, and action-reward pairs from a reference trajectory\footnote{The reference trajectory is always generated from the fully trained RNN agent---also when analyzing RNN agents during training.} in bandit tasks), and then taking the mean-squared error between the (PCA-projected) original states and the mapped states (compare Figure~\ref{fig:structure} in the main paper). Output dissimilarity is computed by comparing the output produced by the original agent with the output produced after projecting the original agent state into the ``surrogate'' agent and evaluating the output. Note that the last step requires inverting the PCA projection in order to create a ``valid'' state in the surrogate agent. For the optimal agent the PCA is invertible since its dimensionality is the same as the agent's state (i.e., the PCA on the optimal agent simply performs a rotation and whitening). On the RNN agent, we use the following scheme: we construct an invertible PCA projection as well, which requires having the same number of components as the internal state's dimensionality. Then, to implant a state from the Bayes-optimal agent the first $n$ components are set according to the mapping~$\phi$, all other principal components are set to their mean-value (across $500$~episodes).

\section{Additional Results}

\subsection{PCA for untrained meta-learner}\label{sec:pca_random_init}
\begin{figure}[htb]
    \centering
    \begin{subfigure}[t]{0.48\textwidth}
        \centering
        \includegraphics[width=\textwidth]{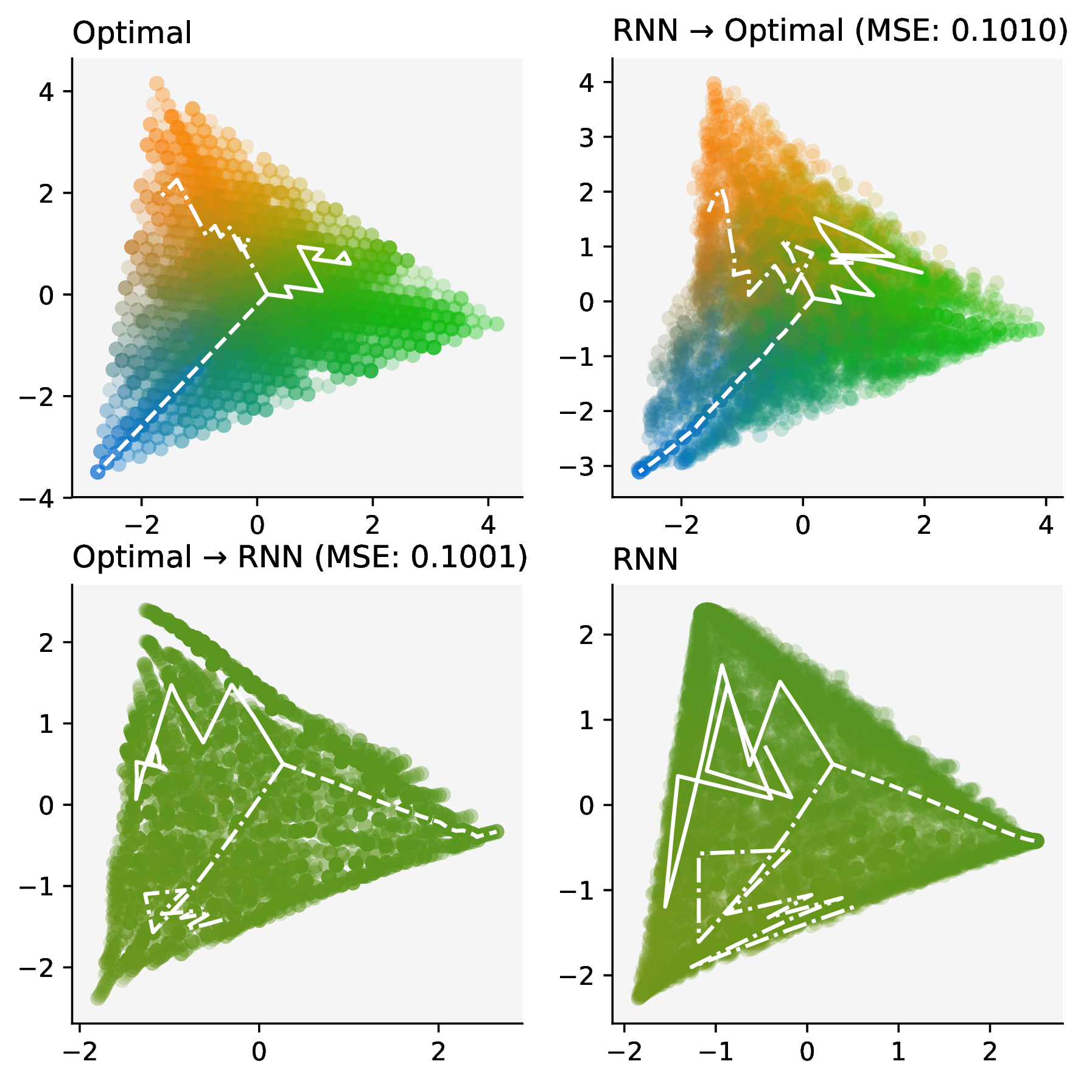}
        \caption{Categorical-variable prediction task $\mathrm{Dirichlet}(1,1,1)$. Colors indicate the output-probabilities (=posterior predictive dist.) for the corresponding state. Lines correspond to the three episodes shown in Figure~\ref{fig:behavior}. Dimensionality of $s_t^\mathrm{rnn}$ is $64$. MLP-regressor $\phi$ has three hidden layers with $64$ neurons each. }
        \label{fig:structure_categorical_init}
    \end{subfigure}
    \hfill
    \begin{subfigure}[t]{0.48\textwidth}
        \centering
        \includegraphics[width=\textwidth]{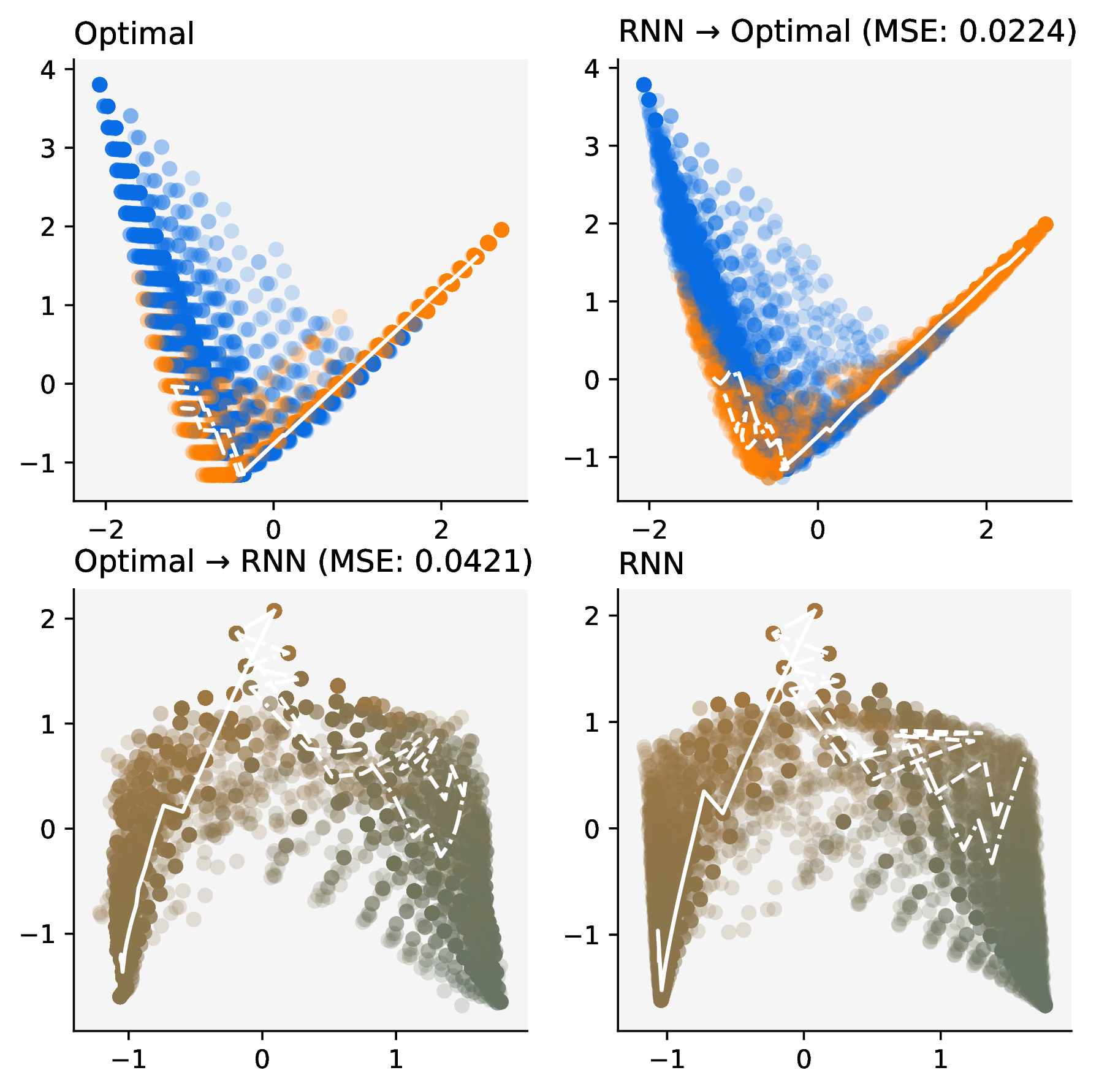}
        \caption{2-armed Bernoulli-bandit task $\sim\mathrm{Beta}(1,1)$. Colors indicate the output-probabilities (=action probabilities) for the corresponding state. Lines correspond to the three episodes shown in Figure~\ref{fig:behavior}. Dimensionality of $s_t^\mathrm{rnn}$ is $512$. MLP-regressor $\phi$ has three hidden layers with $256$ neurons each.}
        \label{fig:structure_gaussian_bandit_init}
    \end{subfigure}
    \caption{Structural comparison for \textbf{untrained agent} (compare Figure~\ref{fig:structure} in main paper). Each sub-figure shows: (i - top left) Projection of Bayes-optimal state onto first two principal components, (iv - bottom right) projection of RNN state onto first two principal components, (ii - top right) learned regression from (iv) to (i), (iii - bottom left) learned regression from (i) to (iv). Scores in panels (ii) and (iii) indicate the mean-squared-error (MSE) of the learned regression (map $\phi$ was trained on training data, plots and numerical results show evaluation on held-out test-data---$500$ data-points for training and test respectively).}
    \label{fig:structure_init}
\end{figure}

Figure~\ref{fig:structure_init} shows the principal component projection and approximate simulation (mapping the state of one agent onto the other and computing the resulting output) for meta-learner after random initialization, without any training. Results for the trained agent (at the end of the training run) are shown in Figure~\ref{fig:structure} in the main paper.

\subsection{Variance explained by PC projections}\label{sec:variance_explained}
Table~\ref{tab:variance-explained} shows the variance explained when projecting the RNN state onto the first $n$ principal components, which is the first step of our structural analysis ($n$ is the dimensionality of the tasks' minimal sufficient statistics, and is between $2$ and $4$ dimensions)---see Section~\ref{sec:structural-comparison}. Numbers indicate the variance explained by projecting $500$ trajectories of length $T=20$ onto first $n$ principal components. Large number indicate that most of the variance in the data is captured by the PCA projection, which is the case for us in all tasks.

\begin{table}[h]
\centering
\caption{Variance of RNN-state explained by PCA projection.}
\label{tab:variance-explained}
\begin{tabular}{llcc}
    \toprule
    & Task & at initialization & after training \\
    \midrule
    \multirow{10}{*}{\rotatebox{90}{Prediction tasks}} & $\text{Beta}(1, 1)$ & 0.98 & 0.94 \\
    & $\text{Beta}(0.5, 0.5)$ & 0.98 & 0.92 \\
    & $\text{Beta}(1, 5)$ & 0.98 &  0.96  \\
    & $\text{Dirichlet}(0.5, 0.5, 0.5)$ & 0.93 & 0.96 \\
    & $\text{Dirichlet}(1, 1, 1)$ & 0.93 & 0.95 \\
    & $\text{Dirichlet}(1, 1, 0.1)$ & 0.94 & 0.96 \\
    &  $\text{Gamma}(1, 0.5)$ & 0.95 & 0.97 \\
    & $\text{Gamma}(5, 1)$ & 0.97 & 0.96 \\
    & $\text{Normal}(0,1)$ & 0.95 & 0.88 \\
    & $\text{Normal}(1,1)$ & 0.97 & 0.94 \\
    \midrule
    \multirow{4}{*}{\rotatebox{90}{Bandits}} & $\text{Beta}(1,1)$ & 0.97 & 0.96 \\
    & $\text{Beta}(2,1),~\text{Beta}(1,2)$ & 0.98 & 0.97 \\
    & $\text{Normal}(0,1)$ & 0.94 & 0.92 \\
    & $\text{Normal}(0,0.1)$ & 0.95 & 0.90 \\
    \bottomrule
\end{tabular}
\end{table}

\subsection{Preliminary architecture sweeps} \label{sec:architecture-ablation}
The meta-learners in our main experiments are three-layer RNNs (a fully connected encoder, followed by a LSTM layer and a fully connected decoder). Each layer has the same width~$N$ which was selected by running preliminary architecture sweeps (on a subset of tasks), shown in Figure~\ref{sec:architecture-ablation}. Generally we found that smaller RNNs suffice to successfully train on the prediction tasks compared to the RNN tasks. For instance a layer-width of~$3$ would suffice in principle to perform well on the prediction tasks (not that the maximum dimensionality of the minimal sufficient statistics is also exactly~$3$). However, we found that the smallest networks also tend to require more iterations to converge, with more noisy convergence in general. We thus selected~$N=32$ for prediction tasks (leading to a $64$-dimensional RNN state, which is the concatenation of cell- and hidden-states) as a compromise between RNN-state dimensionality, runtime-complexity and iterations required for training to converge robustly (in our main experiments we train prediction agents for $1e7$~steps, and bandit agents for $1e8$~steps). Using similar trade-offs we chose~$N=256$ for bandit tasks (leading to a $512$-dimensional RNN state).
\begin{figure}[h]
    \centering
    \begin{subfigure}{0.48\textwidth}
        \centering
        \includegraphics[width=\textwidth]{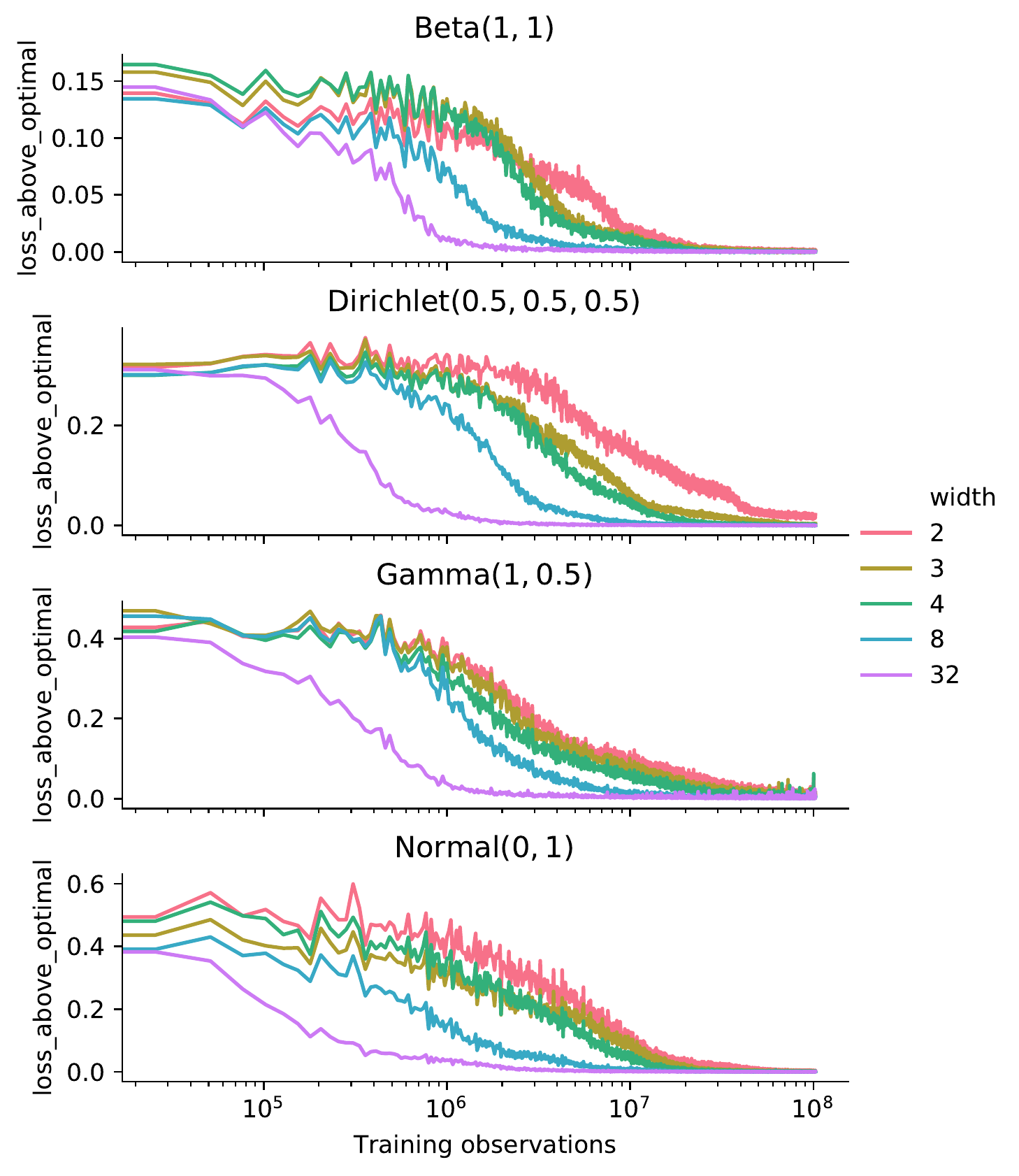}
        \caption{Subset of prediction tasks. Lines show difference between RNN and Bayes-optimal log-loss, averaged over $10$ training runs. }
    \end{subfigure}
    \hfill
    \begin{subfigure}{0.48\textwidth}
        \centering
        \includegraphics[width=\textwidth]{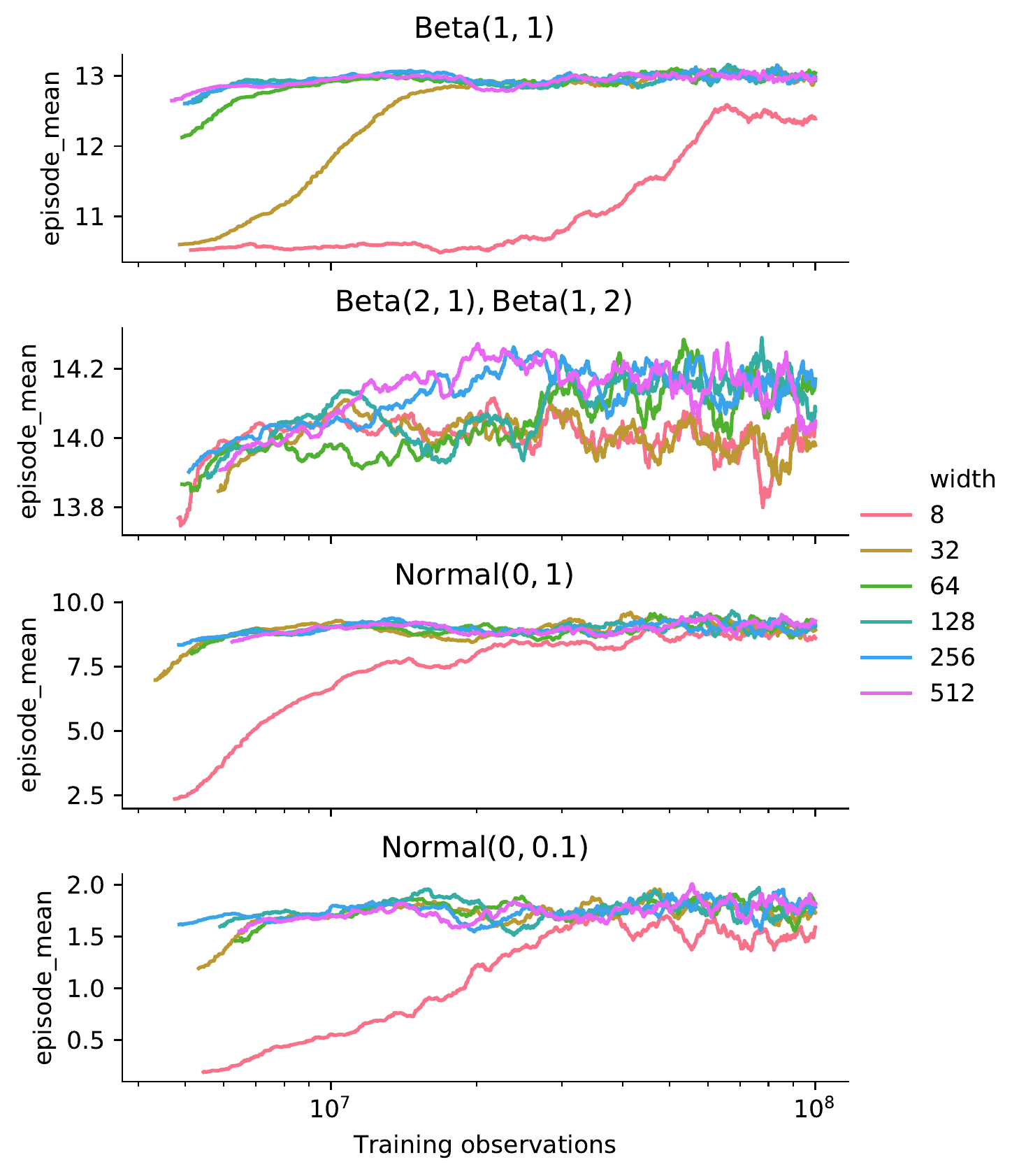}
        \caption{Bandit tasks. Lines show mean reward computed over the last $10$k steps (rolling average) for a single training run.  }
    \end{subfigure}
    \caption{Architecture sweeps.}
    \label{fig:architecture_ablation}
\end{figure}

\newpage

\begin{sidewaysfigure}
\subsection{Behavioral and structural comparison}
    \centering
    \includegraphics[width=1\textwidth]{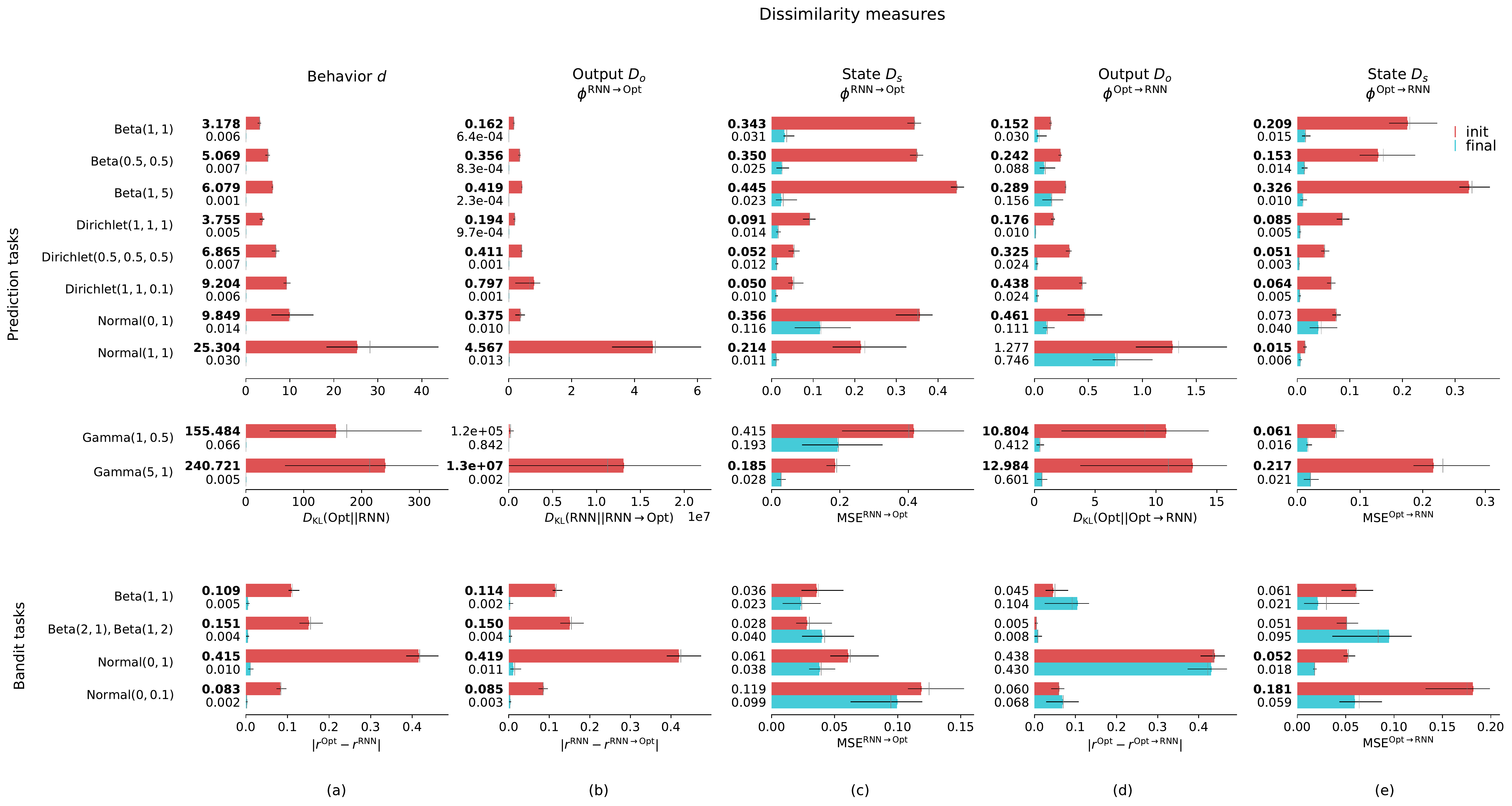}
    \caption{Behavioral and structural comparison for all tasks---same as Figure~\ref{fig:comparison_all_tasks} in main paper. Figure shows dissimilarity measures across $500$ episodes of length $T=20$, and $10$ different training runs of the meta-learner. `init' denotes the untrained meta-learner, `final' denotes evaluation at the end of training. Colored bars show median across training runs (also given as numerical values on y-axis), error bars denote $5$-$95$ quantiles (bold numbers indicate that upper end of `final' error bar is strictly lower than lower end of 'init' error bar), vertical grey ticks indicate mean values (across training runs). \textbf{(a)} Behavioral dissimilarity between meta-learned agent and Bayes-optimal agent (see Section~\ref{sec:behavioral_comparison}). \textbf{(b), (c)} State- and Output-dissimilarity for $\mathrm{RNN}\rightarrow\mathrm{Opt}$.  \textbf{(d), (e)} State- and Output-dissimilarity for $\mathrm{Opt}\rightarrow\mathrm{RNN}$.}
    \label{fig:comparison_all_tasks_annotated}
\end{sidewaysfigure}

\subsection{Structural comparison}\label{sec:structural-comparison-more}
\begin{figure}[htbp]
    \centering
    \begin{subfigure}{0.45\textwidth}
        \centering
        \includegraphics[width=\textwidth]{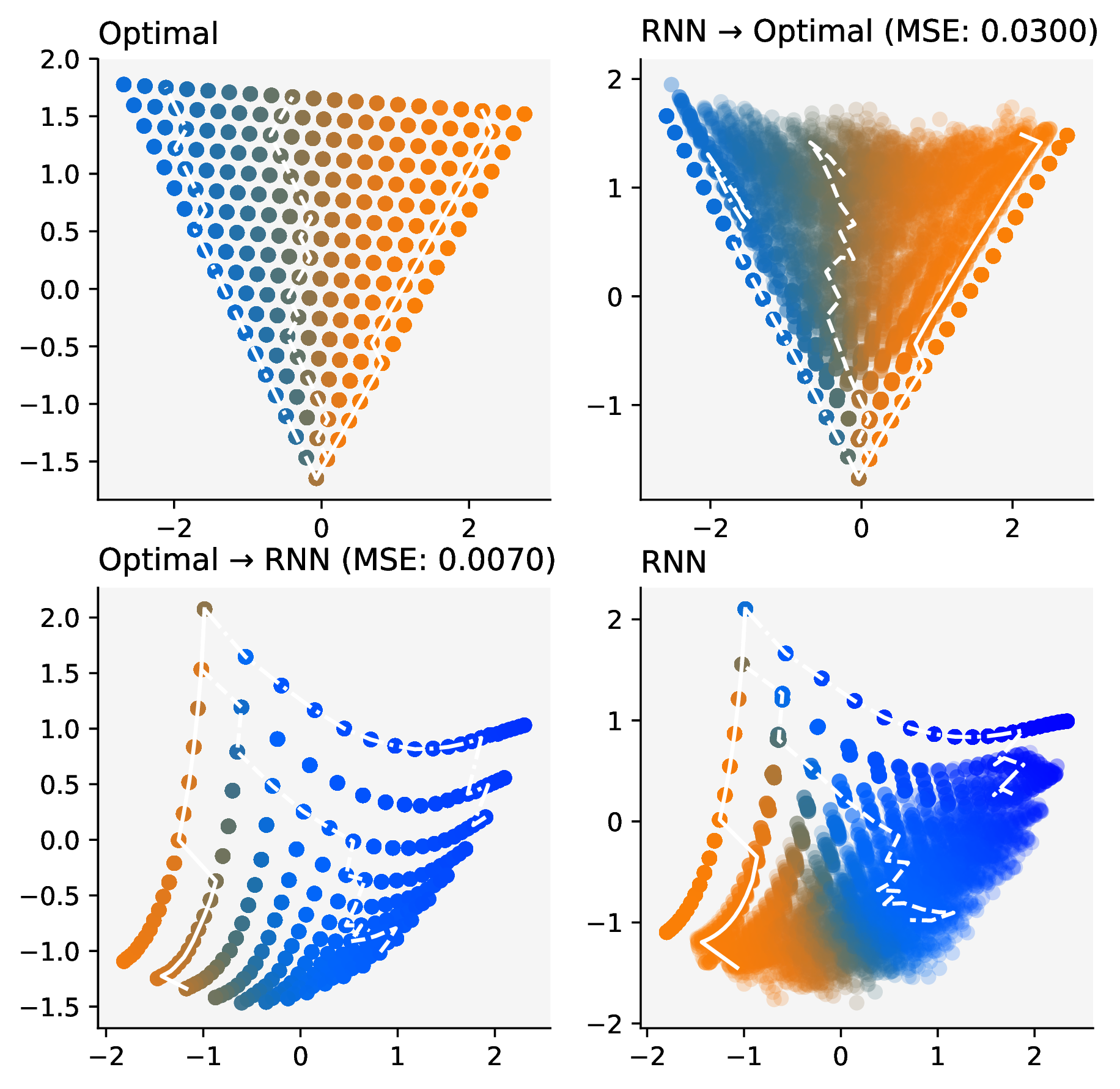}
        \caption{$\mathrm{Bernoulli}(\theta),~\theta \sim \mathrm{Beta}(1, 1)$}
    \end{subfigure}
    \hfill
    \begin{subfigure}{0.45\textwidth}
        \centering
        \includegraphics[width=\textwidth]{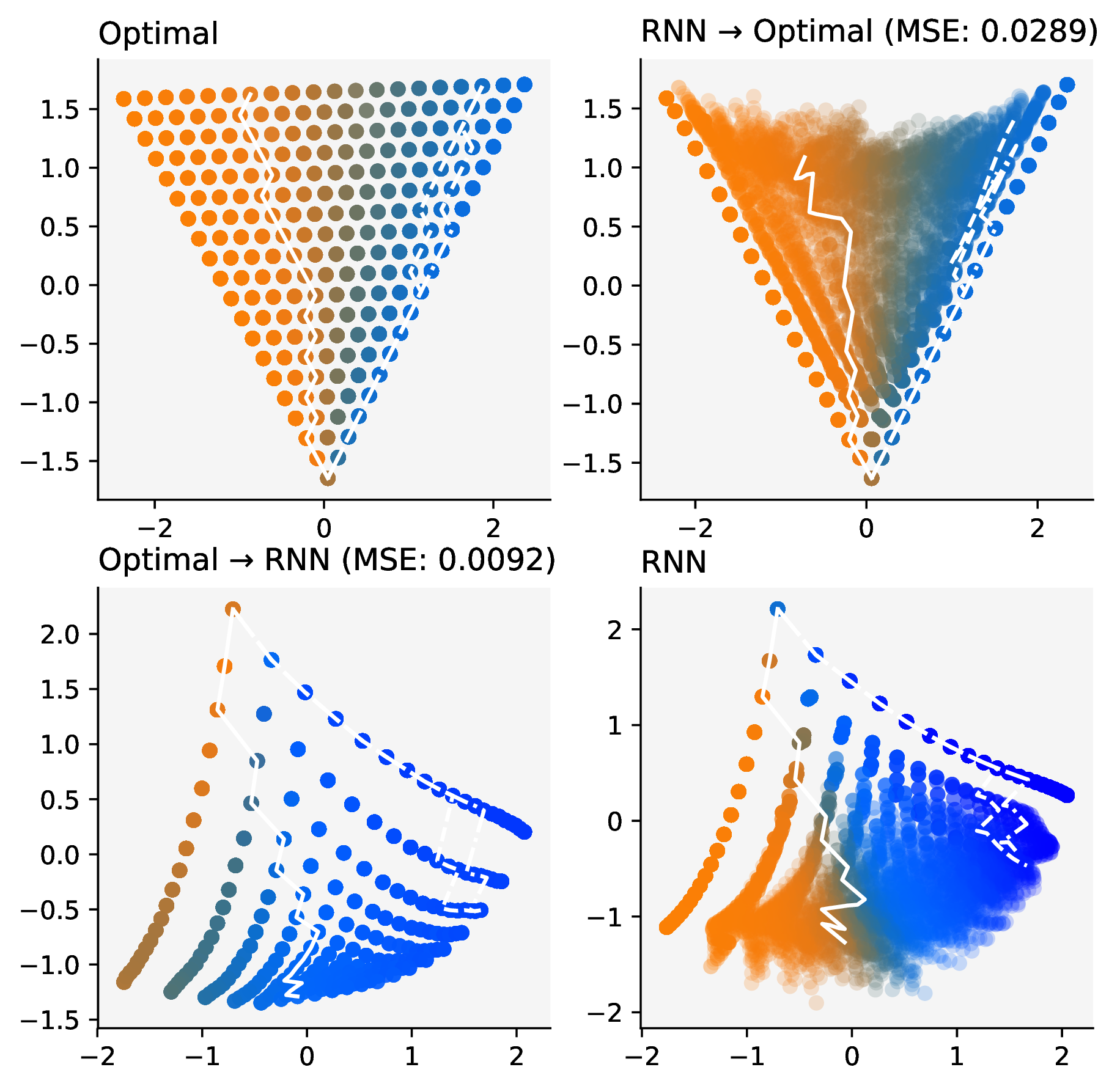}
        \caption{$\mathrm{Bernoulli}(\theta),~\theta \sim \mathrm{Beta}(0.5, 0.5)$}
    \end{subfigure}
    
    \begin{subfigure}{0.45\textwidth}
        \centering
        \includegraphics[width=\textwidth]{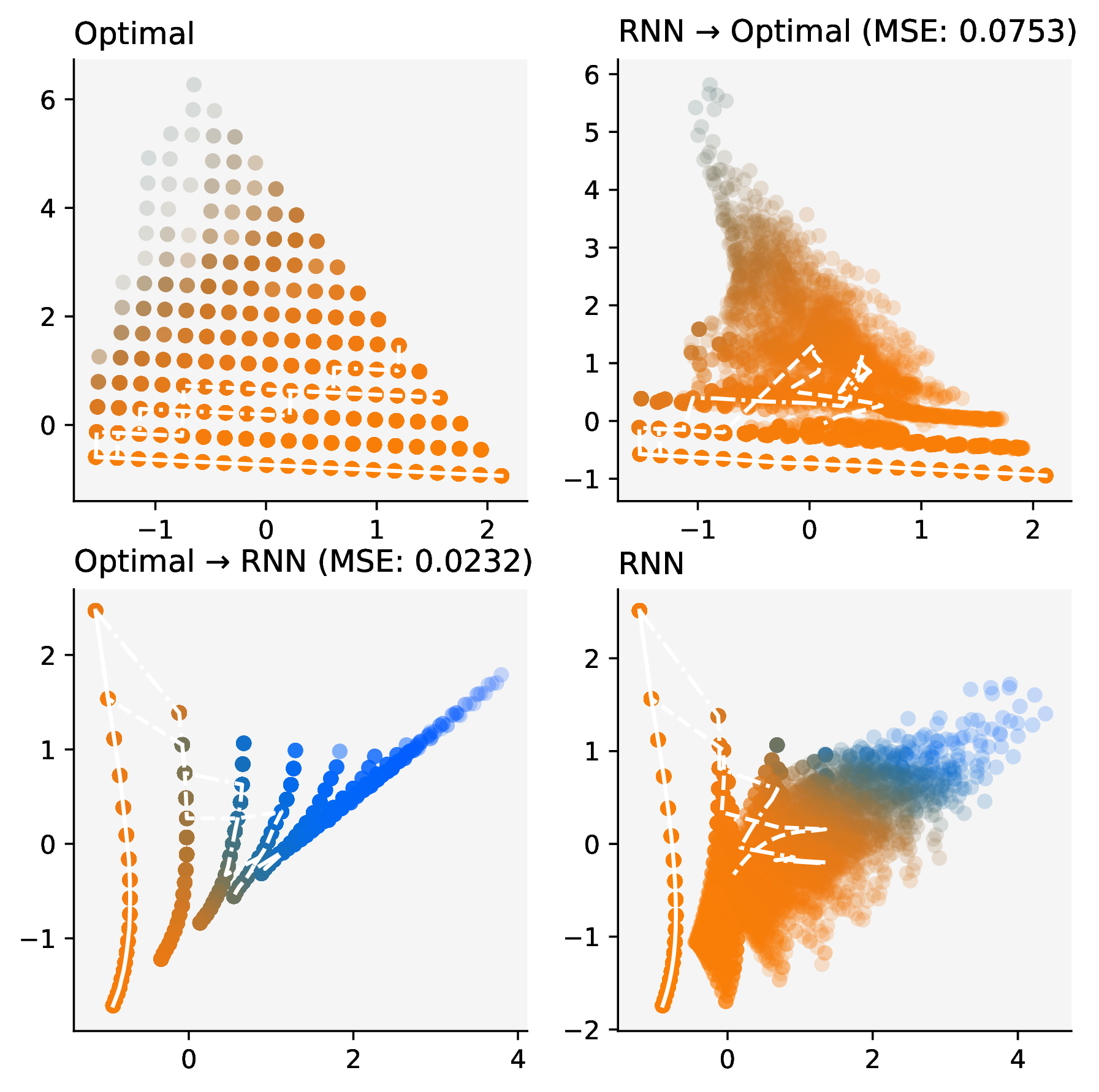}
        \caption{$\mathrm{Bernoulli}(\theta),~\theta \sim \mathrm{Beta}(1, 5)$}
    \end{subfigure}
    \hfill
    \begin{subfigure}{0.45\textwidth}
        \centering
        \includegraphics[width=\textwidth]{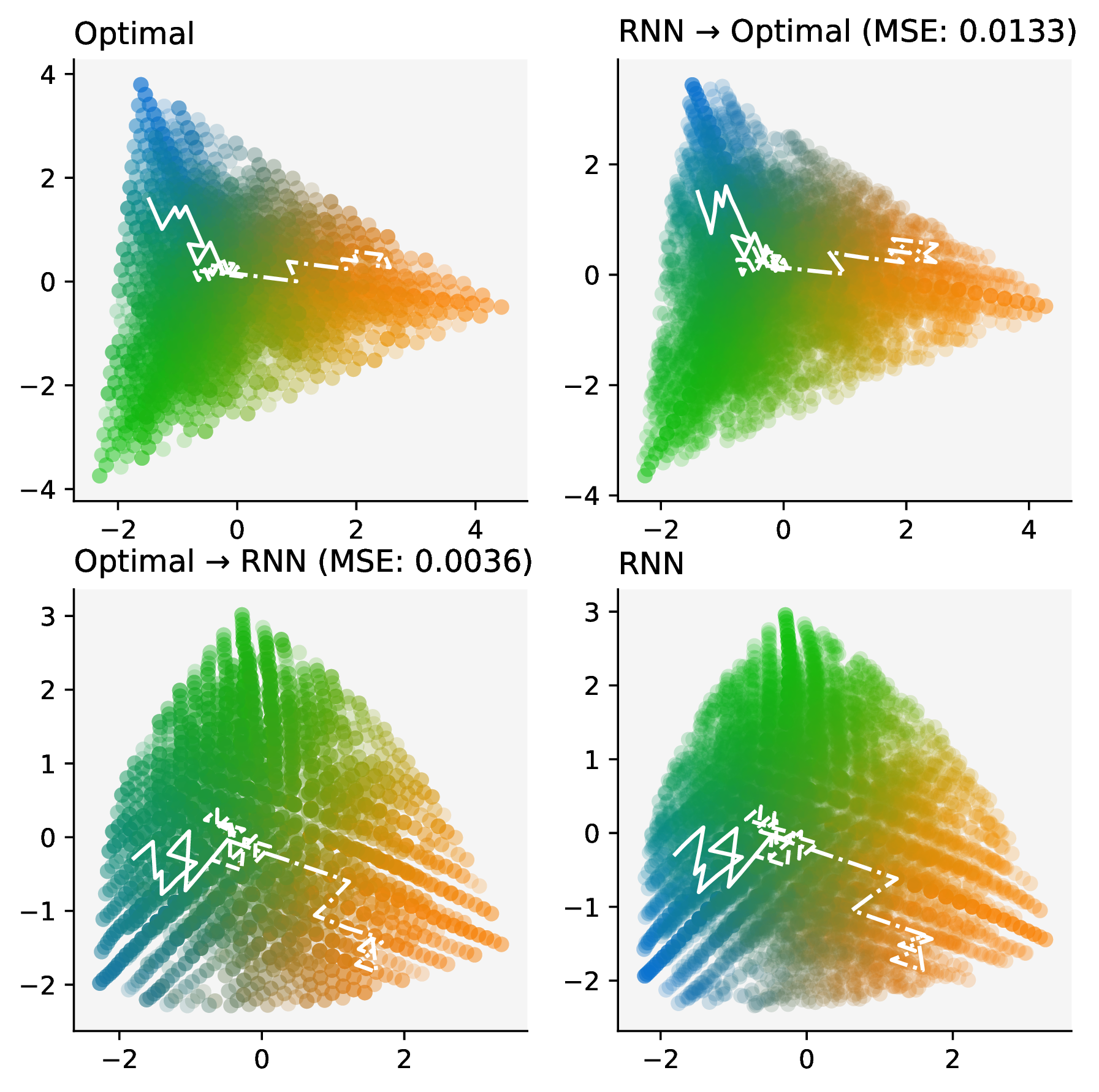}
        \caption{$\mathrm{Categorical}(\vec{\theta}),~\vec{\theta} \sim \mathrm{Dir}(1, 1, 1)$}
    \end{subfigure}
    
    \begin{subfigure}{0.45\textwidth}
        \centering
        \includegraphics[width=\textwidth]{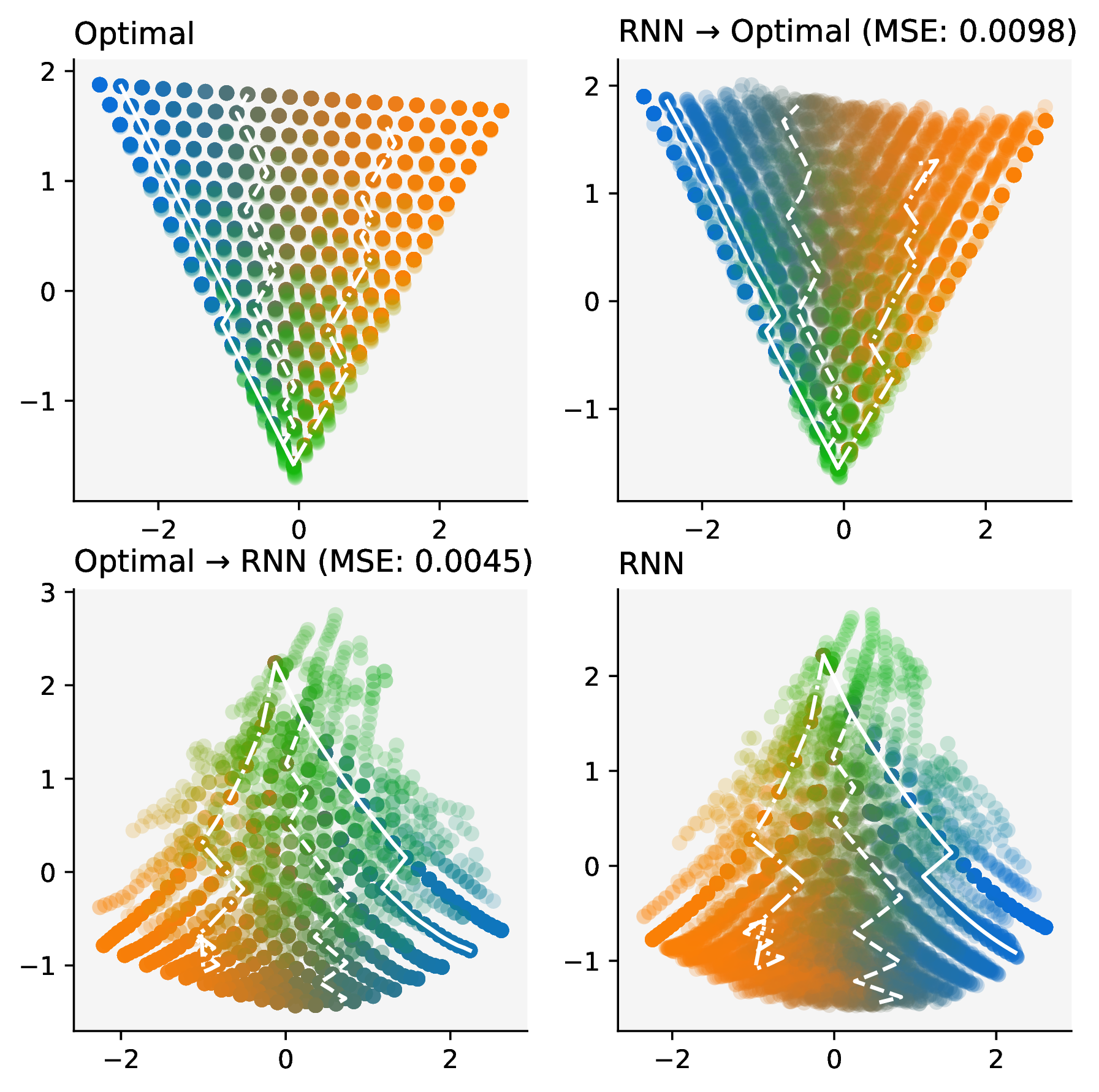}
        \caption{$\mathrm{Categorical}(\vec{\theta}),~\vec{\theta} \sim \mathrm{Dir}(1, 1, 0.1)$}
    \end{subfigure}
    \hfill
    \begin{subfigure}{0.45\textwidth}
        \centering
        \includegraphics[width=\textwidth]{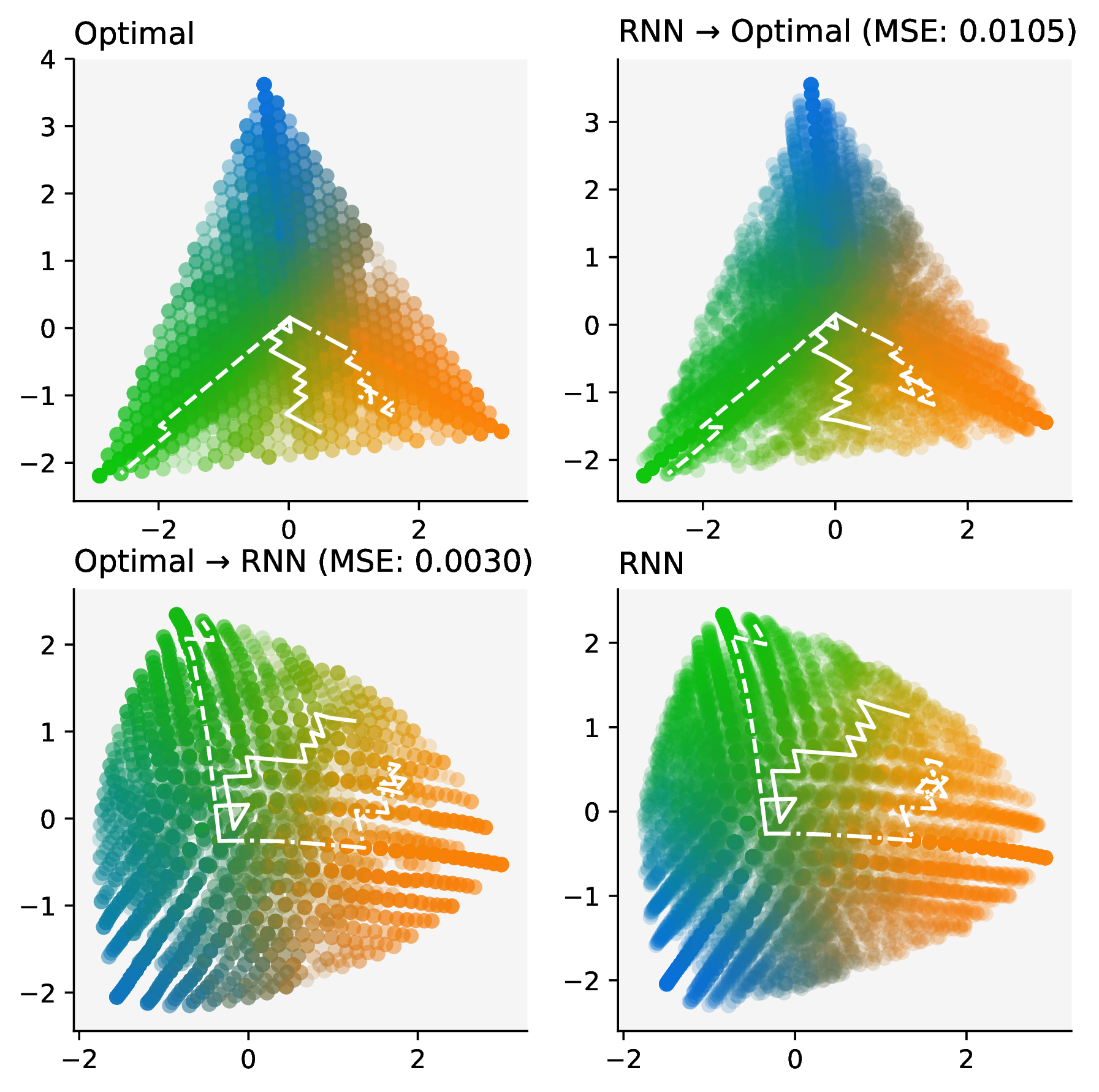}
        \caption{$\mathrm{Categorical}(\vec{\theta}),~\vec{\theta} \sim \mathrm{Dir}(0.5, 0.5, 0.5)$}
    \end{subfigure}
    \caption{Structural comparison I. Prediction probabilities are color-coded.}
    \label{fig:structure-i}
\end{figure}

\begin{figure}[htbp]
    \vspace{-10pt}
    \centering
    \begin{subfigure}{0.45\textwidth}
        \centering
        \includegraphics[width=\textwidth]{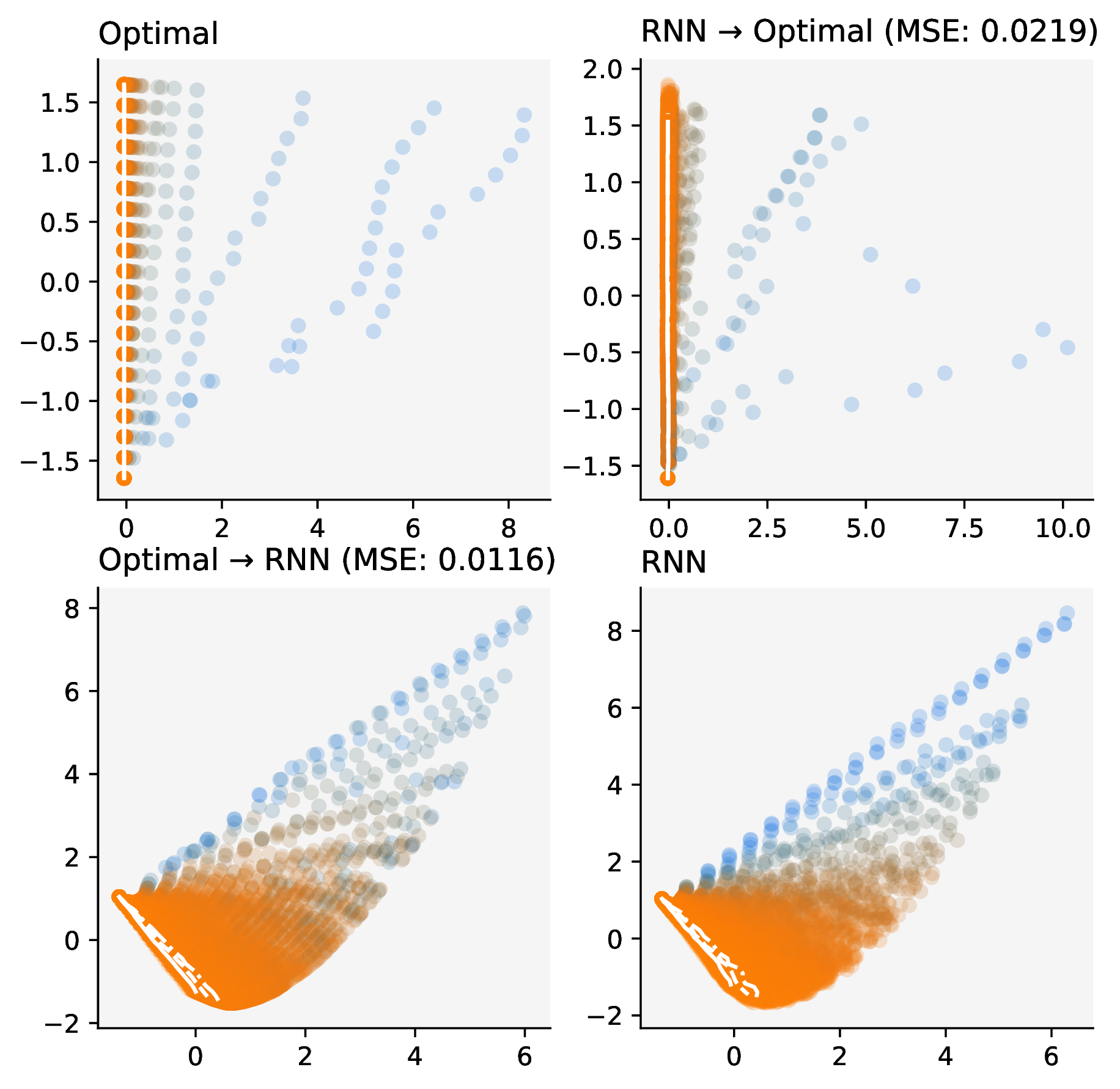}
        \caption{$\mathrm{Exponential}(\lambda),~\lambda \sim \mathrm{Gamma}(1, 0.5)$}
    \end{subfigure}
    \hfill
    \begin{subfigure}{0.45\textwidth}
        \centering
        \includegraphics[width=\textwidth]{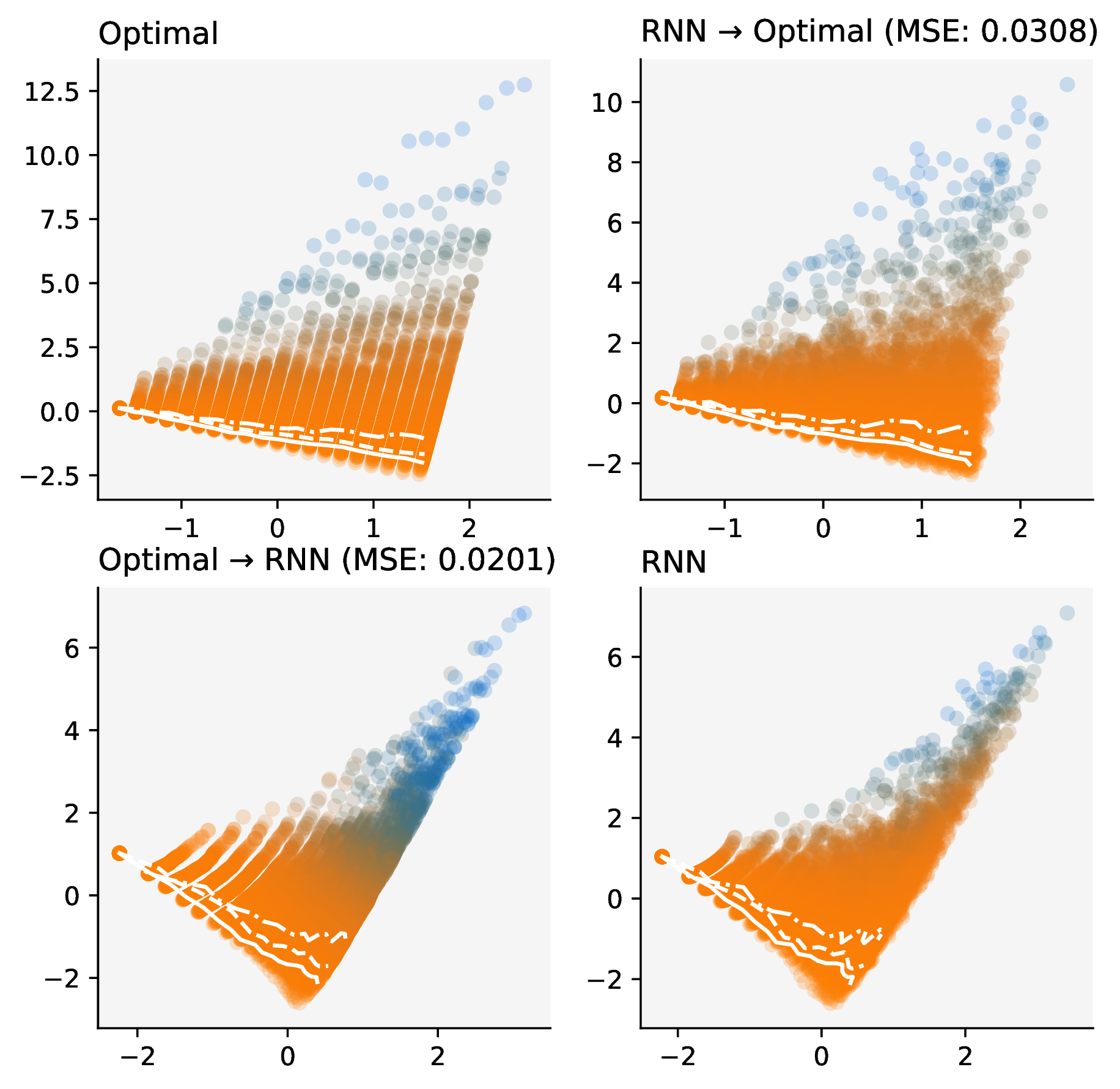}
        \caption{$\mathrm{Exponential}(\lambda),~\lambda \sim \mathrm{Gamma}(5, 1)$}
    \end{subfigure}
    
    \begin{subfigure}{0.45\textwidth}
        \centering
        \includegraphics[width=\textwidth]{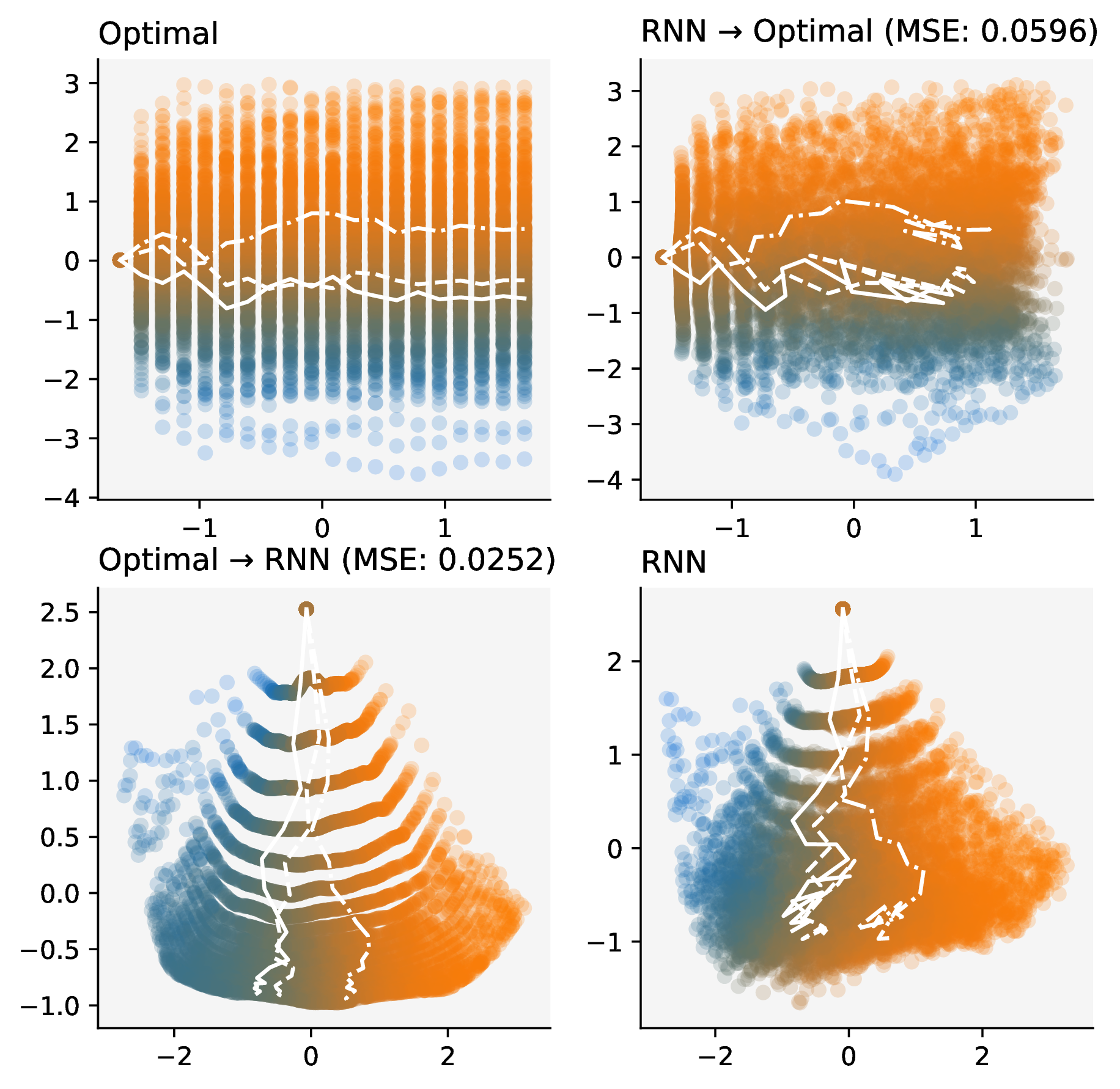}
        \caption{$\mathrm{Normal}(\mu, 1),~\mu \sim \mathrm{Normal}(0,1)$}
    \end{subfigure}
    \hfill
    \begin{subfigure}{0.45\textwidth}
        \centering
        \includegraphics[width=\textwidth]{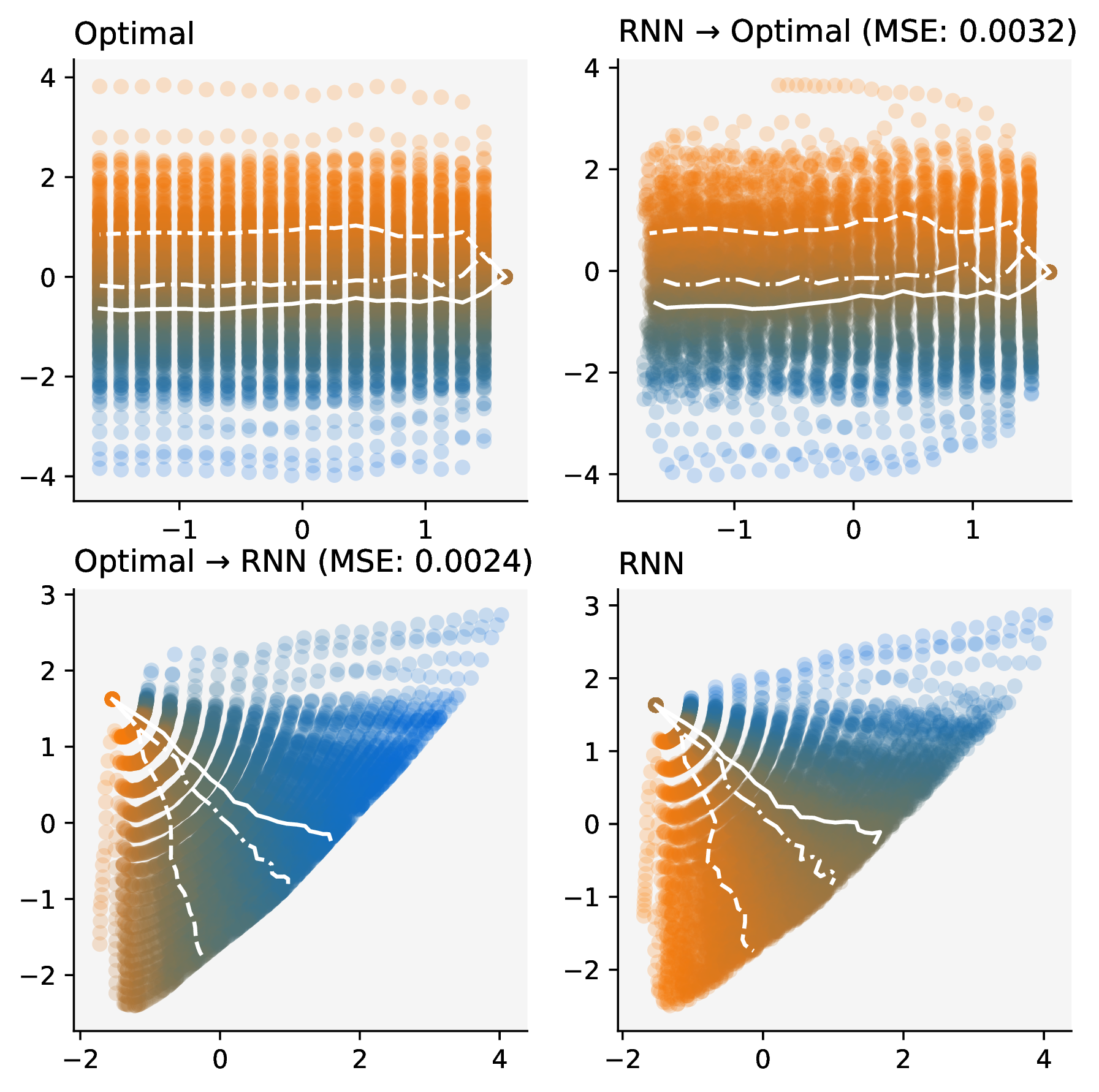}
        \caption{$\mathrm{Normal}(\mu, 0.2),~\mu \sim \mathrm{Normal}(1,1)$}
    \end{subfigure}
    \caption{Structural comparison II. The predicted means are color-coded.}
    \label{fig:structure-ii}
\end{figure}

\begin{figure}[htbp]
    \vspace{-10pt}
    \centering
    \begin{subfigure}{0.45\textwidth}
        \centering
        \includegraphics[width=\textwidth]{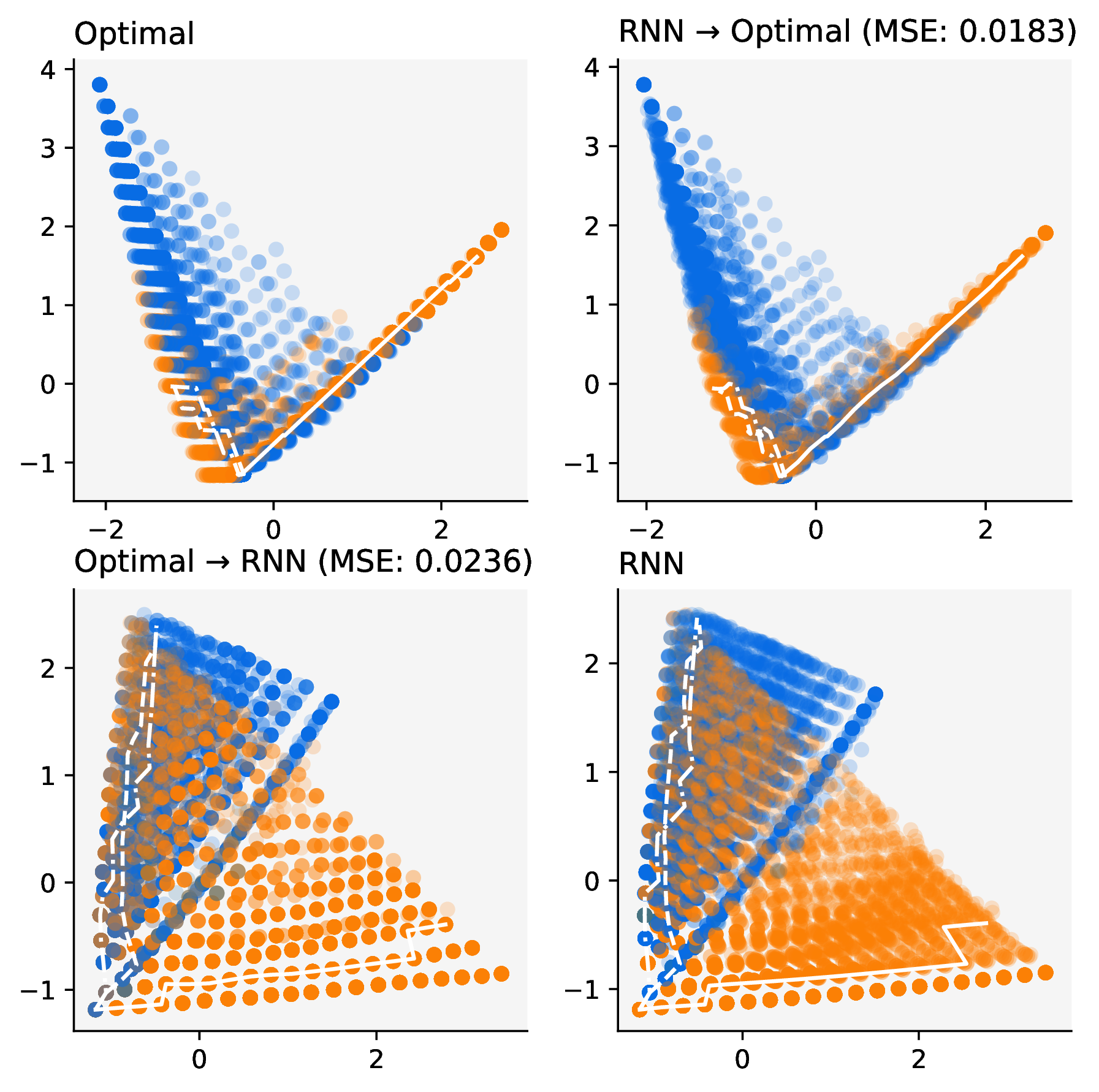}
        \caption{$\theta_1, \theta_2 \sim \mathrm{Beta}(1, 1)$}
    \end{subfigure}
    \hfill
    \begin{subfigure}{0.45\textwidth}
        \centering
        \includegraphics[width=\textwidth]{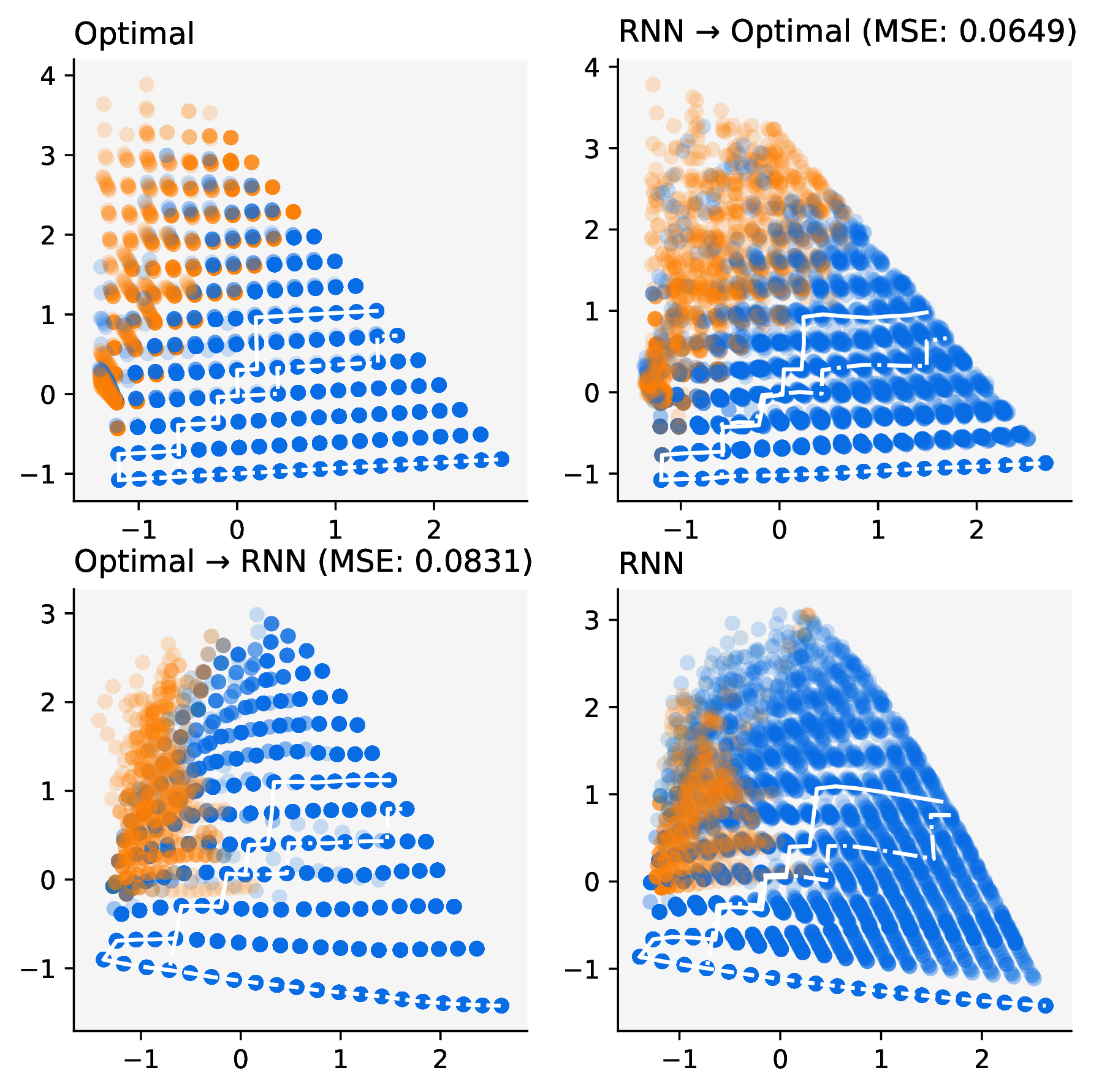}
        \caption{$\theta_1,  \sim \mathrm{Beta}(2, 1),~ \theta_2 \sim \mathrm{Beta}(1, 2)$}
    \end{subfigure}
    
    \begin{subfigure}{0.45\textwidth}
        \centering
        \includegraphics[width=\textwidth]{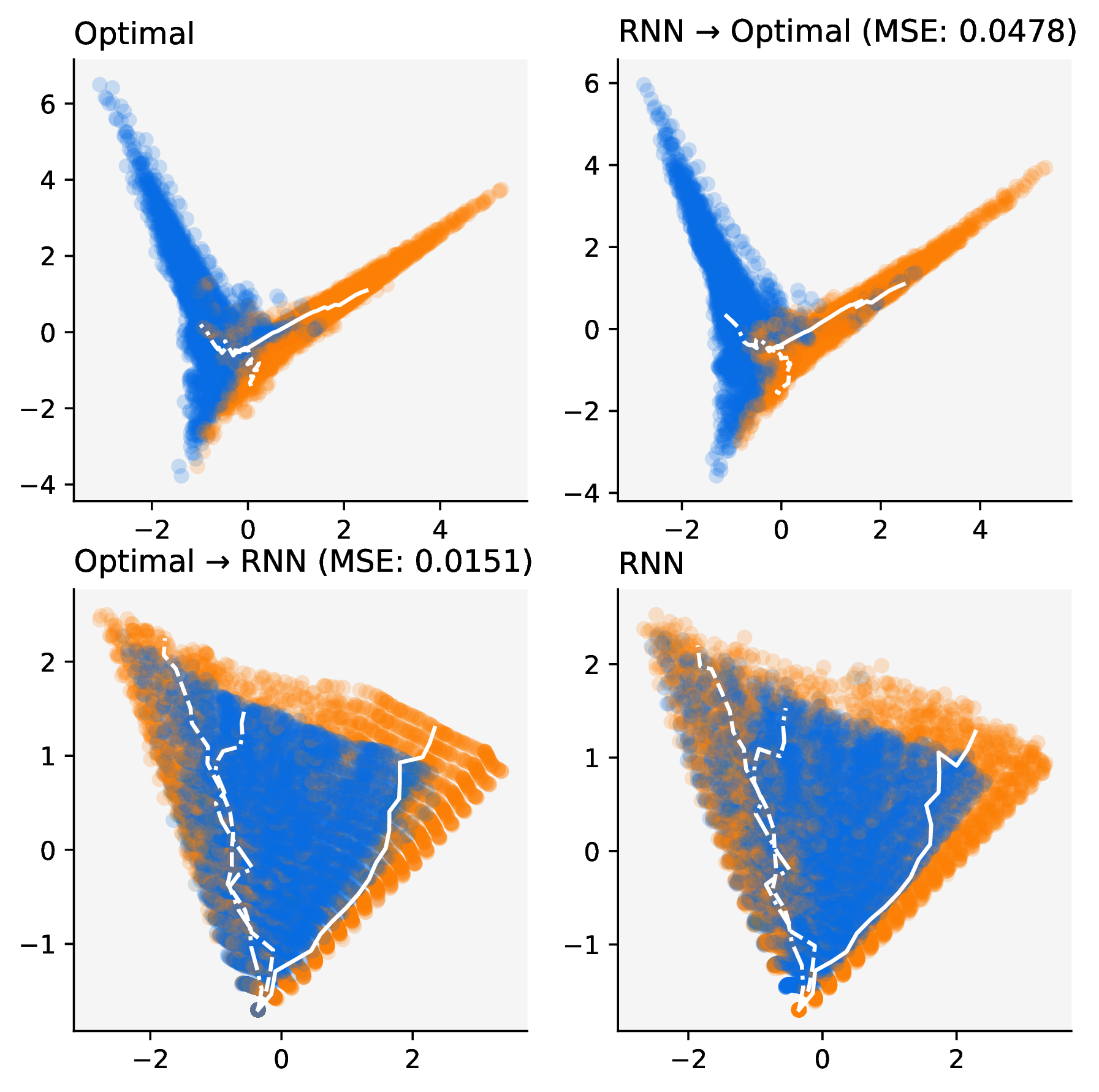}
        \caption{$\mu_1, \mu_2 \sim \mathrm{Normal}(0, 1)$}
    \end{subfigure}
    \hfill
    \begin{subfigure}{0.45\textwidth}
        \centering
        \includegraphics[width=\textwidth]{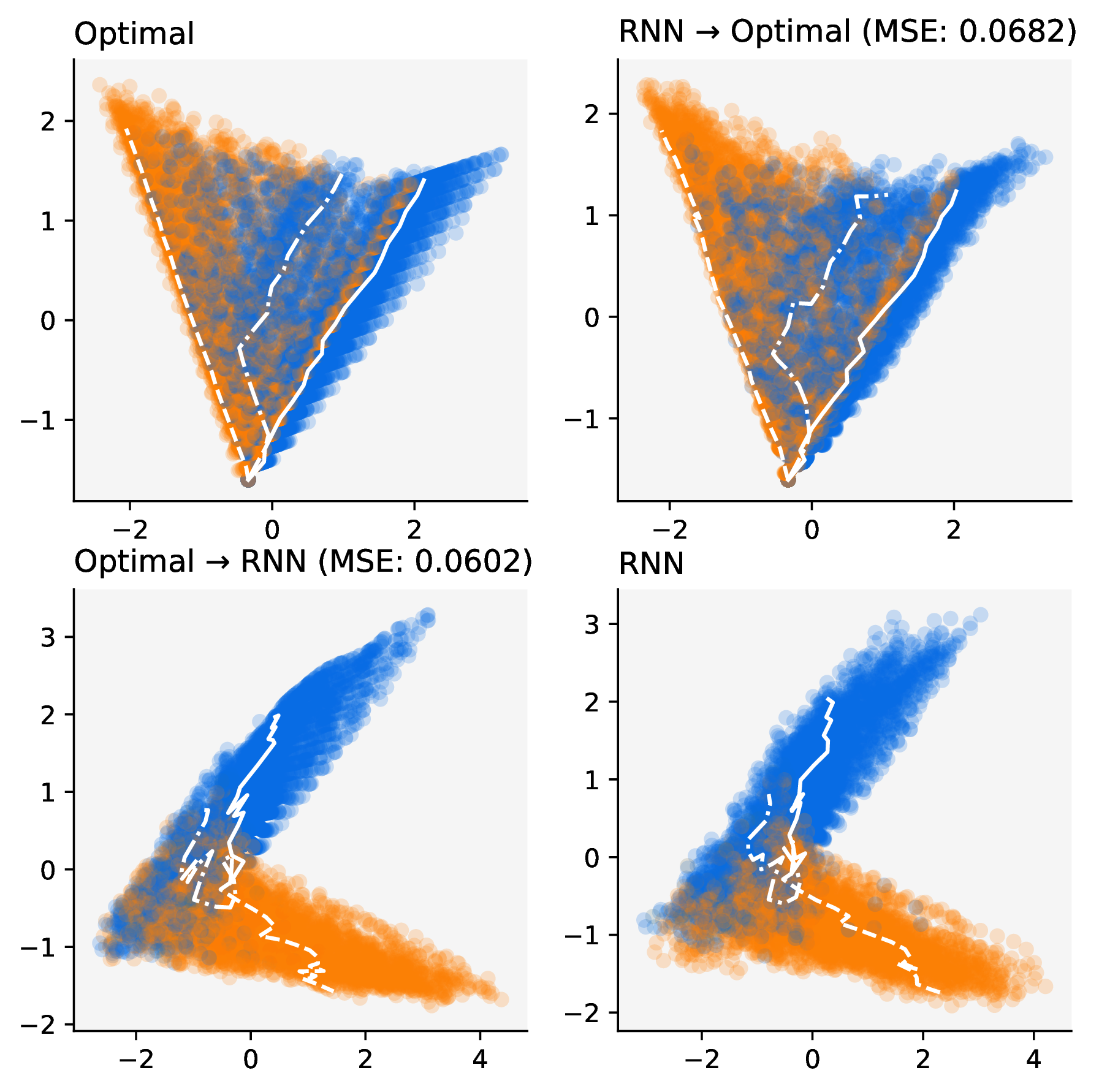}
        \caption{$\mu_1, \mu_2 \sim \mathrm{Normal}(0, 0.1)$}
    \end{subfigure}
    \caption{Structural comparison III (bandit tasks). Action probabilities are color-coded.}
    \label{fig:structure-iii}
\end{figure}

We report the structural comparison plots for all the tasks. These were generated using the same methodology as in Figure~\ref{fig:structure}. Figures~\ref{fig:structure-i}, \ref{fig:structure-ii}, and \ref{fig:structure-iii} show the comparisons for the prediction of discrete observations, prediction of continuous observations, and bandits respectively.

\subsection{Convergence analysis - additional results}\label{sec:appendix-evolution}
\begin{figure}[p]
    \vspace{-10pt}
    \centering
    \begin{subfigure}{0.48\textwidth}
        \centering
        \includegraphics[width=\textwidth]{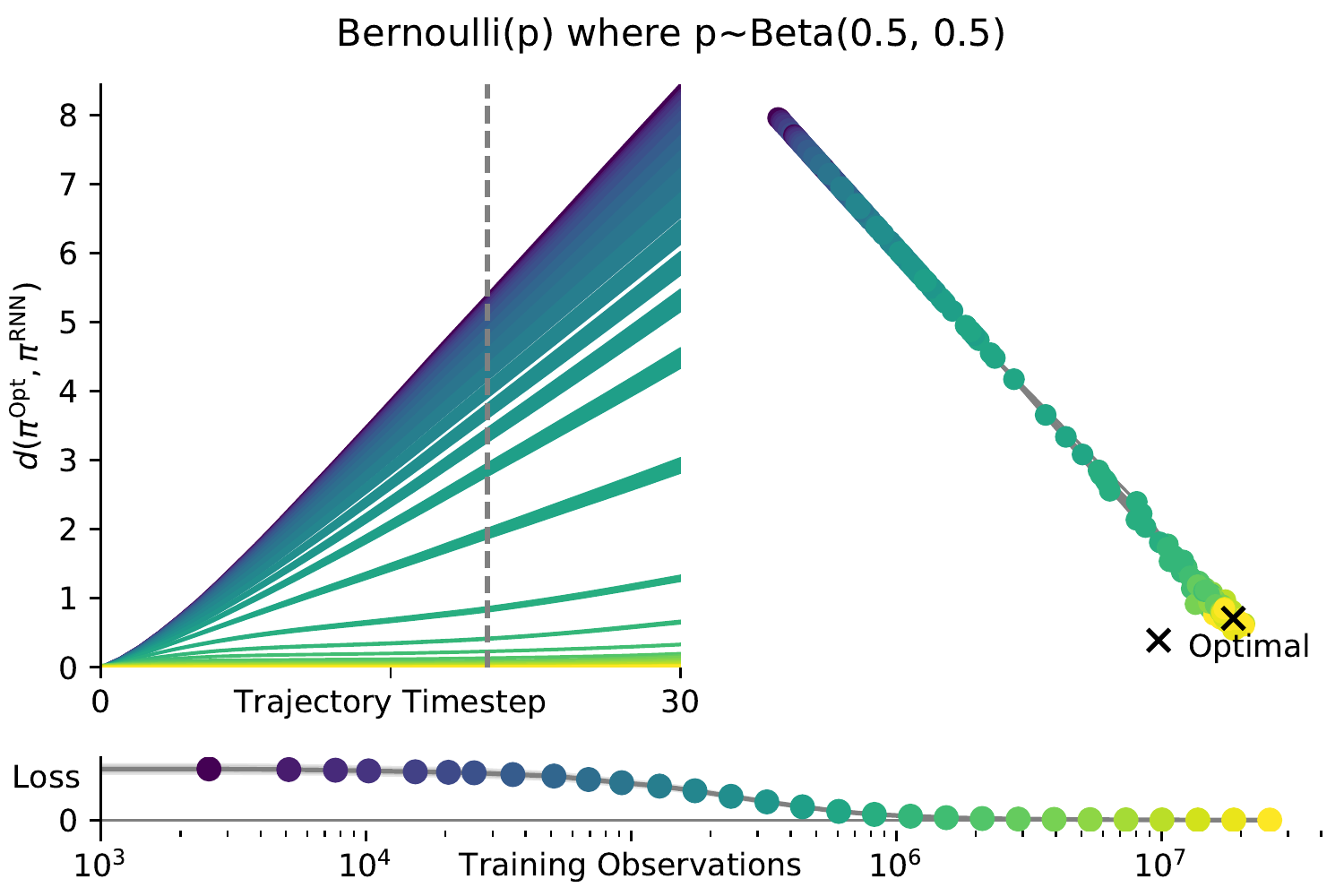}
    \end{subfigure}
    \hfill
    \begin{subfigure}{0.48\textwidth}
        \centering
        \includegraphics[width=\textwidth]{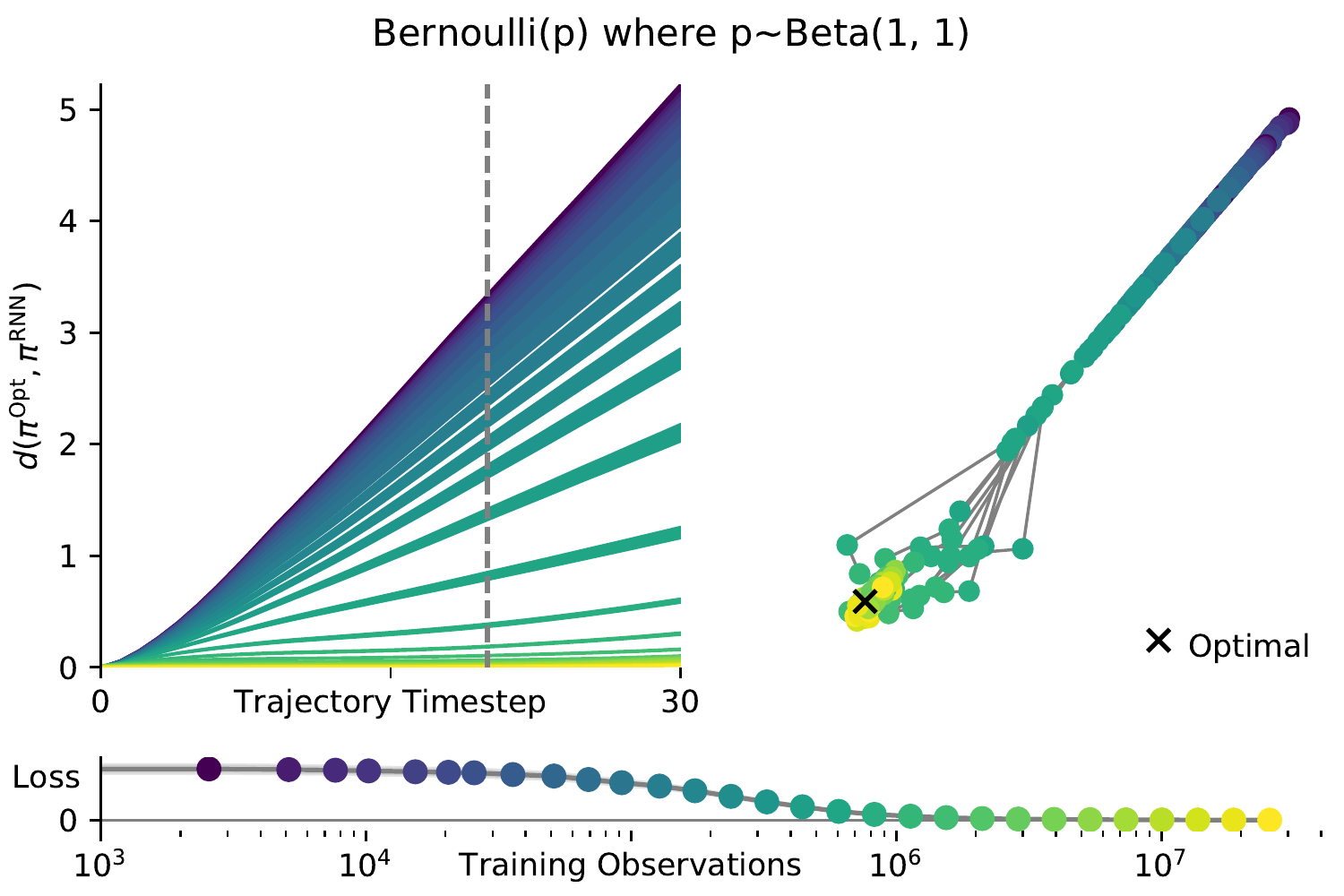}
    \end{subfigure}
    
    \begin{subfigure}{0.48\textwidth}
        \centering
        \includegraphics[width=\textwidth]{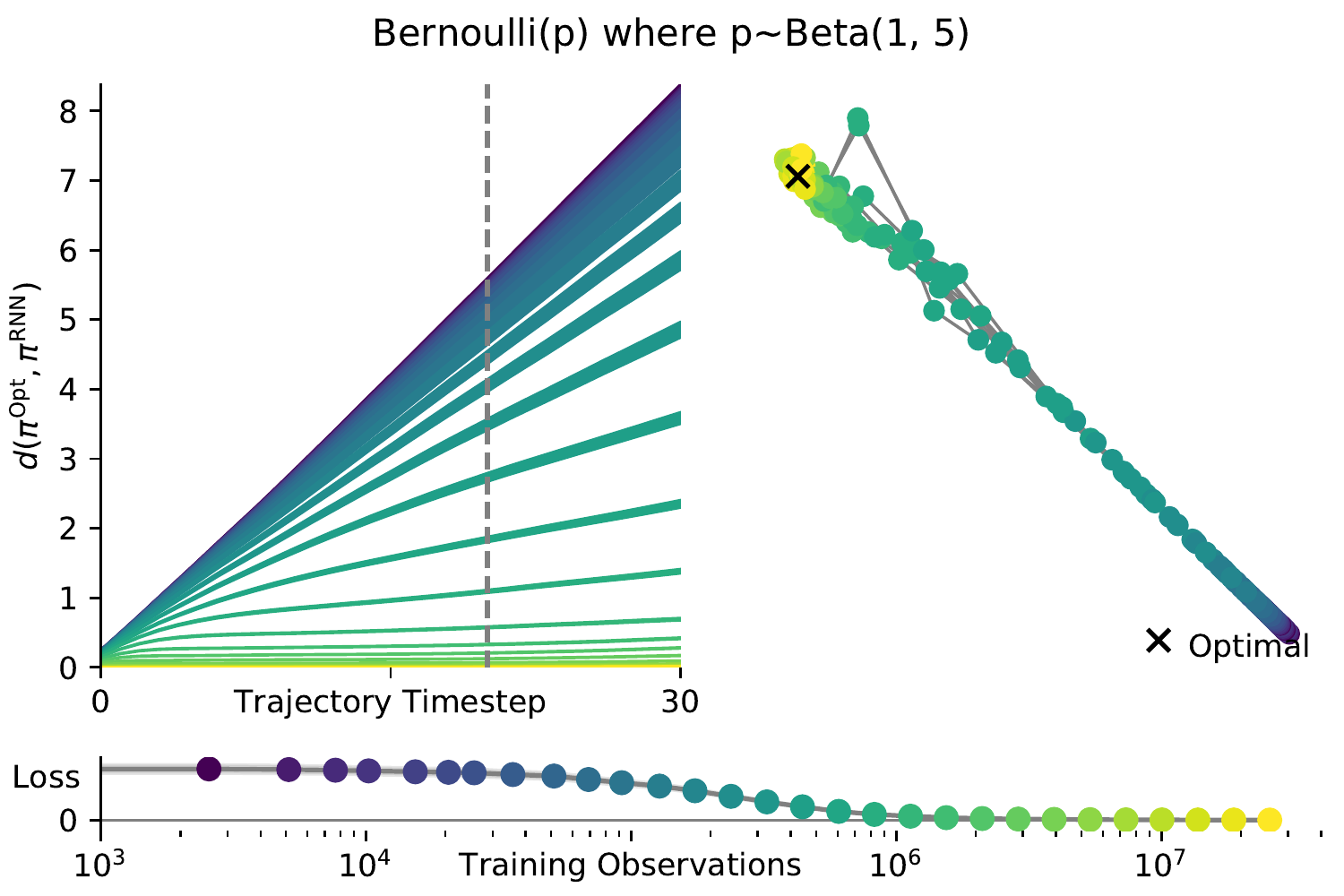}
    \end{subfigure}
    \hfill
    \begin{subfigure}{0.48\textwidth}
        \centering
        \includegraphics[width=\textwidth]{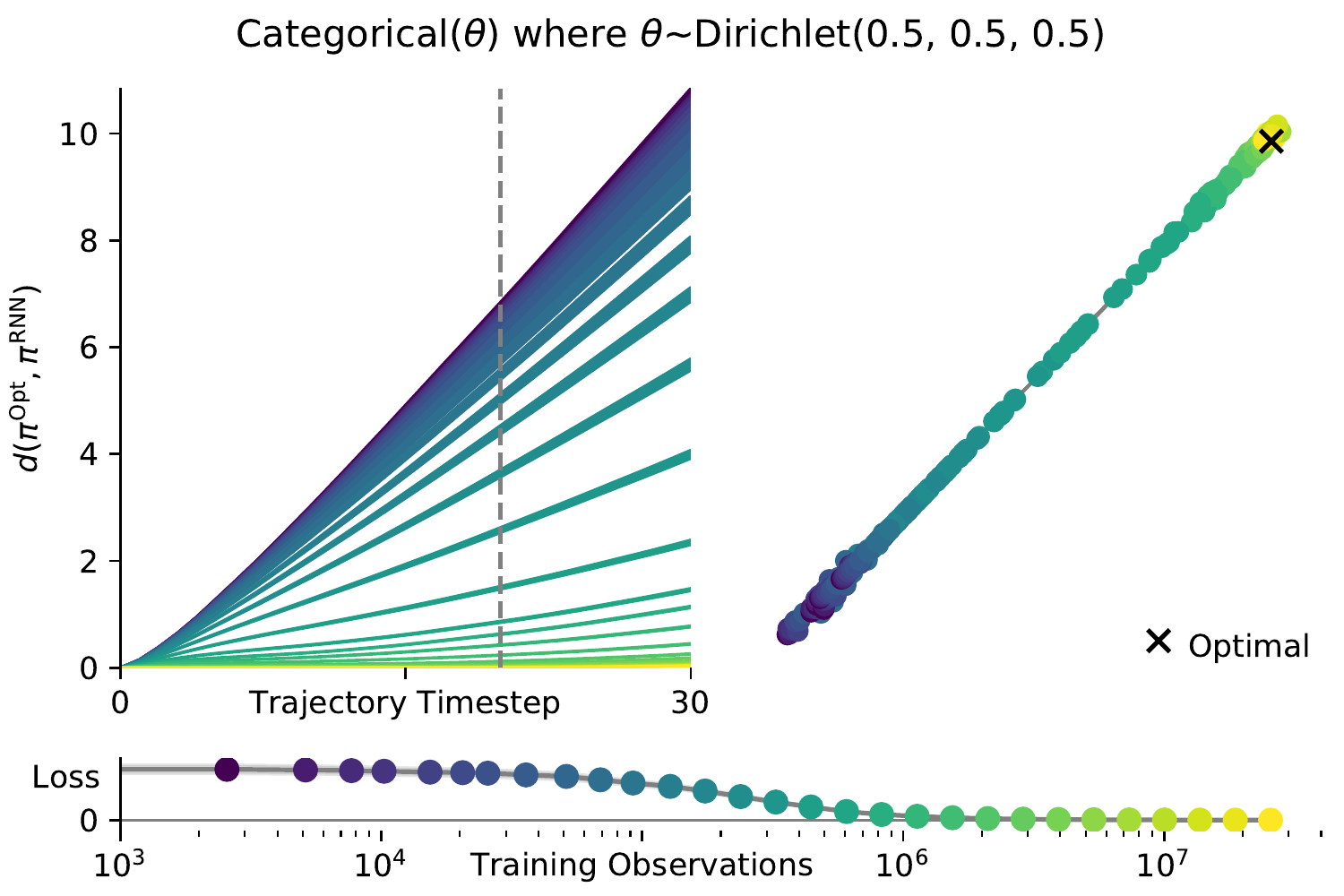}
    \end{subfigure}
    
    \begin{subfigure}{0.48\textwidth}
        \centering
        \includegraphics[width=\textwidth]{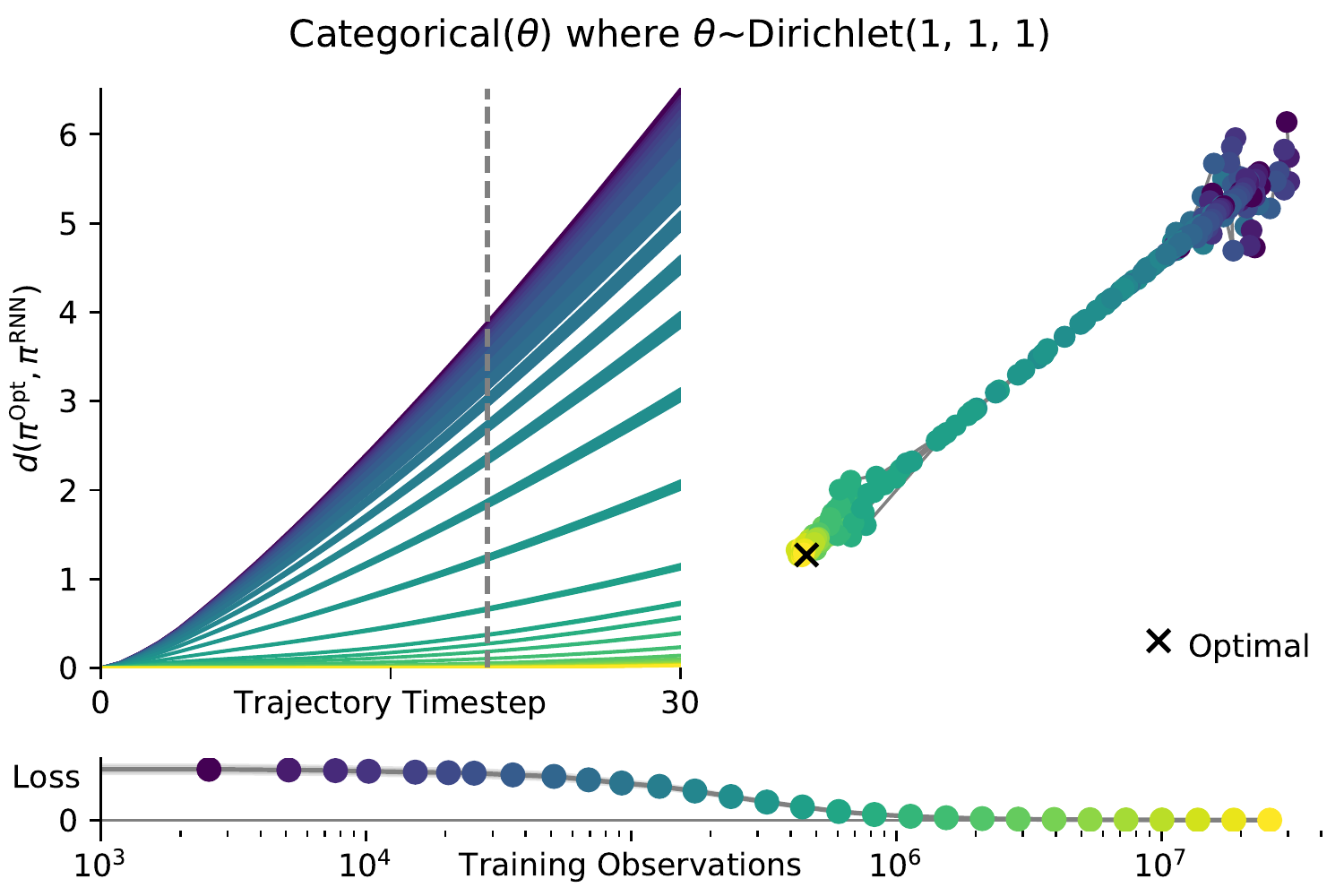}
    \end{subfigure}
    \hfill
    \begin{subfigure}{0.48\textwidth}
        \centering
        \includegraphics[width=\textwidth]{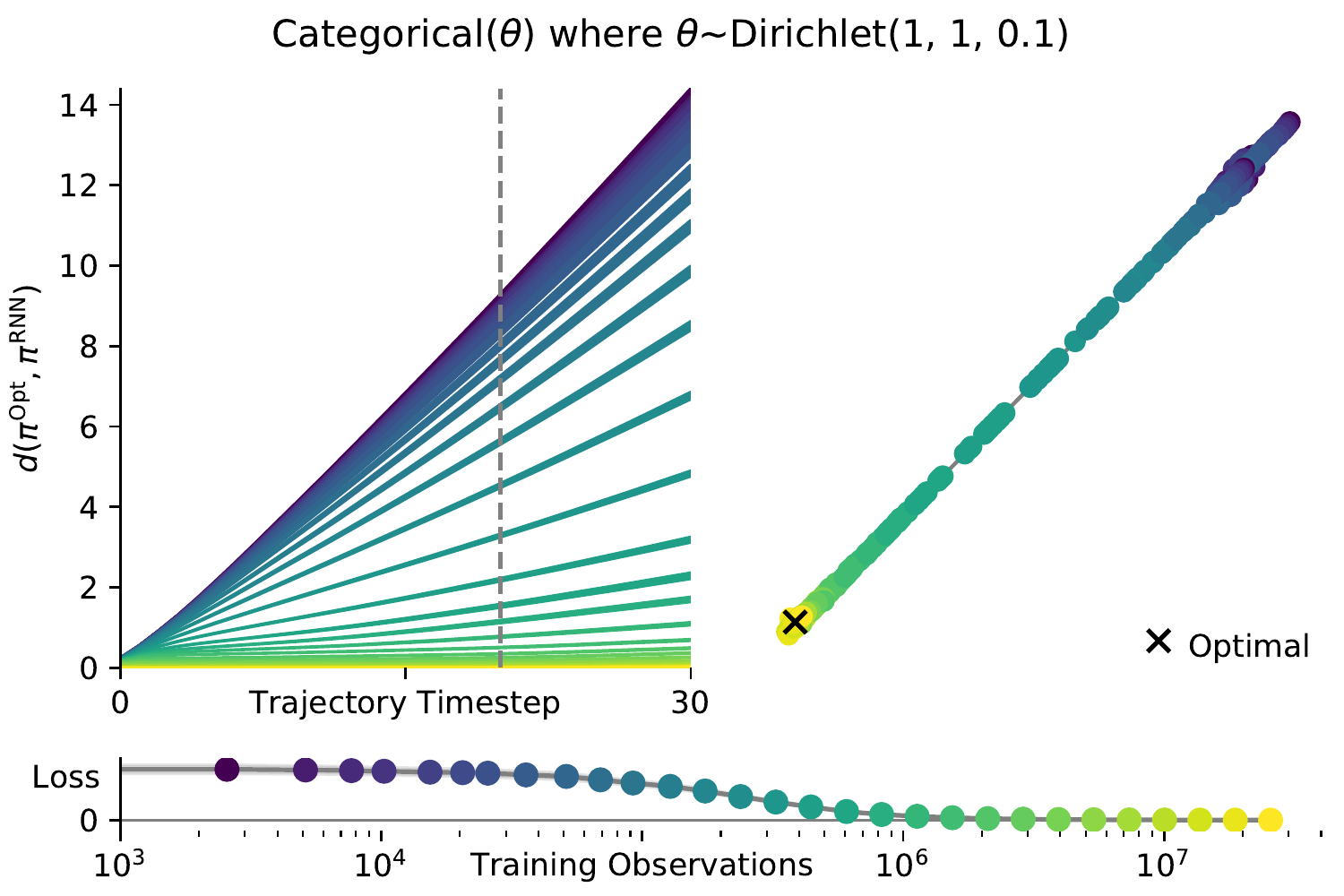}
    \end{subfigure}
    
    \begin{subfigure}{0.48\textwidth}
        \centering
        \includegraphics[width=\textwidth]{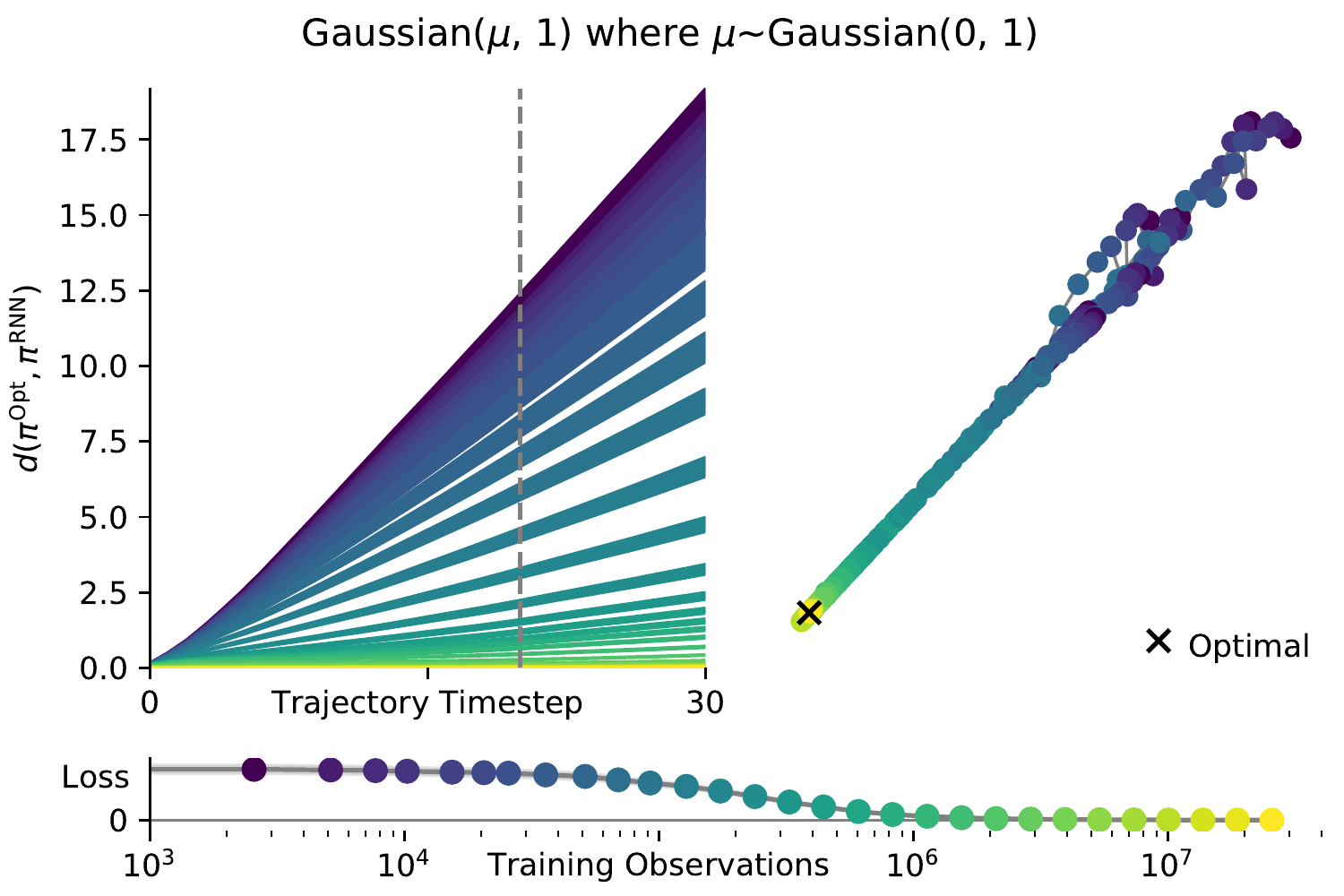}
    \end{subfigure}
    \hfill
    \begin{subfigure}{0.48\textwidth}
        \centering
        \includegraphics[width=\textwidth]{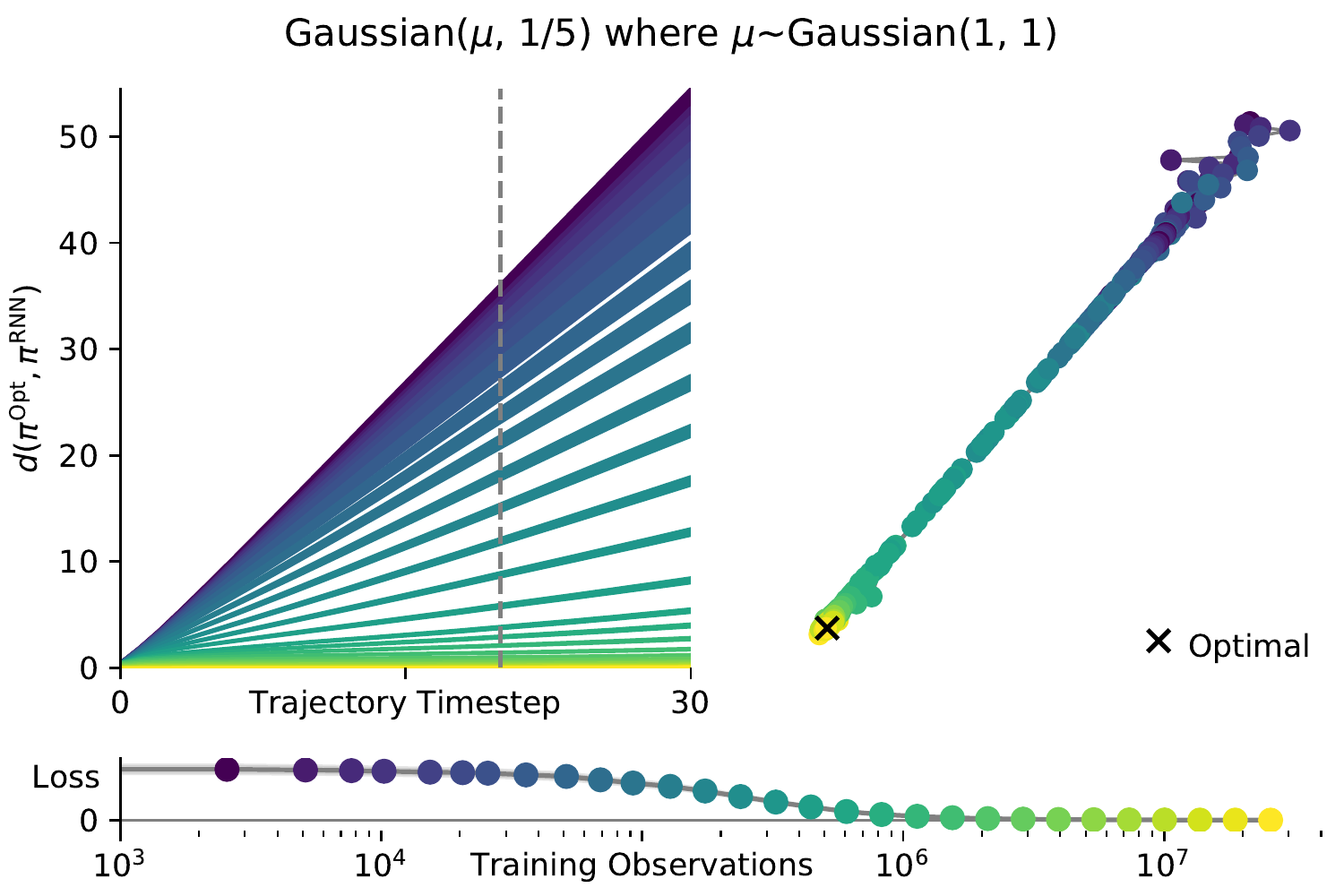}
    \end{subfigure}
    
    \caption{Convergence plots for our prediction tasks, showing 10 steps of generalisation (demarcated by grey dashed line).}
    \label{fig:convergence-other-envs-supervised}
\end{figure}

\begin{figure}[p]
    \vspace{-10pt}
    \centering
    \begin{subfigure}{0.48\textwidth}
        \centering
        \includegraphics[width=\textwidth]{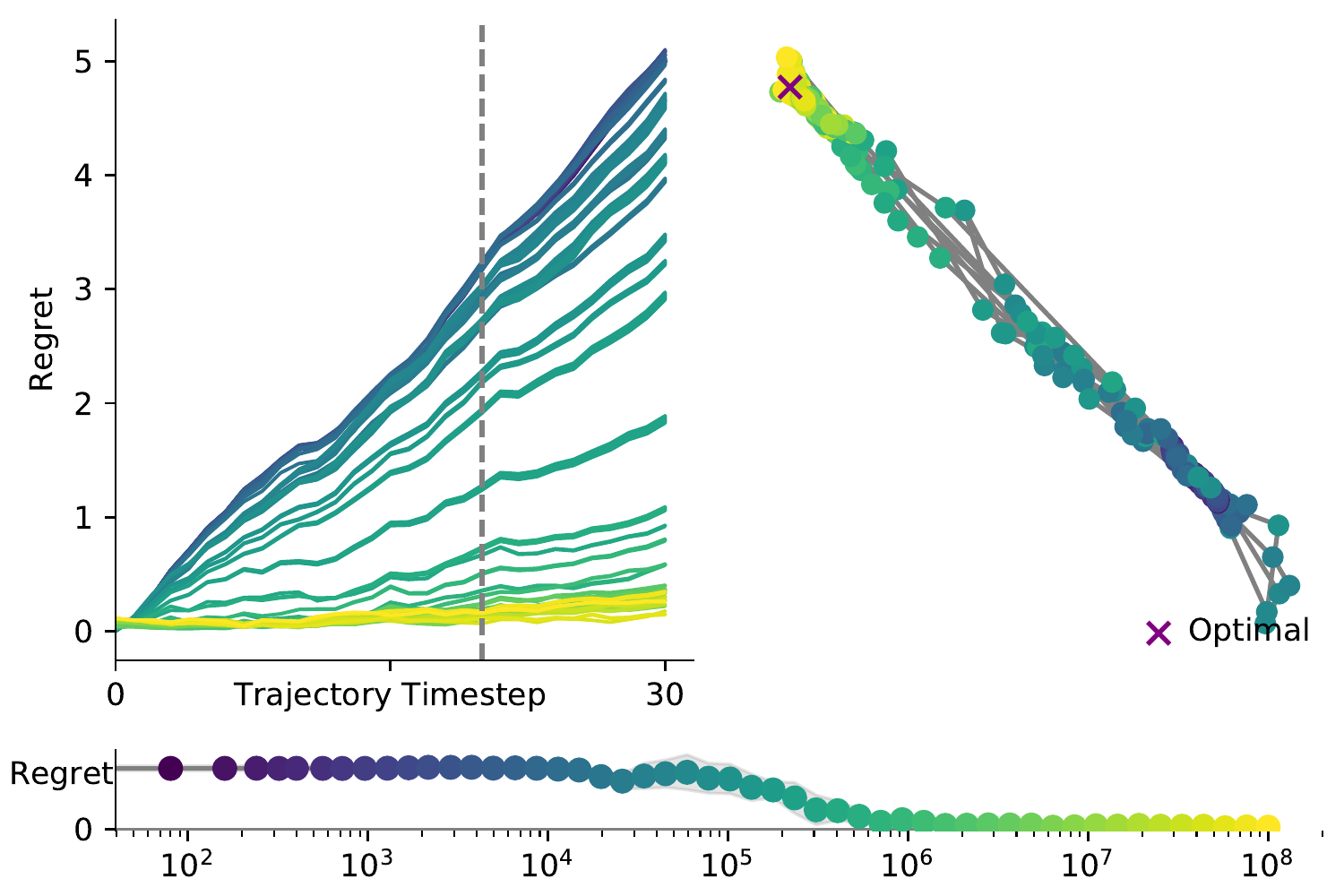}
    \caption{$\theta_1$, $\theta_2 \sim\rm{Beta}(1,1)$.}
    \end{subfigure}
    \hfill
    \begin{subfigure}{0.48\textwidth}
        \centering
        \includegraphics[width=\textwidth]{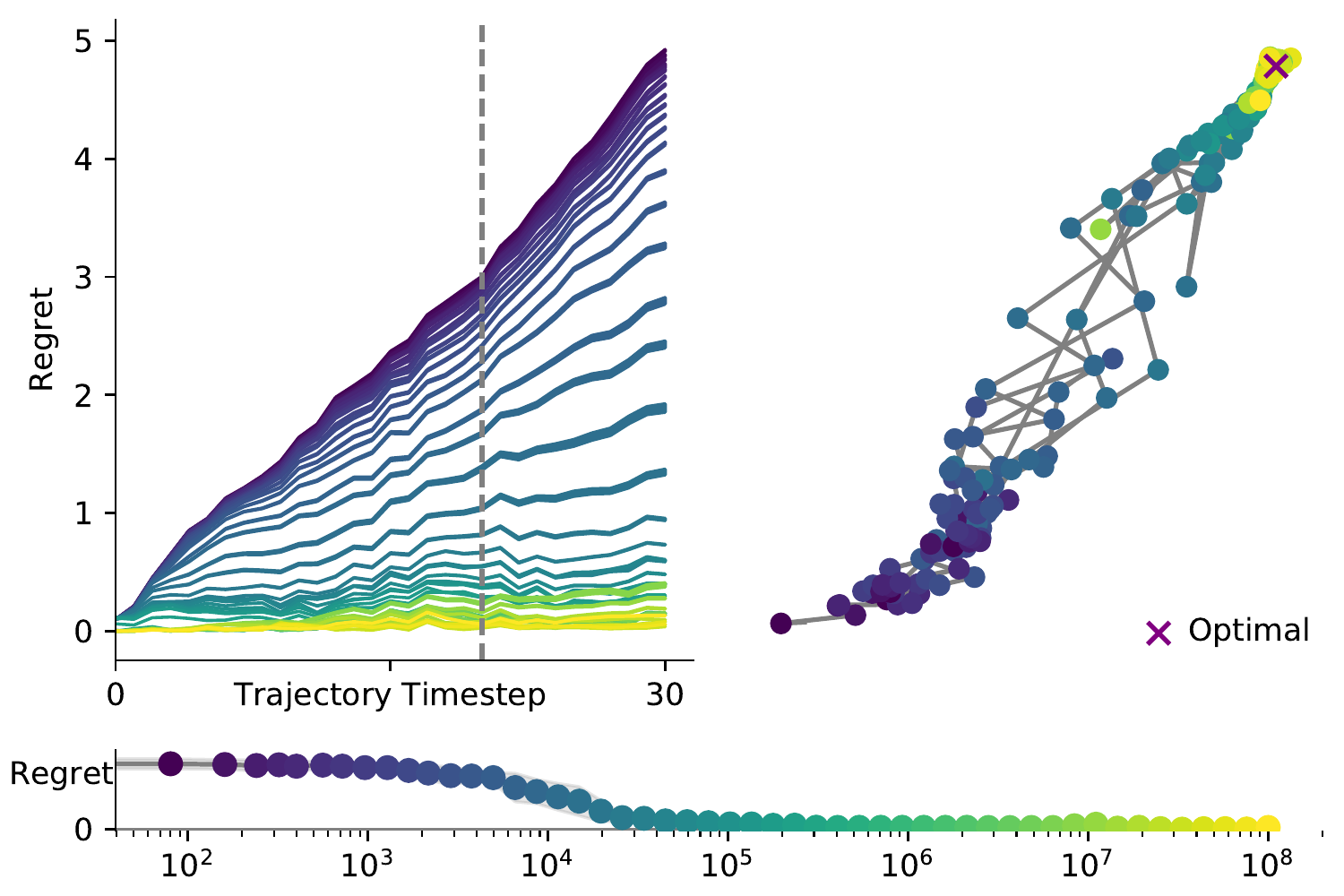}
    \caption{$\theta_1\sim\rm{Beta}(2,1),~\theta_2 \sim\rm{Beta}(1,2)$}
    \end{subfigure}
    
    \begin{subfigure}{0.48\textwidth}
        \centering
        \includegraphics[width=\textwidth]{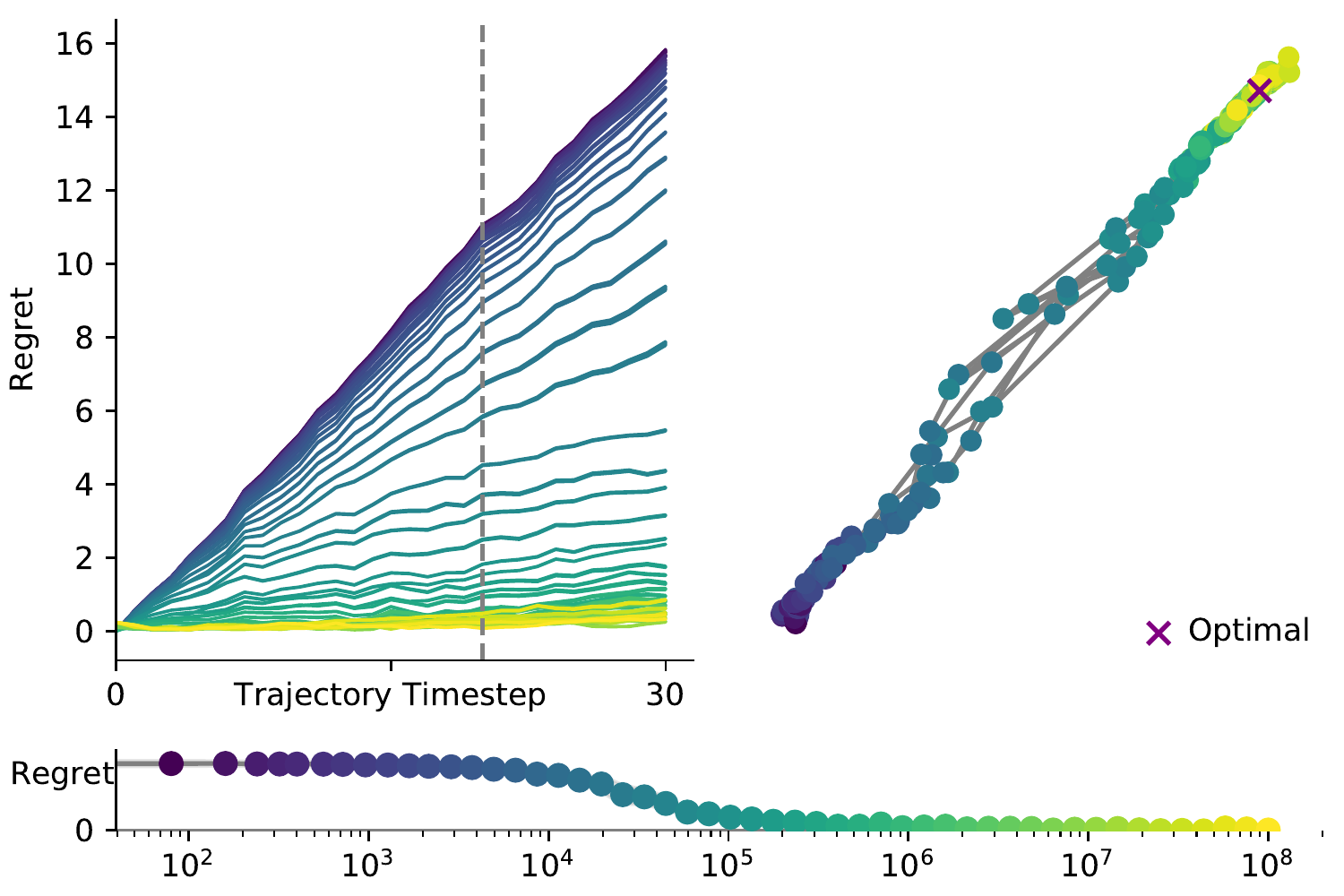}
    \caption{$\mu_1$, $\mu_2\sim\mathcal{N}(0, 1)$}
    \end{subfigure}
        \hfill
    \begin{subfigure}{0.48\textwidth}
        \centering
        \includegraphics[width=\textwidth]{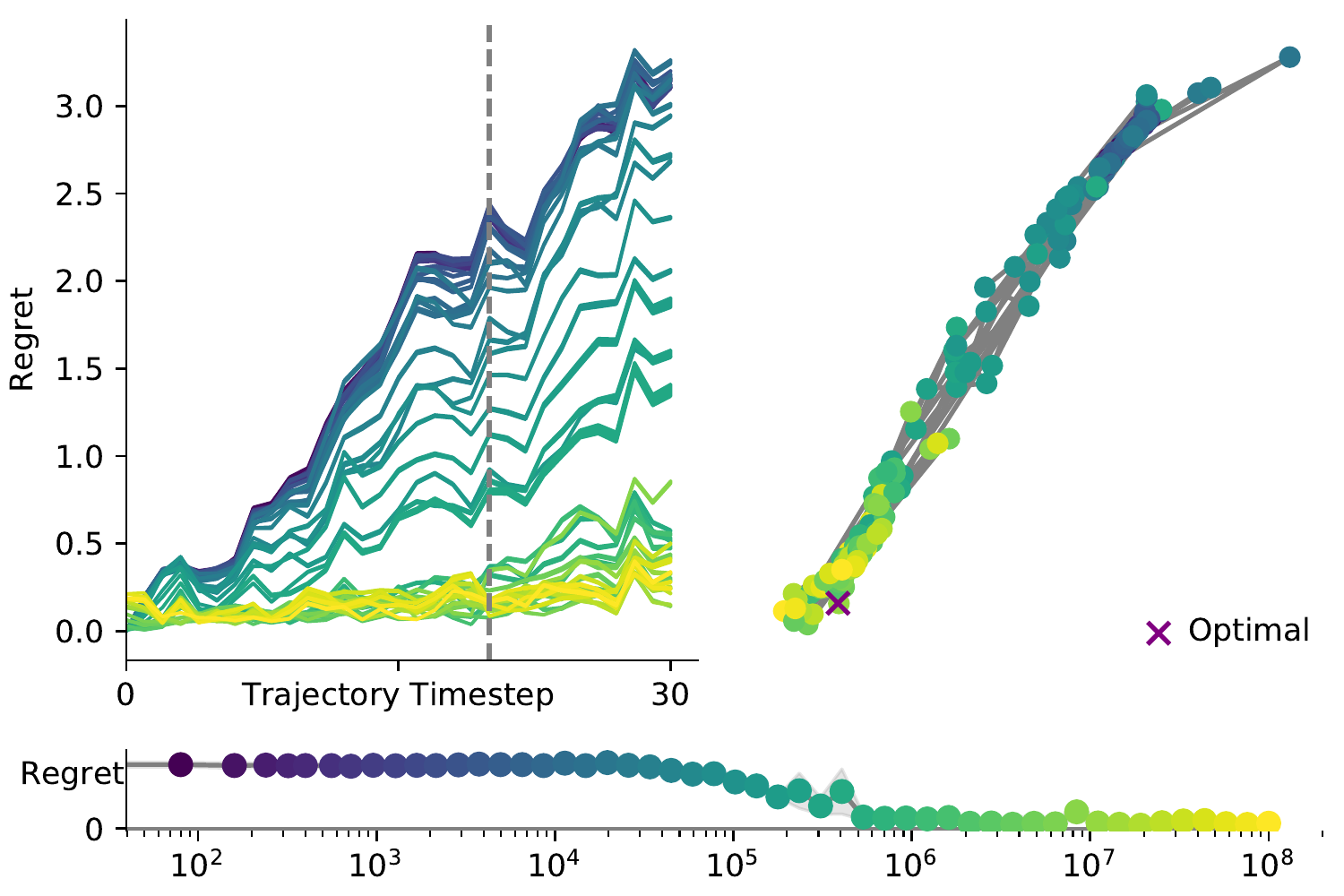}
    \caption{$\mu_1$, $\mu_2 \sim\mathcal{N}(0, 0.01)$}
    \end{subfigure}
    
    \caption{Convergence plots for bandit tasks, showing 10 steps of generalisation (demarcated by grey dashed line).}
    \label{fig:convergence-other-envs-bandit}
\end{figure}

Convergence plots for all our tasks (except the two exponential prediction tasks, where the KL-divergence estimation for the Lomax distribution can cause numerical issues that lead to bad visual results) are shown in Figure~\ref{fig:convergence-other-envs-supervised} and Figure~\ref{fig:convergence-other-envs-bandit}. Note that our agents were trained with episodes of $20$ steps, and the figures show how agents generalize when evaluated on episodes of $30$ steps.

\newpage
\subsection{Reduced-memory agents}\label{ss:reduced-mem}
In order to understand outcomes when the optimal policy is not in the search space we investigated the performance of a series of reduced-memory baselines. These were implemented with purely feedfoward architectures, which observed a context window of the previous $k$ timesteps (padded for $t<k$), rather than with an LSTM. Short context windows dramatically impaired performance, and the degree to which longer context windows allowed for improved performance was strongly task-dependent. In some cases (Dirichlet and high-precision Gaussian), extending the context window to match the episode length almost completely recovers performance, whereas in other cases performance plateaus.
\begin{figure}[ht]
    \centering
    \begin{subfigure}{0.45\textwidth}
        \centering
        \includegraphics[width=\textwidth]{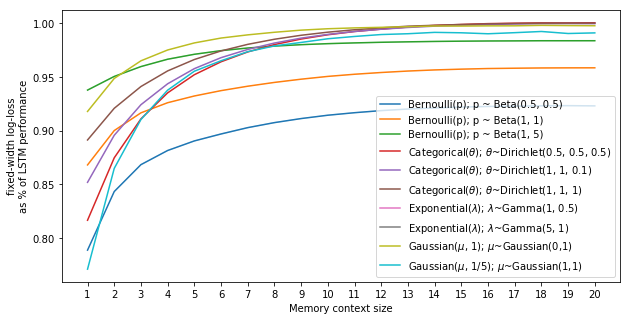}
        \caption{Prediction tasks.}
        \label{fig:reduced_memo_predict}
    \end{subfigure}
        \begin{subfigure}{0.45\textwidth}
        \centering
        \includegraphics[width=\textwidth]{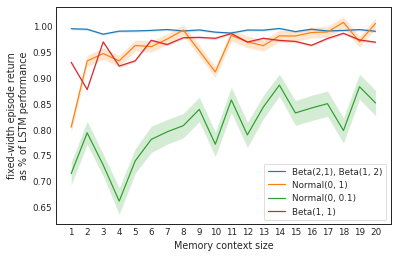}
        \caption{Bandit tasks}
        \label{fig:reduced_memo_bandits}
    \end{subfigure}
    \caption{Performance as a percentage of LSTM agent score for reduced-memory baselines. Solid line is mean over 20 trials, shaded area shows standard error of the mean over 20 repetitions. Reduced-memory baselines are feedforward agents trained with a fixed-width context of past observations. Adjusting the context width scales the amount of history the agent can use when computing a prediction/action decision.}\label{fig:reducedmemo}
\end{figure}

\end{document}